\documentclass[10pt,twocolumn,letterpaper]{article}

\usepackage{cvpr}              

\usepackage{booktabs}       
\usepackage{multirow}       
\usepackage{xcolor}         
\usepackage{pifont}         
\usepackage{threeparttable} 
\usepackage{makecell}

\usepackage{graphicx}
\usepackage[table]{xcolor} 

\usepackage{algorithm}
\usepackage{algorithmic}
\usepackage{amsmath} 

\usepackage{tcolorbox}
\usepackage{colortbl}

\definecolor{cvprblue}{rgb}{0.21,0.49,0.74}
\usepackage[pagebackref,breaklinks,colorlinks,allcolors=cvprblue]{hyperref}

\def\paperID{} 
\def\confName{CVPR}
\def\confYear{2026}

\title{ENC-Bench: A Benchmark for Evaluating Multimodal Large Language Models in Electronic Navigational Chart Understanding}

\author{
Ao Cheng$^{1}$ \quad
Xingming Li$^{1}$ \quad
Xuanyu Ji$^{1}$ \quad
Xixiang He$^{1}$ \quad
Qiyao Sun$^{1}$ \\
Chunping Qiu$^{2}$ \quad
Runke Huang$^{3}$ \quad
Qingyong Hu$^{2}$\thanks{Corresponding author.} \\
$^{1}$National University of Defense Technology \quad
$^{2}$Intelligent Game and Decision Lab \\
$^{3}$The Chinese University of Hong Kong, Shenzhen \\
}

\begin{document}
\maketitle

\begin{abstract}
Electronic Navigational Charts (ENCs) are the safety-critical backbone of modern maritime navigation, yet it remains unclear whether multimodal large language models (MLLMs) can reliably interpret them. Unlike natural images or conventional charts, ENCs encode regulations, bathymetry, and route constraints via standardized vector symbols, scale-dependent rendering, and precise geometric structure---requiring specialized maritime expertise for interpretation. 
We introduce \textbf{ENC-Bench}, the first benchmark dedicated to professional ENC understanding. ENC-Bench contains 20{,}490 expert-validated samples from 840 authentic National Oceanic and Atmospheric Administration (NOAA) ENCs, organized into a three-level hierarchy: \textbf{Perception} (symbol and feature recognition), \textbf{Spatial Reasoning} (coordinate localization, bearing, distance), and \textbf{Maritime Decision-Making} (route legality, safety assessment, emergency planning under multiple constraints). All samples are generated from raw S\textendash57 data through a calibrated vector-to-image pipeline with automated consistency checks and expert review. We evaluate \textbf{10} state-of-the-art MLLMs such as GPT-4o, Gemini 2.5, Qwen3-VL, InternVL-3, and GLM-4.5V, under a unified zero-shot protocol. The best model achieves only 47.88\% accuracy, with systematic challenges in symbolic grounding, spatial computation, multi-constraint reasoning, and robustness to lighting and scale variations. By establishing the first rigorous ENC benchmark, we open a new research frontier at the intersection of specialized symbolic reasoning and safety-critical AI, providing essential infrastructure for advancing MLLMs toward professional maritime applications.
\end{abstract}    
\section{Introduction}
\label{sec:intro}

\begin{figure*}[!t]
\centering
\includegraphics[width=\textwidth]{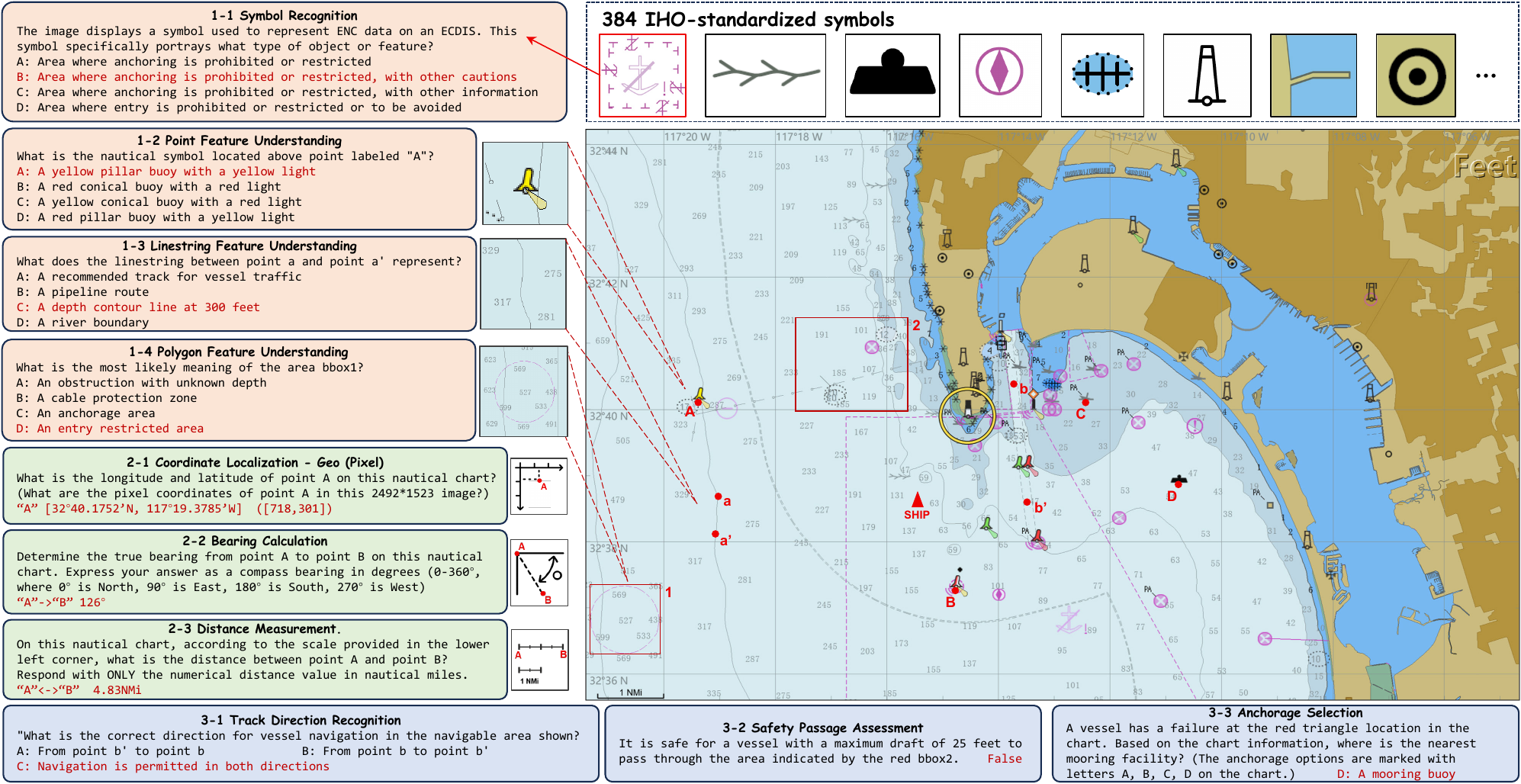}
\caption{\textbf{Overview of ENC-Bench.} Our benchmark evaluates MLLMs across three hierarchical tiers: Perception (L-1), Spatial Reasoning (L-2), and Maritime Decision-Making (L-3). Example tasks are shown on the left with corresponding visual elements on an authentic NOAA chart (right), demonstrating the progression from basic symbol interpretation to complex multi-constraint decision-making required in professional maritime navigation.}
\label{fig:overview}
\end{figure*}

    Maritime transportation underpins over 90\% of global trade, yet over 26,000 marine casualties occurred in EU waters between 2014 and 2023~\cite{allianz2024safety,emsa2024overview}, with catastrophic incidents exceeding USD 400 million in losses~\cite{frontiers2025maritime}. \textbf{Electronic Navigational Charts} (ENCs) are now mandatory for commercial vessels under International Maritime Organization regulations, with paper charts being phased out by 2030~\cite{hartis2024enc,noaa2024enc}. As maritime AI reaches 4.13 billion in market value with demonstrated potential to reduce collision risks by 75\%~\cite{orcaai2025maritime,lloyds2024ai}, reliable AI interpretation of ENCs becomes critical for next-generation safety systems.

    However, ENCs represent a \textbf{highly specialized vertical domain} that fundamentally differs from general visual content~\cite{palikaris2020electronic, blindheim2021electronic,hecht2006electronic}. Unlike natural images or statistical charts, ENCs encode safety-critical information through three distinctive characteristics that demand professional maritime expertise: (1) \textbf{standardized symbolic systems} (International Hydrographic Organisation, IHO S-57)~\cite{iho200057} where regulatory information, bathymetric features, and navigational aids are represented through formal vector symbols with precise semantic meanings; (2) \textbf{scale-dependent multi-layer rendering} where cartographic features dynamically appear, disappear, or transform across zoom levels following cartographic generalization principles; and (3) \textbf{multi-constraint spatial geometry} requiring simultaneous evaluation of water depth, vessel draft, restricted zones, traffic separation schemes, and environmental conditions. This domain specificity necessitates systematic evaluation of whether AI systems possess the requisite symbolic grounding and structured reasoning capabilities.

    Multimodal Large Language Models (MLLMs)  ~\cite{xu2025qwen3,hurst2024gpt,comanici2025gemini,meta2025llama,glm45v,wang2025internvl3} have demonstrated remarkable capabilities in general visual understanding tasks, excelling on benchmarks spanning everyday photographs, common objects, and natural scenes  \cite{lin2014microsoft,goyal2017making,gurari2018vizwiz,hudson2019gqa,marino2019ok,singh2019towards,zhou2018towards}. Yet research consistently reveals a critical limitation: models trained predominantly on general-domain visual data struggle when transferred to specialized domains requiring domain-specific symbolic systems and structured spatial reasoning~\cite{schulze2025visual,wu2023multimodal,zhang2024mm,yin2024survey,caffagni2024revolution,cui2024survey}. This raises a fundamental question: \textit{Can current MLLMs bridge the gap between general visual understanding and the symbolic, structured, and safety-critical domain of ENCs?}

    Unfortunately, existing benchmarks fail to tackle this issue. Chart understanding benchmarks~\cite{masry2022chartqa,kembhavi2016diagram,mathew2022infographicvqa,xia2025chartx,huang2025evochart} focus on statistical plots with unstructured visual elements rather than formal geospatial symbology. Document benchmarks~\cite{mathew2021docvqa,liu2024ocrbench,deng2025longdocurl} emphasize text extraction and layout understanding over geometric spatial reasoning. Geographic reasoning benchmarks~\cite{srivastava2025mapiq,yerramilli2025geochain,shiri2024empirical} explore location identification but lack the intersection of \textbf{standardized symbolic interpretation}, \textbf{multi-scale cartographic reasoning}, and \textbf{multi-constraint safety decision-making} that defines professional maritime navigation. No benchmark systematically evaluates whether MLLMs possess the specialized cognitive faculties required for ENC comprehension---a critical capability gap as AI systems increasingly integrate into maritime safety infrastructure.

    To bridge this gap, we introduce \textbf{ENC-Bench}, the first comprehensive benchmark for evaluating MLLM capabilities in ENCs understanding. Designed to mirror the cognitive pipeline of certified maritime navigators---from symbol recognition to safety-critical decision-making under multiple constraints---ENC-Bench provides a rigorous testbed for assessing model readiness in specialized, high-stakes visual domains. Our contributions are:

\newcommand{\cmark}{\textcolor[HTML]{00B050}{\ding{51}}}
\newcommand{\pmark}{\textcolor[HTML]{FF8C00}{\(\blacktriangle\)}}
\newcommand{\xmark}{\textcolor[HTML]{C00000}{\ding{55}}}

\begin{table*}[thb]
\centering
\caption{Comparison of \textbf{ENC-Bench} with existing benchmarks. \cmark, \pmark, and  \xmark  separately represent \textbf{full support} (supports the core capability), \textbf{partial support} (e.g., informal/math symbols, Euclidean/layout reasoning, or multi-resolution), and \textbf{no support}.}
\label{tab:benchmark_comparison}
\small
\begin{tabular}{@{}r|l|c|c|c|c@{}}
\toprule
& & \textbf{Standardized} & \textbf{Precise Geospatial} & \textbf{Multi-Scale} & \textbf{Multi-Light} \\
\textbf{Benchmark} & \textbf{Domain} & \textbf{Symbol Recognition} & \textbf{Reasoning (Numerical)} & \textbf{(Cartographic)} & \textbf{(Rendered Modes)} \\
\midrule
MME~\cite{fu2024mme} & \textit{General} & \xmark & \xmark & \xmark & \xmark \\
MMBench~\cite{liu2024mmbench} & \textit{General} & \xmark & \xmark & \xmark & \xmark \\
MMMU~\cite{yue2024mmmu} & \textit{General} & \pmark & \pmark & \xmark & \xmark \\
\midrule
DocVQA~\cite{mathew2021docvqa} & \textit{Chart/Doc} & \xmark & \pmark & \xmark & \xmark \\
ChartQA~\cite{masry2022chartqa} & \textit{Chart/Doc} & \xmark & \cmark & \xmark & \xmark \\
\midrule
MathVista~\cite{lu2024mathvista} & \textit{Geometry/Math} & \pmark & \cmark & \xmark & \xmark \\
GeoQA~\cite{chen2021geoqa} & \textit{Geometry/Math} & \pmark & \cmark & \xmark & \xmark \\
\midrule
MapQA~\cite{deng2025mapqa} & \textit{Map} & \xmark & \pmark & \xmark & \xmark \\
MapEval~\cite{dihan2025mapeval} & \textit{Map} & \xmark & \cmark & \xmark & \xmark \\
\midrule
RSVQA~\cite{lobry2020rsvqa} & \textit{Remote Sensing} & \xmark & \pmark & \pmark & \xmark \\
SkyScript~\cite{wang2024skyscript} & \textit{Remote Sensing} & \xmark & \xmark & \pmark & \xmark \\
\midrule
\textbf{ENC-Bench (Ours)} & \textit{Maritime} & \cmark & \cmark & \cmark & \cmark \\
\bottomrule
\end{tabular}
\end{table*}
    
\begin{itemize}[leftmargin=*]
\setlength{\itemsep}{0pt}
\setlength{\parsep}{0pt}
\setlength{\parskip}{0pt}
\item \textbf{Professional-Grade Dataset.} We construct 20,490 samples from 840 authentic NOAA charts conforming to IHO S-57 standards. Each sample undergoes rigorous quality control through automated consistency validation and systematic  expert review, ensuring alignment with professional maritime practice and safety protocols.

\item \textbf{Hierarchical Evaluation Framework.} We design a three-tier assessment pipeline (Perception $\rightarrow$ Spatial Reasoning $\rightarrow$ Maritime Decision-Making) that decomposes ENC understanding from basic symbol recognition through coordinate-based geometric reasoning to complex, multi-constraint safety judgments---mirroring the cognitive hierarchy required in maritime navigation.

\item \textbf{Comprehensive Evaluation and Analysis.} We conduct extensive zero-shot evaluation of \textbf{10} state-of-the-art MLLMs including GPT-4o, Gemini 2.5, Qwen3-VL, GLM-4.5V, InternVL-3, and Llama-4, revealing severe capability gaps: the best model achieves only 47.88\% accuracy. Through in-depth error analysis, we identify three fundamental limitations---\textit{symbolic grounding bottleneck} (failure to interpret formal notation like coordinate grids and scale bars), \textit{multi-constraint reasoning deficiency} (greedy local optimization rather than explicit global constraint satisfaction), and \textit{lack of robustness} across lighting modes and scale variations---providing concrete directions for advancing vision-language models toward deployment readiness in professional domains.
\end{itemize}
\section{Related Work}
\label{sec:related}

\noindent\textbf{Multimodal LLMs and General Visual Understanding.}
Recent MLLMs~\cite{liu2024llava,openai2023gpt4v,team2024gemini,bai2023qwen2vl,xu2025qwen3,chen2024internvl} have achieved remarkable success on general visual understanding benchmarks such as MME~\cite{fu2024mme}, MMBench~\cite{liu2024mmbench}, and MMMU~\cite{yue2024mmmu}, which evaluate perception and reasoning over natural images and everyday scenes~\cite{lin2014microsoft,goyal2017making,hudson2019gqa}. However, research consistently reveals their limitations when transferred to specialized professional domains requiring formal symbolic systems and domain-specific expertise~\cite{schulze2025visual,wu2023multimodal,zhang2024mm}.

\noindent\textbf{Chart and Document Understanding.}
Benchmarks including DocVQA~\cite{mathew2021docvqa}, ChartQA~\cite{masry2022chartqa}, and InfoVQA~\cite{mathew2022infographicvqa} assess information extraction from business documents and statistical visualizations. Specialized models such as Pix2Struct~\cite{lee2023pix2struct} and ChartLlama~\cite{han2024chartllama} advance these capabilities through targeted training. However, these works focus on \textit{pixel-rendered data visualizations} where values are encoded through informal visual geometry (bar heights, pie angles)---not \textit{vector-based standardized symbology} governed by international regulatory frameworks encoding legal constraints and safety-critical information.

\noindent\textbf{Geospatial Visual Understanding.}
Remote sensing benchmarks (RSVQA~\cite{lobry2020rsvqa}, SkyScript~\cite{wang2024skyscript}, LHRS-Bench~\cite{muhtar2024lhrs}) evaluate satellite imagery interpretation, while map understanding benchmarks (MapQA~\cite{deng2025mapqa}, MapEval~\cite{dihan2025mapeval}) assess reasoning on consumer web maps. Domain-adapted models like GeoChat~\cite{kuckreja2024geochat} and RS-LLaVA~\cite{bazi2024rsllava} demonstrate improved performance through specialized pretraining. Yet these works evaluate either \textit{observational photographic data} or \textit{informal consumer cartography} with approximate spatial representations---neither addresses \textit{safety-certified professional navigation charts} employing standardized symbology (IHO S-57), legally mandated positional accuracy, and regulatory compliance for international maritime operations.

\noindent\textbf{Structured Visual Reasoning.}
Geometry-focused benchmarks including GeoQA~\cite{chen2021geoqa}, UniGeo~\cite{chen2022unigeo}, and MathVista~\cite{lu2024mathvista} evaluate mathematical reasoning over educational diagrams and abstract geometric problems. While these assess structured reasoning capabilities, they operate within \textit{idealized Euclidean coordinate systems} suitable for textbook scenarios---distinct from \textit{spherical geodetic coordinate systems} required for real-world nautical computations involving Haversine distance, meridian convergence, and DMS notation critical for maritime route planning.

\noindent\textbf{Our Positioning: Safety-Critical Symbolic Reasoning in Maritime Navigation.}
Existing benchmarks (Table~\ref{tab:benchmark_comparison}) evaluate general vision, informal visualizations, geospatial maps, or abstract geometry---none address professional maritime navigation charts requiring regulatory compliance. ENC-Bench uniquely combines four capabilities essential for safety-critical ENC interpretation: (1)~\textit{Standardized Symbolic Recognition}---IHO S-57 regulated symbology encoding legal constraints; (2)~\textit{Precise Geospatial Reasoning}---exact coordinate computation with nautical accuracy requirements; (3)~\textit{Multi-Scale Cartographic Rendering}---scale-dependent feature aggregation following professional cartographic principles; and (4)~\textit{Multi-Lighting Operational Modes}---robustness across day/dusk/night rendering modes. By simultaneously evaluating these dimensions in authentic navigational contexts where interpretation errors lead to maritime casualties, ENC-Bench establishes essential infrastructure for advancing MLLMs from general-purpose understanding toward deployment in high-stakes professional domains.

\begin{figure*}[!t]
\centering
\includegraphics[width=\textwidth]{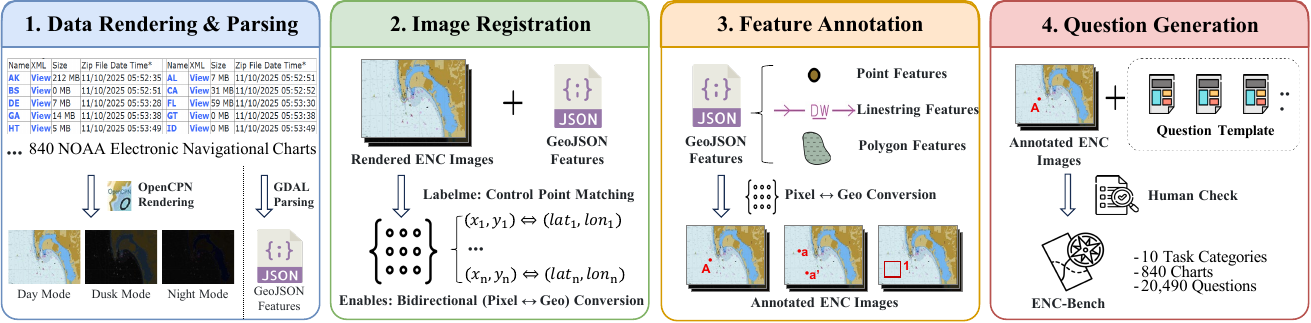}
\caption{\textbf{ENC-Bench Data Generation Pipeline.} Four-stage process transforms 840 NOAA S-57 charts into 20,490 validated samples: \textbf{(1) Rendering \& Parsing} produces multi-condition images and extracts GeoJSON features; \textbf{(2) Image Registration} establishes pixel-to-geo coordinate conversion; \textbf{(3) Feature Annotation} marks point/line/polygon features with expert verification; \textbf{(4) Question Generation} applies templates to create structured QA pairs with validated ground truth.}
\label{fig:pipeline}
\end{figure*}

\section{ENC-Bench}
\label{sec:benchmark}

We introduce ENC-Bench, a comprehensive benchmark with 20,490 expert-validated samples from 840 authentic NOAA S-57 charts, organized into three tiers (Perception $\rightarrow$ Spatial Reasoning $\rightarrow$ Maritime Decision-Making) with systematic variations across lighting modes and scale levels.

\subsection{Dataset Construction}

\subsubsection{Data Source and Coverage}

We collected 840 Electronic Navigational Charts from the National Oceanic and Atmospheric Administration (NOAA)~\cite{noaa2024enc}, which provides publicly accessible, quality-assured official ENCs actively used in operational navigation. All charts conform to the IHO S-57 standard~\cite{iho200057}, the globally recognized format for official ENCs. The dataset encompasses operational scenarios: shallow harbors, deep-water shipping lanes, traffic separation schemes, restricted zones, and ecologically sensitive areas, ensuring comprehensive representation of real-world navigation contexts.

\subsubsection{Data Generation Pipeline}

We developed a semi-automated four-stage pipeline to transform raw S-57 vector data into structured question-answer pairs (Figure~\ref{fig:pipeline}):

\noindent\textbf{Stage 1: Data Rendering \& Parsing.}
S-57 binary charts are rendered via OpenCPN~\cite{OpenCPN} across three lighting modes (day/dusk/night) following Electronic Chart Display and Information System (ECDIS) operational standards~\cite{imo2024ecdis} and six scale levels (1:50k, 1:70k, 1:100k, 1:130k, 1:200k, 1:300k). Charts are parsed into GeoJSON features using GDAL~\cite{Rouault_GDAL_2025}, with abbreviated attribute codes mapped to human-readable descriptions via official IHO lookup tables~\cite{iho200057}. Cross-referenced features are merged based on LNAM identifiers to maintain semantic integrity.

\noindent\textbf{Stage 2: Image Registration.}
We establish bidirectional pixel-to-geographic coordinate conversion through control point matching. Two geographic control points are manually labeled per chart in pixel space using Labelme~\cite{Wada_Labelme_Image_Polygonal}, enabling affine transformation matrices for precise coordinate conversion required by spatial reasoning tasks.

\noindent\textbf{Stage 3: Feature Annotation.}
Point, linestring, and polygon features are systematically annotated. Point features are grouped using graph-coloring algorithms to prevent visual overlap, with manual refinement for ambiguous cases. Linear features are marked at endpoints. Polygons use axis-aligned bounding boxes. Density control ensures visual clarity with a median of 8 features per image. All annotations are verified through expert review to ensure correctness and nautical plausibility.

\noindent\textbf{Stage 4: Question Generation.}
We apply task-specific templates to generate 20,490 questions across 10 categories. Ground truth for spatial reasoning tasks is computed using validated nautical formulas: Haversine distance for nautical mile calculations, bearing computation via arctangent of coordinate differences adjusted for true north, and affine transformation for coordinate localization. Multiple-choice distractors are systematically generated based on common navigation errors (e.g., reversed bearings, incorrect unit conversions, overlooked constraints).

\subsubsection{Quality Control}

We implement a rigorous two-stage validation protocol. \textbf{Stage 1 (Automated Verification):} Each question's answer is cross-checked against original chart file attributes, validating coordinates through transformation matrices, depth values via S-57 SOUNDG objects, feature classifications through object class codes, and computed spatial metrics using reference implementations. \textbf{Stage 2 (Expert Review):} All generated questions undergo review by maritime navigation professionals for formatting consistency, linguistic clarity, and plausibility of multiple-choice options, ensuring alignment with professional practice.

\subsection{Benchmark Design}

We structure ENC-Bench into three tiers with 10 task dimensions reflecting the cognitive hierarchy of maritime navigation: perception, spatial reasoning, and decision-making, as illustrated in Figure~\ref{fig:overview}.

\subsubsection{L-1 Perception (4 Tasks)}

Perception tasks assess the ability to interpret standardized maritime symbology in authentic chart contexts, testing both isolated symbol recognition and contextual feature understanding across three geometry types.

\noindent\textbf{Symbol Recognition.} Models identify IHO-standardized symbols from isolated visual appearance, testing direct visual-to-semantic mapping of maritime notation.

\noindent\textbf{Point Feature Understanding.} Models extract structured attributes from point features (buoys, lighthouses) embedded in complete chart scenes.

\noindent\textbf{Linestring Feature Understanding.} Models identify types and directional semantics of linear features (tracks, contours) marked with endpoints in full chart contexts.

\noindent\textbf{Polygon Feature Understanding.} Models interpret regional semantics and safety attributes of area features (zones, restricted areas) marked with bounding boxes.

\subsubsection{L-2 Spatial Reasoning (3 Tasks)}

Spatial reasoning demands quantitative geometric computation grounded in cartographic principles, testing numerical reasoning with coordinate systems and scale conversion.

\noindent\textbf{Coordinate Localization.} Given a marked point, models predict geographic location through two approaches: outputting decimal-degree coordinates converted to pixel space, or directly predicting pixel coordinates. Both are evaluated using pixel-space error, enabling comparison of symbolic notation interpretation versus visual localization.

\noindent\textbf{Bearing Calculation.} Given two points, models compute compass bearing (0--360°, true north = 0°) between them, requiring point identification, vector angle calculation, and alignment with chart orientation.

\noindent\textbf{Distance Measurement.} Given two points, models calculate distance in nautical miles, extracting scale information, measuring pixel distance, and applying unit conversion.

\subsubsection{L-3 Maritime Decision-Making (3 Tasks)}

The highest tier evaluates synthesis of perceptual and spatial information into actionable navigation decisions under real-world multi-constraint scenarios.

\noindent\textbf{Track Direction Recognition.} Given a marked route with endpoints, models determine legal navigation direction: ``from $a$ to $a'$'', ``from $a'$ to $a$'', or ``non-navigable'', interpreting traffic separation schemes and routing regulations.

\noindent\textbf{Safety Passage Assessment.} Given a vessel's draft and passage, models judge if safe transit is possible (True/False), requiring extraction of depth soundings, identification of minimum depth, and application of safety margins.

\begin{table}[t]
\centering
\renewcommand{\arraystretch}{0.9}
\caption{Key statistics of the ENC-Bench benchmark. $^*$Excluding 384 Symbol Recognition samples (isolated symbols without contextual rendering). Contextual tasks total: 20,106 samples.}
\label{tab:statistics}
\small
\begin{tabular}{lc}
\toprule
\textbf{Statistics} & \textbf{Number} \\
\midrule
Total Questions & 20,490 \\
Total Charts & 840 \\
Task Categories & 10 \\
\midrule
\textbf{Task Hierarchy} & \\
\quad Perception & 14,886 (72.6\%) \\
\quad\quad - Symbol Recognition & 384 \\
\quad\quad - Point Features & 7,830 \\
\quad\quad - Linestring Features & 2,562 \\
\quad\quad - Polygon Features & 4,110 \\
\quad Spatial Reasoning & 4,338 (21.2\%) \\
\quad\quad - Coordinate Localization & 2,016 \\
\quad\quad - Bearing Calculation & 1,161 \\
\quad\quad - Distance Measurement & 1,161 \\
\quad Maritime Decision-Making & 1,266 (6.2\%) \\
\quad\quad - Track Direction Recognition & 141 \\
\quad\quad - Safety Passage Assessment & 702 \\
\quad\quad - Anchorage Selection & 423 \\
\midrule
\textbf{Lighting Modes} (Contextual Tasks)$^*$ & \\
\quad Day Mode & 6,702 (33.3\%) \\
\quad Dusk Mode & 6,702 (33.3\%) \\
\quad Night Mode & 6,702 (33.3\%) \\
\midrule
\textbf{Scale Levels} (Contextual Tasks)$^*$ & \\
\quad Large Scale (1:50k, 1:70k) & 6,507 (32.4\%) \\
\quad Intermediate Scale (1:100k, 1:130k) & 7,557 (37.6\%) \\
\quad Small Scale (1:200k, 1:300k) & 6,042 (30.0\%) \\
\bottomrule
\end{tabular}
\end{table}

\noindent\textbf{Anchorage Selection.} When a vessel encounters an emergency, models select the nearest safe anchorage from four candidates: minimizing distance while ensuring adequate depth and verifying non-restricted status.

\subsection{Dataset Statistics}

\noindent\textbf{Scale and Distribution.}
Table~\ref{tab:statistics} presents key statistics. ENC-Bench comprises 20,490 samples with hierarchical distribution: Perception (72.6\%), Spatial Reasoning (21.2\%), and Maritime Decision-Making (6.2\%). This pyramid structure reflects cognitive complexity progression---models may excel at perception yet fail at spatial computation or multi-constraint judgment.

\noindent\textbf{Multi-Condition Rendering.}
Contextual samples are evenly distributed across three lighting modes and six scale levels. \textit{Lighting modes} include day (bright backgrounds), dusk (reduced contrast), and night (dark backgrounds, high-contrast symbols). \textit{Scale levels} span large scales (1:50k, 1:70k) with maximum detail, intermediate scales (1:100k, 1:130k) with moderate density, and small scales (1:200k, 1:300k) with cartographic generalization. This systematic variation tests model robustness across conditions mariners encounter when adjusting display settings.
\begin{table*}[t]
\centering
\renewcommand{\arraystretch}{0.9}
\caption{Performance on Perception and Decision-Making tasks. Results in accuracy (\%). \textbf{Bold} indicates best performance.}
\label{tab:main_results}
\footnotesize
\begin{tabular}{l|cccc|ccc|c}
\toprule
& \multicolumn{4}{c|}{\textbf{Perception}} & \multicolumn{3}{c|}{\textbf{Decision-Making}} & \\
\textbf{Model} & Symbol & Point & Line & Polygon & Track & Safety & Anchor & \textbf{Average} \\
\midrule
Gemini-2.5-Pro~\cite{comanici2025gemini} & \textbf{69.53} & 45.38 & \textbf{30.05} & \textbf{39.95} & 63.12 & 57.55 & 29.55 & \textbf{47.88} \\
Gemini-2.5-Flash~\cite{comanici2025gemini} & 63.80 & 41.89 & 29.08 & 39.44 & 59.57 & 55.84 & 24.35 & 44.85 \\
GPT-4o~\cite{hurst2024gpt} & 50.78 & 36.39 & 21.58 & 23.97 & 45.39 & 45.44 & 20.57 & 34.87 \\
\midrule
Qwen3-VL-235B-Instruct~\cite{qwen3technicalreport} & 57.03 & \textbf{51.79} & 29.70 & 29.93 & 74.47 & 58.97 & 26.48 & 46.91 \\
Qwen3-VL-235B-Thinking~\cite{qwen3technicalreport} & 54.17 & 49.34 & 28.18 & 27.83 & \textbf{75.18} & 61.54 & \textbf{30.50} & 46.68 \\
Qwen3-VL-32B-Instruct~\cite{qwen3technicalreport} & 50.26 & 48.77 & 29.39 & 26.86 & 68.79 & 58.40 & 17.26 & 42.82 \\
Qwen3-VL-32B-Thinking~\cite{qwen3technicalreport} & 46.88 & 35.93 & 28.45 & 24.67 & 73.76 & 55.98 & 18.44 & 40.59 \\
GLM-4.5V~\cite{glm45v} & 38.80 & 43.44 & 20.92 & 21.61 & 53.19 & \textbf{65.67} & 26.24 & 38.55 \\
InternVL-3-38B~\cite{wang2025internvl3} & 55.99 & 27.36 & 19.87 & 30.29 & 51.06 & 54.13 & 20.57 & 37.04 \\
Llama-4-Maverick-17B~\cite{meta2025llama} & 47.66 & 42.75 & 19.83 & 18.54 & 53.90 & 57.55 & 14.89 & 36.44 \\
\midrule
Random Chance & 25.00 & 25.00 & 25.00 & 25.00 & 33.33 & 50.00 & 25.00 & 29.76 \\
\bottomrule
\end{tabular}
\end{table*}

\begin{table*}[t]
\centering
\caption{Performance on Spatial Reasoning tasks. Results in Acc@T (\%) and Mean Error. \textbf{Bold} indicates best performance.}
\label{tab:spatial}
\resizebox{\textwidth}{!}{
\begin{tabular}{l|cc|cc|cc|cc}
\toprule
& \multicolumn{2}{c|}{\textbf{Coordinate (Geo)}} & \multicolumn{2}{c|}{\textbf{Coordinate (Pixel)}} & \multicolumn{2}{c|}{\textbf{Direction}} & \multicolumn{2}{c}{\textbf{Distance}} \\
\textbf{Model} & Acc@200px (\%) $\uparrow$ & Err (px) $\downarrow$ & Acc@200px (\%) $\uparrow$ & Err (px) $\downarrow$ & Acc@20° (\%) $\uparrow$ & Err (°) $\downarrow$ & Acc@0.2 (\%) $\uparrow$ & Err (\%) $\downarrow$ \\
\midrule
Gemini-2.5-Pro~\cite{comanici2025gemini} & \textbf{17.36} & \textbf{495.2} & \textbf{21.43} & 480.5 & 46.86 & \textbf{29.95} & 25.67 & \textbf{42.31} \\
Gemini-2.5-Flash~\cite{comanici2025gemini} & 16.77 & 502.1 & 21.03 & \textbf{466.0} & 28.77 & 66.89 & \textbf{25.93} & 42.72 \\
GPT-4o~\cite{hurst2024gpt} & 12.20 & 540.5 & 6.94 & 552.1 & 25.15 & 73.88 & 19.29 & 48.78 \\
\midrule
Qwen3-VL-235B-Instruct~\cite{qwen3technicalreport} & 11.31 & 570.4 & 12.20 & 510.9 & 36.95 & 33.53 & 16.71 & 56.36 \\
Qwen3-VL-235B-Thinking~\cite{qwen3technicalreport} & 11.01 & 555.2 & 12.90 & 501.7 & \textbf{55.64} & 34.15 & 20.33 & 51.02 \\
Qwen3-VL-32B-Instruct~\cite{qwen3technicalreport} & 5.26 & 630.3 & 5.16 & 635.3 & 26.44 & 52.11 & 12.75 & 48.70 \\
Qwen3-VL-32B-Thinking~\cite{qwen3technicalreport} & 6.94 & 615.1 & 7.34 & 610.2 & 44.10 & 52.08 & 4.91 & 58.20 \\
GLM-4.5V~\cite{glm45v} & 9.52 & 580.6 & 14.38 & 497.9 & 28.60 & 32.22 & 9.04 & 63.03 \\
InternVL-3-38B~\cite{wang2025internvl3} & 13.19 & 512.7 & 10.52 & 565.1 & 41.09 & 54.80 & 13.09 & 47.95 \\
Llama-4-Maverick-17B~\cite{meta2025llama} & 9.13 & 590.3 & 12.90 & 574.4 & 14.90 & 74.22 & 7.41 & 68.99 \\
\bottomrule
\end{tabular}
}
\end{table*}

\section{Experiments}

\subsection{Experimental Setup}

\noindent\textbf{Evaluated Models.} 
We evaluate ten state-of-the-art MLLMs spanning diverse architectures and parameter scales, categorized into two types: 1) \textit{Closed-source models} include GPT-4o~\cite{hurst2024gpt}, Gemini 2.5 Pro, and Gemini 2.5 Flash~\cite{comanici2025gemini}, representing commercial frontier systems; 2) \textit{Open-source models} comprise Qwen3-VL-235B and Qwen3-VL-32B~\cite{qwen3technicalreport} (each available in both Instruct and Thinking variants), InternVL-3-38B~\cite{wang2025internvl3}, GLM-4.5V~\cite{glm45v}, and Llama-4-Maverick-17B-128E~\cite{meta2025llama}. This selection spans 17B to 235B parameters, enabling analysis across model sizes, architectural paradigms (dense vs. MoE), and reasoning strategies (instruct vs. thinking).

\noindent\textbf{Evaluation Metrics.}
Model performance is evaluated under zero-shot setting with uniform prompts across all models. Perception and Decision-Making tasks use multiple-choice format with chart images and natural language questions, while Spatial Reasoning tasks require numeric outputs with explicit unit specifications. Closed-source models are accessed via official APIs; open-source models via HuggingFace Transformers. Perception and Decision-Making tasks use \textbf{Accuracy (\%)}. Spatial Reasoning tasks employ \textbf{Accuracy at Tolerance (Acc@T)}—predictions within error thresholds—and \textbf{Mean Error} in task-specific units. Coordinate localization evaluates geographic coordinates (latitude/longitude converted to pixels) and direct pixel prediction, both with 200px tolerance. Bearing and distance use 20° and 20\% relative error thresholds. Full prompts and additional results are in the Appendix.

\subsection{Main Results}
\label{sec:results}

We present evaluation results across the three tiers. Table~\ref{tab:main_results} reports performance on Perception and Decision-Making tasks, while Table~\ref{tab:spatial} presents Spatial Reasoning results.

\noindent\textbf{Results on Perception Tasks.}
Perception tasks reveal a substantial performance gap between isolated and contextual understanding. Symbol Recognition on isolated symbols proves most accessible, with Gemini-2.5-Pro achieving approximately 70\% accuracy. However, when these same symbols appear embedded in complete chart scenes, performance degrades dramatically—contextual feature recognition drops to a range of 30\%-52\% depending on feature type and visual complexity. This degradation demonstrates that overlapping layers, dense bathymetry, and competing visual elements fundamentally challenge current vision architectures trained predominantly on natural images. All models substantially exceed random baseline performance, yet the best average across perception tasks reaches only 48\%, falling far short of professional maritime requirements where near-perfect accuracy is expected.

\noindent\textbf{Results on Spatial Reasoning Tasks.}
Spatial Reasoning exposes fundamental limitations in visually-grounded numerical computation (Table~\ref{tab:spatial}). Coordinate localization reveals a counterintuitive pattern: models achieve better accuracy through direct pixel prediction than through geographic coordinate conversion, despite the latter being theoretically more precise for cartographic localization. This pattern reveals a critical bottleneck—models struggle more with interpreting formal coordinate notation systems than with visual feature localization itself. The geographic approach requires reading grid annotations, interpolation, and unit conversion, with errors cascading at each step. Direction calculation shows that extended reasoning variants substantially outperform standard models, while distance measurement proves universally challenging with best performance around 26\% accuracy and mean relative errors exceeding 40\%. The consistently poor results suggest fundamental architectural limitations in maintaining precision through multi-step numerical reasoning chains.

\definecolor{BadChange}{rgb}{0.8, 0, 0}
\definecolor{GoodChange}{rgb}{0, 0.5, 0}

\begin{table}[t]
\centering
\small
\caption{Average model performance across lighting conditions. Delta shows change relative to day mode baseline. \textcolor{GoodChange}{green} indicates improvement, \textcolor{BadChange}{Red} indicates degradation.}
\label{tab:lighting}
\resizebox{\columnwidth}{!}{
\begin{tabular}{lccc}
\toprule
\textbf{Task Category} & \textbf{Day} & \textbf{Dusk} & \textbf{Night} \\
\midrule
\multicolumn{4}{c}{\textit{Perception \& Decision-Making (Accuracy \% $\uparrow$)}} \\
\cmidrule{1-4}
Point Features & \textbf{42.88} & 41.76 (\textcolor{BadChange}{-1.12}) & 42.27 (\textcolor{BadChange}{-0.61}) \\
Linestring Features & \textbf{27.01} & 24.86 (\textcolor{BadChange}{-2.15}) & 25.26 (\textcolor{BadChange}{-1.75}) \\
Polygon Features & 27.32 & \textbf{29.92} (\textcolor{GoodChange}{+2.60}) & 27.70 (\textcolor{GoodChange}{+0.38}) \\
Track Direction & \textbf{65.35} & 61.27 (\textcolor{BadChange}{-4.08}) & 58.92 (\textcolor{BadChange}{-6.43}) \\
Safety Assessment & \textbf{57.80} & 55.99 (\textcolor{BadChange}{-1.81}) & 57.52 (\textcolor{BadChange}{-0.28}) \\
Anchorage Selection & \textbf{24.63} & 22.18 (\textcolor{BadChange}{-2.45}) & 21.85 (\textcolor{BadChange}{-2.78}) \\
\midrule
\multicolumn{4}{c}{\textit{Spatial Reasoning (Mean Error $\downarrow$)}} \\
\cmidrule{1-4}
Coordinate (Geo) Error (px) & 592.4 & 587.6 (\textcolor{GoodChange}{-4.8}) & \textbf{490.5} (\textcolor{GoodChange}{-101.9}) \\
Coordinate (Pixel) Error (px) & 558.5 & 539.1 (\textcolor{GoodChange}{-19.4}) & \textbf{520.6} (\textcolor{GoodChange}{-37.9}) \\
Direction Error (°) & 55.11 & 51.26 (\textcolor{GoodChange}{-3.85}) & \textbf{44.78} (\textcolor{GoodChange}{-10.33}) \\
Distance Error (\%) & 53.01 & \textbf{52.46} (\textcolor{GoodChange}{-0.55}) & 52.94 (\textcolor{GoodChange}{-0.07}) \\
\bottomrule
\end{tabular}
}
\end{table}

\noindent\textbf{Results on Decision-Making Tasks.}
Decision-Making tasks reveal divergent difficulty levels that directly correlate with constraint complexity. Simpler tasks with constrained decision spaces achieve moderate success, reaching approximately 65\%-75\% accuracy. In stark contrast, Anchorage Selection—requiring simultaneous optimization of distance, depth, and regulatory constraints—proves catastrophic, with best performance barely exceeding random baseline. Extended reasoning mechanisms consistently improve performance, with gains ranging from minimal on simple tasks to substantial on complex ones. However, these improvements prove insufficient for operational deployment. The results expose a fundamental architectural limitation: current models lack explicit mechanisms for constraint enumeration, independent verification, and multi-objective synthesis required in safety-critical professional navigation scenarios.


\begin{table}[t]
\centering
\caption{Average model performance across scale levels. Delta shows change relative to large scale baseline. \textcolor{GoodChange}{green} indicates improvement, \textcolor{BadChange}{Red} indicates degradation.}
\label{tab:scale}
\resizebox{\columnwidth}{!}{
\begin{tabular}{lccc}
\toprule
\textbf{Task Category} & \textbf{Large Scale} & \textbf{Intermediate} & \textbf{Small Scale} \\
& \textbf{(1:50k, 1:70k)} & \textbf{(1:100k, 1:130k)} & \textbf{(1:200k, 1:300k)} \\
\midrule
\multicolumn{4}{c}{\textit{Perception Tasks (Accuracy \% $\uparrow$)}} \\
\cmidrule{1-4}
Point Features & 42.24 & \textbf{47.72} (\textcolor{GoodChange}{+5.48}) & 38.19 (\textcolor{BadChange}{-4.05}) \\
Linestring Features & \textbf{28.83} & 20.23 (\textcolor{BadChange}{-8.60}) & 20.26 (\textcolor{BadChange}{-8.57}) \\
Polygon Features & \textbf{28.93} & 26.73 (\textcolor{BadChange}{-2.20}) & 25.37 (\textcolor{BadChange}{-3.56}) \\
\midrule
\multicolumn{4}{c}{\textit{Decision-Making Tasks (Accuracy \% $\uparrow$)}} \\
\cmidrule{1-4}
Track Direction & \textbf{68.16} & 58.33 (\textcolor{BadChange}{-9.83}) & 52.30 (\textcolor{BadChange}{-15.86}) \\
Safety Assessment & 51.19 & \textbf{59.72} (\textcolor{GoodChange}{+8.53}) & 48.96 (\textcolor{BadChange}{-2.23}) \\
Anchorage Selection & 16.47 & \textbf{20.34} (\textcolor{GoodChange}{+3.87}) & 11.22 (\textcolor{BadChange}{-5.25}) \\
\bottomrule
\end{tabular}
}
\end{table}

\begin{figure*}[!t]
\centering
\includegraphics[width=\textwidth]{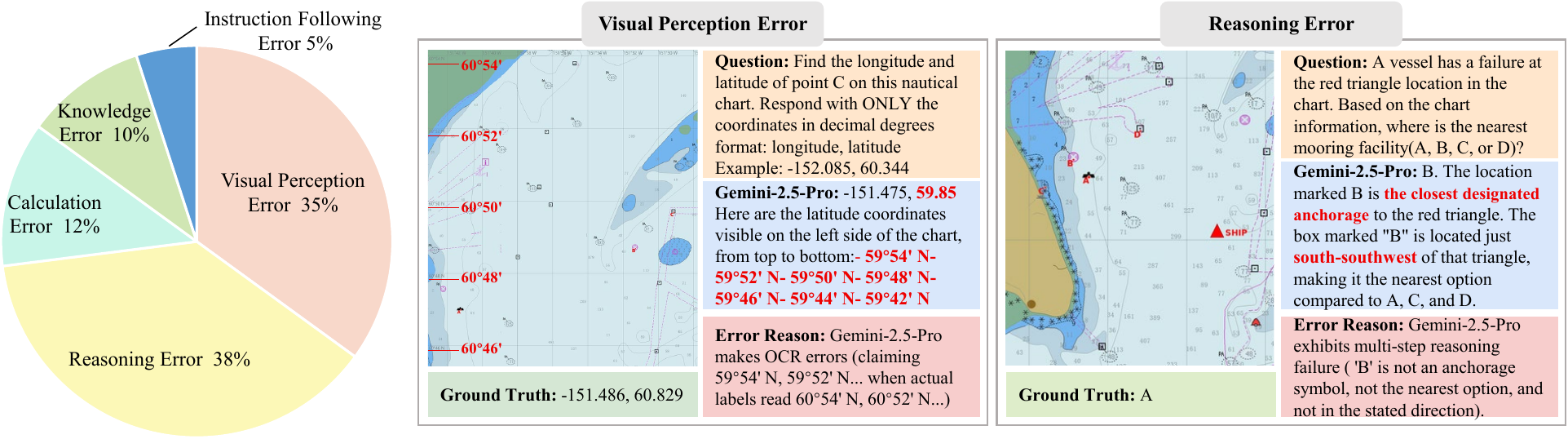}
\caption{Error distribution of Gemini-2.5-Pro’s incorrect results and example errors in its responses.}
\label{fig:e}
\end{figure*}

\subsection{Ablation Studies}
\label{sec:ablation}

We analyze model performance robustness across two critical rendering conditions: lighting modes and scale levels. These dimensions directly impact operational ECDIS usage as mariners regularly adjust display settings based on ambient conditions and navigation phases.

\noindent\textbf{Performance Across Lighting Modes.}
Table~\ref{tab:lighting} presents average performance across three lighting modes, revealing divergent task-specific dependencies on color information. Perception and Decision-Making tasks favor day mode, degrading up to 6\% under night mode as color-dependent features lose discriminability. In stark contrast, Spatial Reasoning tasks paradoxically improve under night mode rendering. Geographic coordinate localization shows dramatic improvement, substantially exceeding direct pixel localization gains. This counterintuitive pattern reveals an underlying trade-off: high-contrast monochromatic rendering enhances symbolic notation reading by reducing visual ambiguity, while simultaneously impairing semantic feature classification that relies on color discrimination. Direction calculation shows substantial improvements, while distance measurement demonstrates minimal sensitivity.

\noindent\textbf{Performance Across Scale Levels.}
Table~\ref{tab:scale} analyzes performance across large, intermediate, and small scales, exposing adaptation failures to cartographic generalization. Perception tasks exhibit complex non-monotonic patterns as scale decreases: linear and polygon features degrade consistently with each scale reduction, while point features paradoxically improve at intermediate scales before declining. This demonstrates models cannot adjust interpretation strategies as rendering simplifies at smaller scales—a capability humans employ intuitively. Decision-Making shows greater instability, with Track Direction suffering severe degradation (approaching 16\%), while other tasks fluctuate unpredictably. Overall, Perception degrades progressively while Decision-Making maintains stability at intermediate scales before collapsing. These variations fundamentally undermine reliability for operational use, where mariners routinely adjust zoom levels.

\subsection{Error Analysis and Discussions}

\noindent\textbf{Error Analysis.} 
We conduct error analysis on 200 randomly sampled incorrect predictions from Gemini-2.5-Pro, as visualized in Figure~\ref{fig:e}. Reasoning errors and visual perception errors dominate failures, accounting for 38\% and 35\% of total errors respectively. Reasoning errors primarily manifest as multi-step failures in spatial judgment and constraint integration, where models correctly perceive individual elements but fail to reason about their relationships. For instance, in anchorage selection tasks, models misidentify feature types, misjudge spatial proximity, and provide incorrect directional descriptions—failing across multiple reasoning dimensions simultaneously. Visual perception errors commonly involve OCR failures on coordinate grid annotations: models consistently misread latitude values by approximately one degree, directly explaining why geographic coordinate prediction underperforms direct pixel localization. Calculation errors occur when numerical computations fail despite correct visual perception, while knowledge errors and instruction-following issues represent secondary failure modes likely stemming from insufficient domain-specific pretraining.

\noindent\textbf{Findings and Discussions.} 
We discuss key findings from our experiments to inspire future work: \textbf{1)} Current MLLMs fall significantly short of professional maritime requirements, with best performance reaching only 47.88\% compared to near-perfect accuracy expected in operational settings. The gap between perception and decision-making capabilities indicates that symbolic recognition alone is insufficient—structured reasoning remains the critical bottleneck. \textbf{2)} Geographic coordinate prediction consistently underperforms direct pixel localization despite being theoretically more precise, revealing that models struggle more with formal notation systems than with visual feature detection. This exposes a fundamental limitation in symbolic-to-spatial translation. \textbf{3)} Extended reasoning mechanisms substantially improve performance on decomposable geometric tasks but fail at multi-constraint optimization. Thinking modes excel at procedural decomposition yet lack explicit constraint verification mechanisms for complex decisions. \textbf{4)} Models show limited robustness across rendering variations. Non-monotonic patterns across scale levels and paradoxical improvements under night mode demonstrate inability to adjust interpretation strategies—a capability humans employ intuitively. \textbf{5)} Multi-constraint decision-making reveals architectural limitations of current models. Tasks requiring simultaneous optimization of distance, depth, and regulatory compliance demand structured reasoning capabilities absent in existing models. These findings demonstrate that professional maritime applications require innovations in symbolic grounding, verifiable reasoning, and rendering-aware processing beyond merely scaling existing architectures.

\section{Conclusion}
\label{sec:conclusion}
We introduce ENC-Bench, the first comprehensive benchmark for evaluating multimodal large language models on Electronic Navigational Chart understanding. ENC-Bench addresses the gap between general visual understanding and professional maritime navigation through 20,490 expert-validated samples from 840 NOAA charts, organized into a three-tier framework spanning perception, spatial reasoning, and maritime decision-making. Evaluation of 10 state-of-the-art MLLMs reveals severe capability gaps, with the best model achieving only 47.88\% accuracy and catastrophic failures in multi-constraint reasoning (30.50\% on anchorage selection). Error analysis identifies fundamental limitations in symbolic notation interpretation, constraint satisfaction, and adaptive visual processing across rendering variations. Looking ahead, our benchmark methodology extends to other professional domains requiring standardized symbolic systems—aviation charts, infrastructure blueprints, and scientific visualization. We anticipate that ENC-Bench will inspire research in domain-specific visual understanding and benefit the development of deployable AI systems for highly specialized vertical domains.

\section*{Acknowledgments}
This work was supported by the National Natural Science Foundation of China under Grant 62306331 and CAAI Youth Talent Lifting Project under Grant CAAI2023-2025QNRC001.

{
    \small
    \bibliographystyle{ieeenat_fullname}
    \bibliography{main}
}

\clearpage
\clearpage
\setcounter{page}{1}
\maketitlesupplementary
\appendix

\section*{Overview of the Appendix}

\begin{figure*}[t]
    \centering
    \includegraphics[width=0.9\textwidth, keepaspectratio]{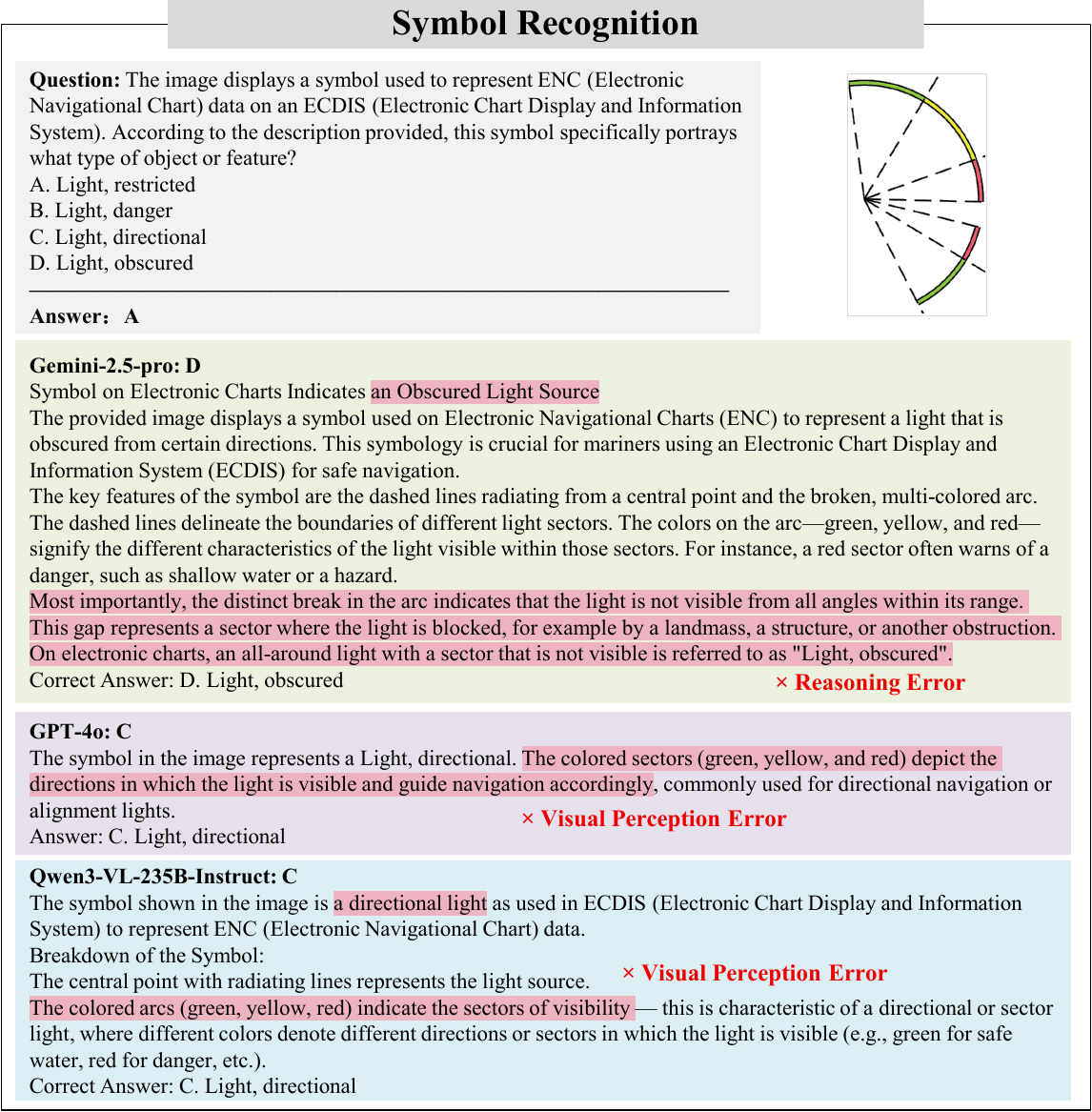}
    \caption{A sample case of \textbf{Symbol Recognition}. The model must identify an "Obscured Light" symbol.}
    \label{fig:case_1}
\end{figure*}

\begin{figure*}[t]
    \centering
    \includegraphics[width=0.9\textwidth, keepaspectratio]{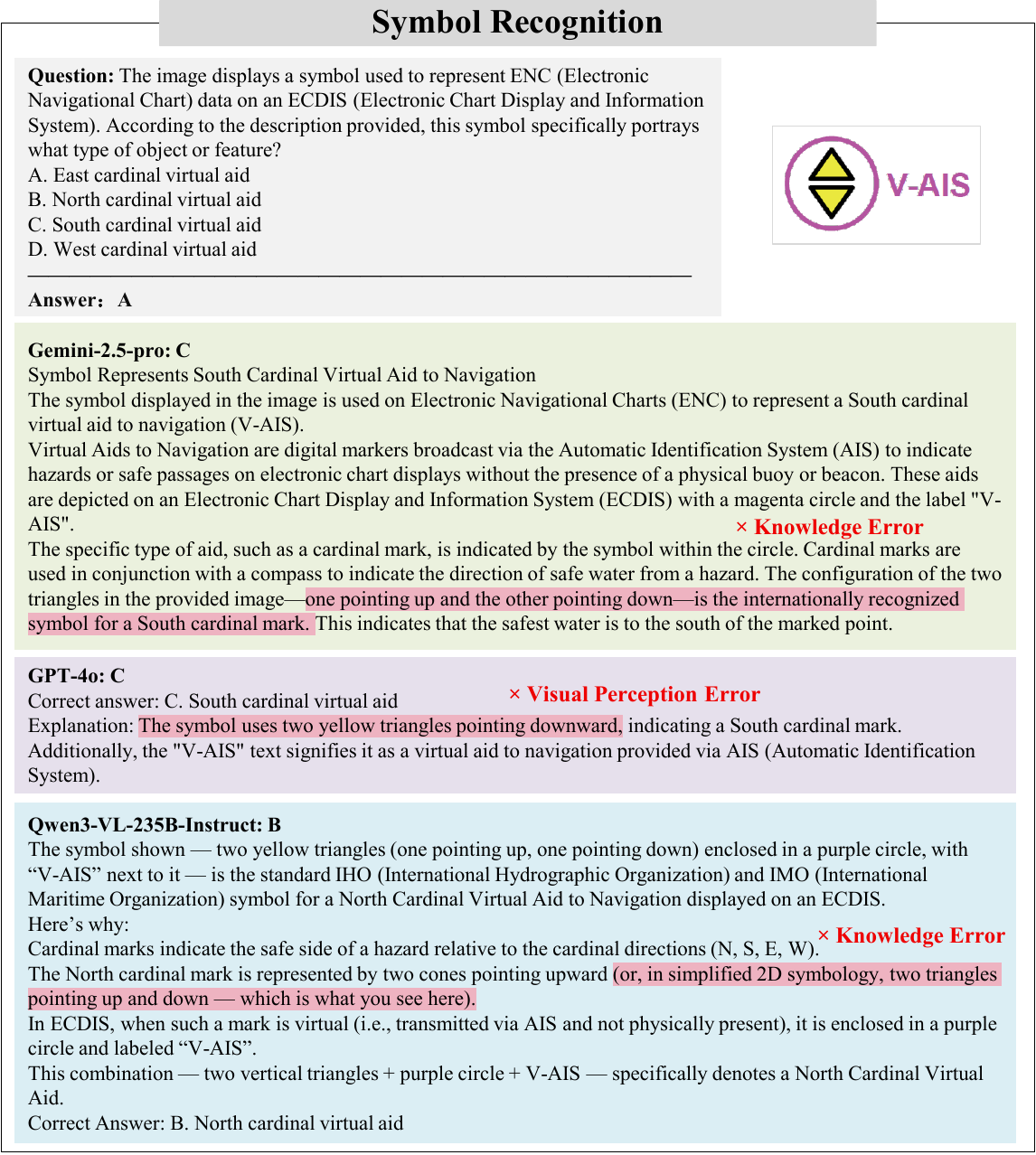}
    \caption{A sample case of \textbf{Symbol Recognition}. The model must distinguish between East and North cardinal marks.}
    \label{fig:case_2}
\end{figure*}

\begin{figure*}[t]
    \centering
    \includegraphics[width=0.9\textwidth, keepaspectratio]{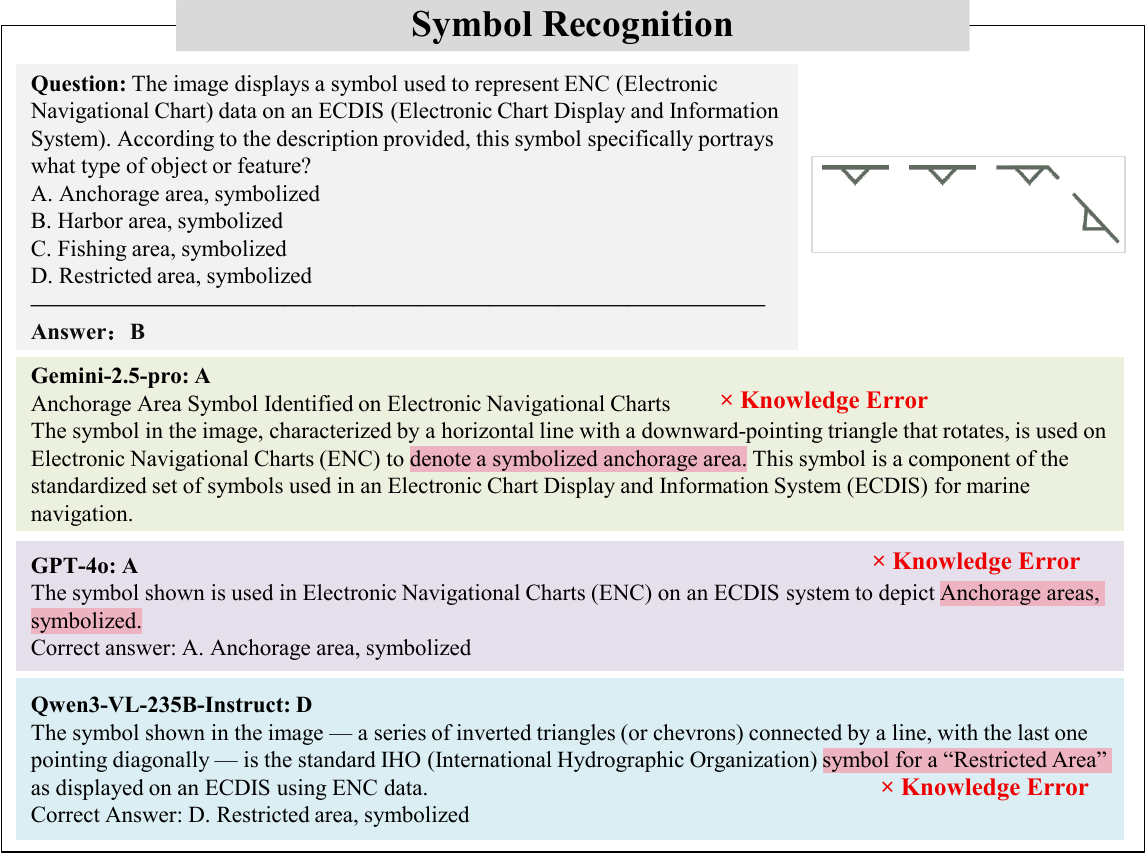}
    \caption{A sample case of \textbf{Symbol Recognition}. Identifying a symbolized Anchorage Area.}
    \label{fig:case_3}
\end{figure*}

\begin{figure*}[t]
    \centering
    \includegraphics[width=0.9\textwidth, keepaspectratio]{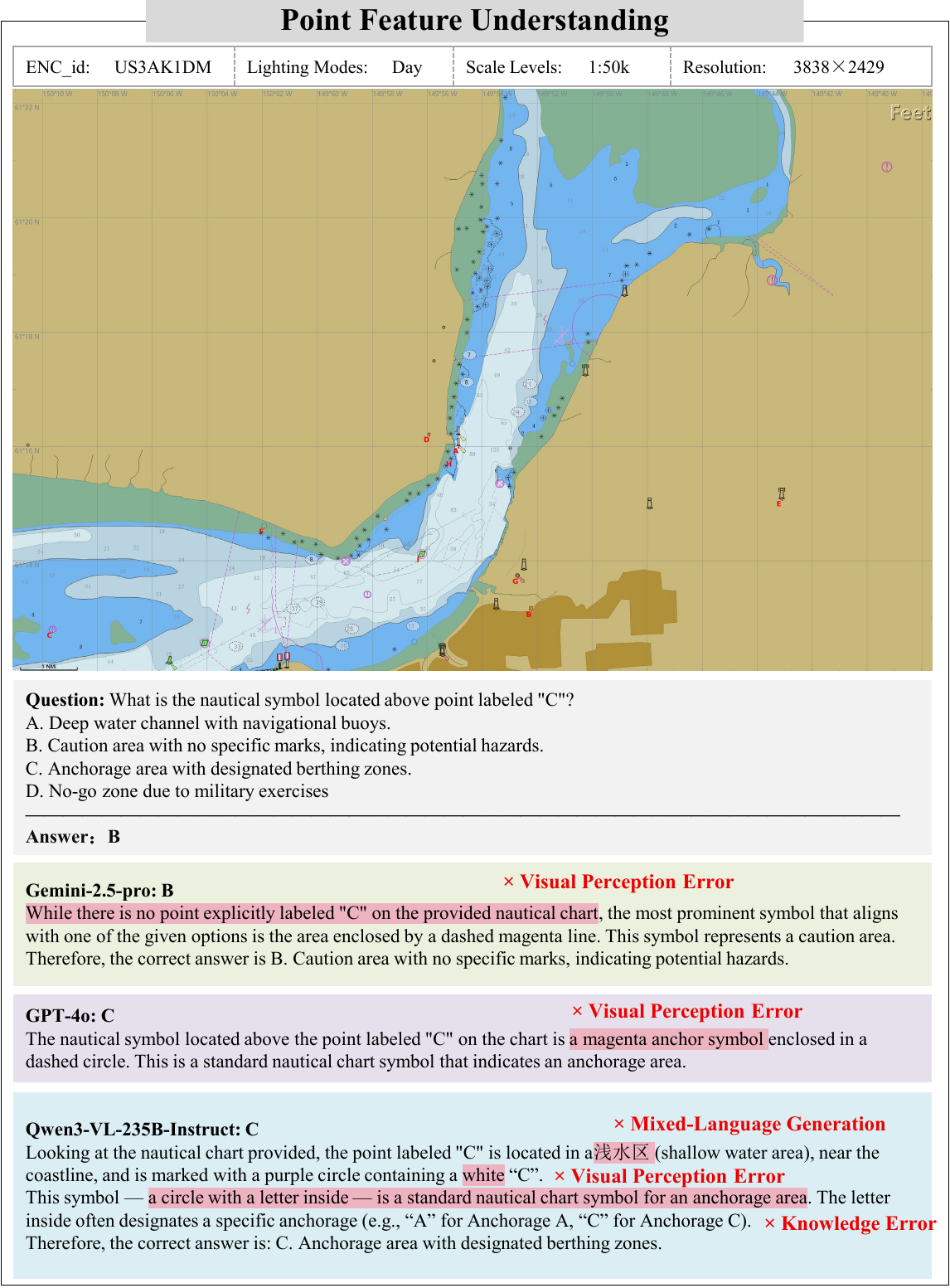}
    \caption{A sample case of \textbf{Point Feature Understanding} (Day Mode). Identifying a Caution Area symbol.}
    \label{fig:case_4}
\end{figure*}

\begin{figure*}[t]
    \centering
    \includegraphics[width=0.9\textwidth, keepaspectratio]{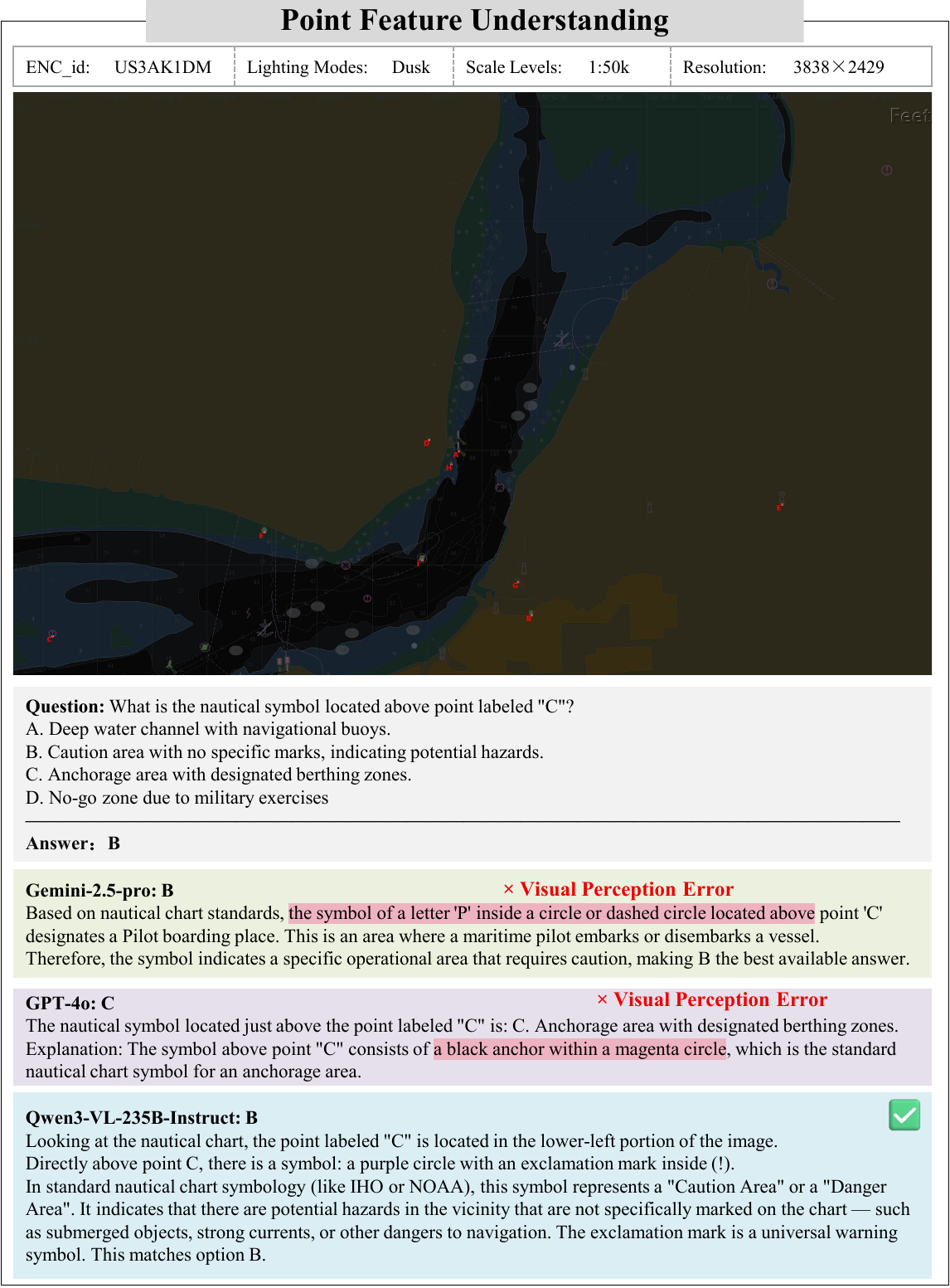}
    \caption{A sample case of \textbf{Point Feature Understanding} (Dusk Mode).}
    \label{fig:case_5}
\end{figure*}

\begin{figure*}[t]
    \centering
    \includegraphics[width=0.9\textwidth, keepaspectratio]{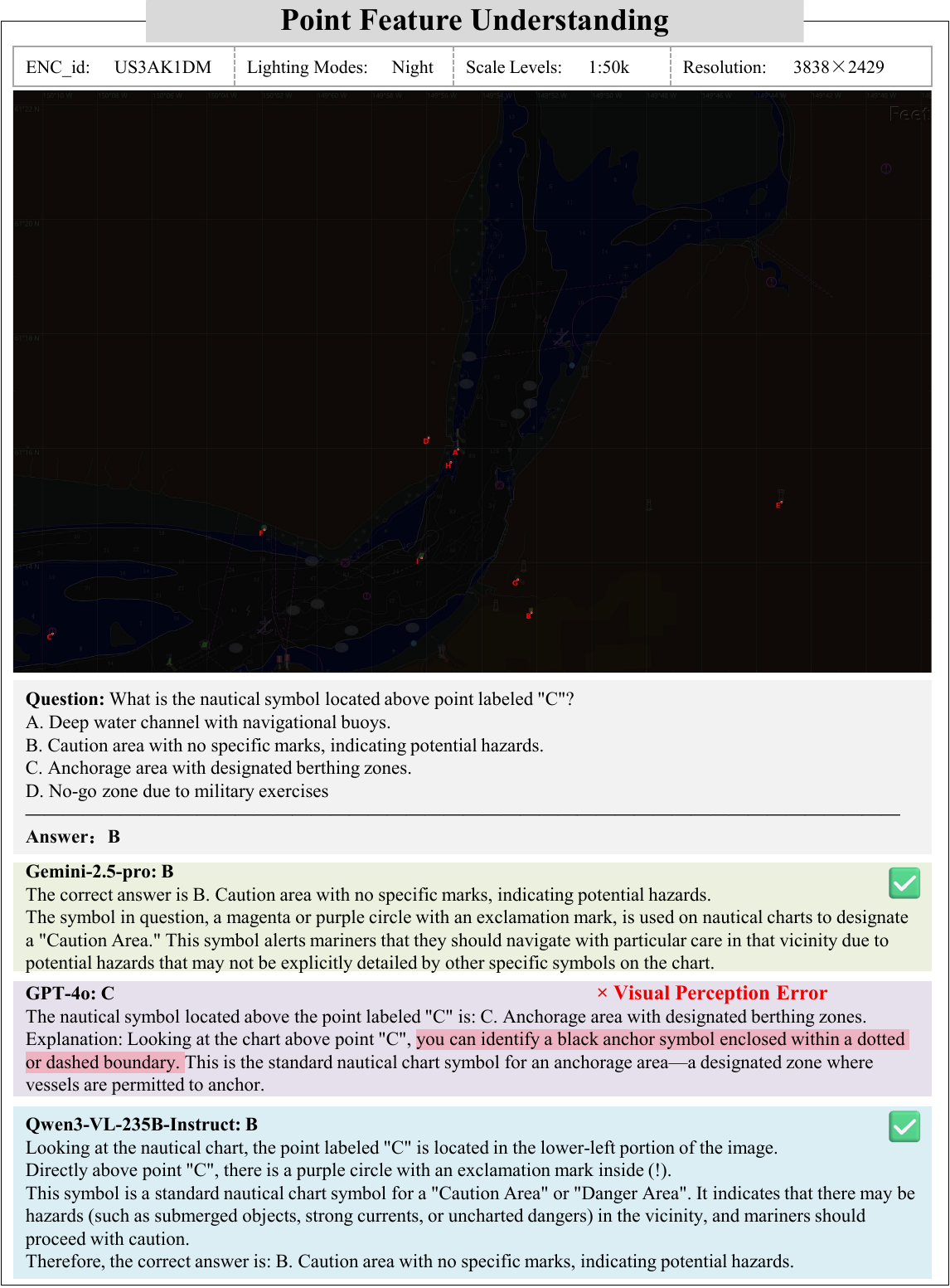}
    \caption{A sample case of \textbf{Point Feature Understanding} (Night Mode).}
    \label{fig:case_6}
\end{figure*}

\begin{figure*}[t]
    \centering
    \includegraphics[width=0.9\textwidth, keepaspectratio]{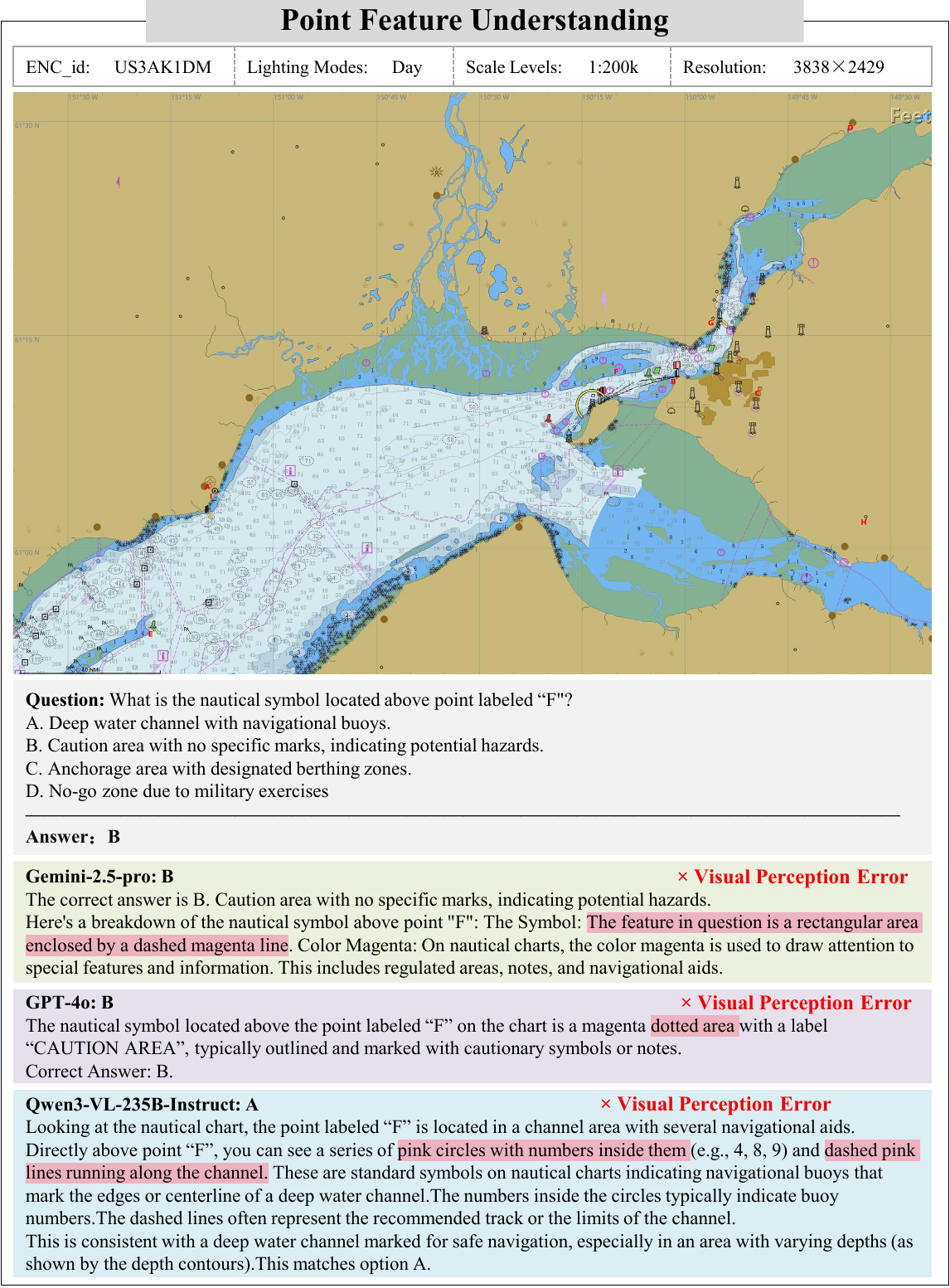}
    \caption{A sample case of \textbf{Point Feature Understanding} (Small Scale 1:200k).}
    \label{fig:case_7}
\end{figure*}

\begin{figure*}[t]
    \centering
    \includegraphics[width=0.9\textwidth, keepaspectratio]{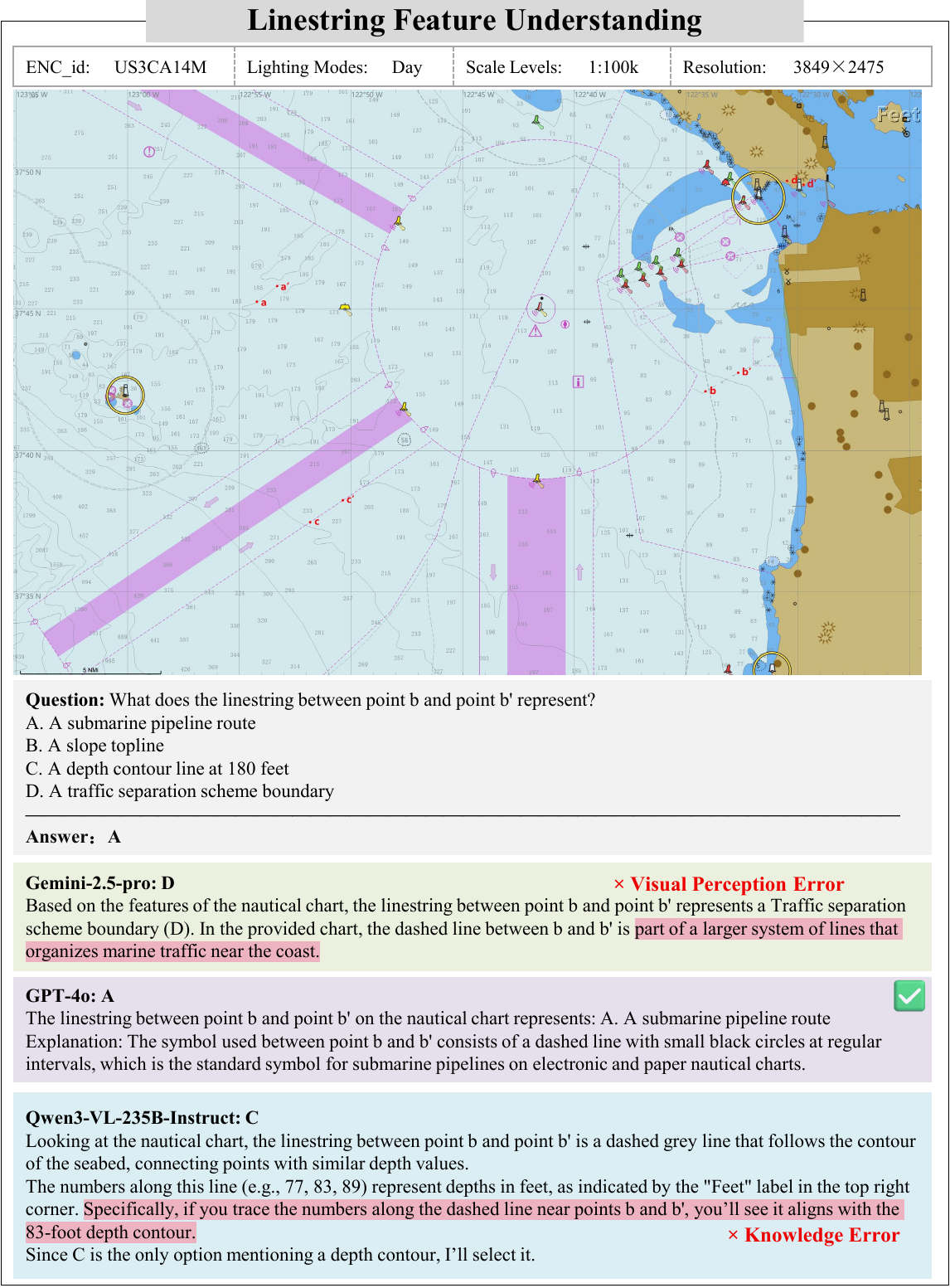}
    \caption{A sample case of \textbf{Linestring Feature Understanding} (Day Mode). Identifying a Submarine Pipeline.}
    \label{fig:case_8}
\end{figure*}

\begin{figure*}[t]
    \centering
    \includegraphics[width=0.9\textwidth, keepaspectratio]{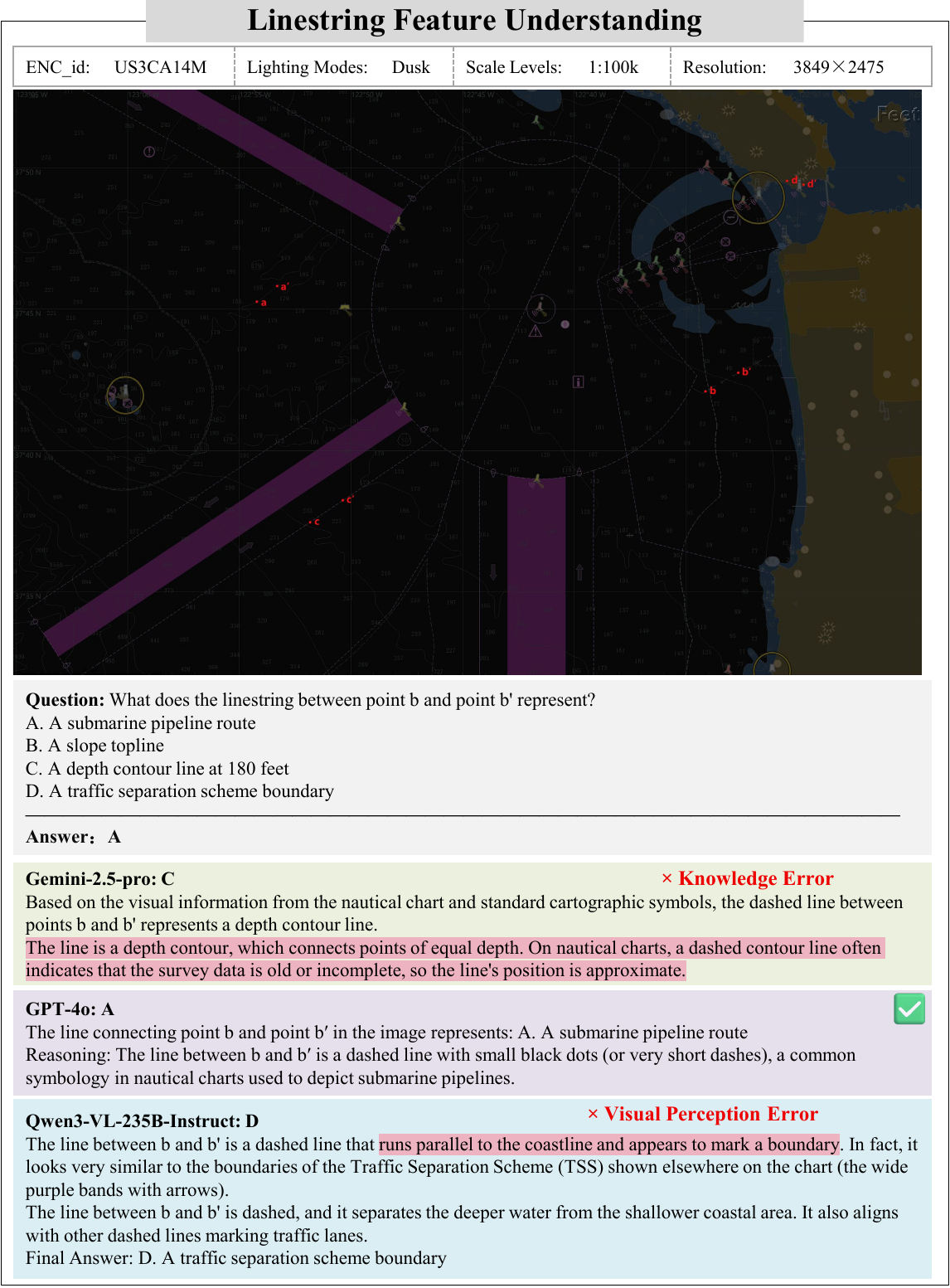}
    \caption{A sample case of \textbf{Linestring Feature Understanding} (Dusk Mode).}
    \label{fig:case_9}
\end{figure*}

\begin{figure*}[t]
    \centering
    \includegraphics[width=0.9\textwidth, keepaspectratio]{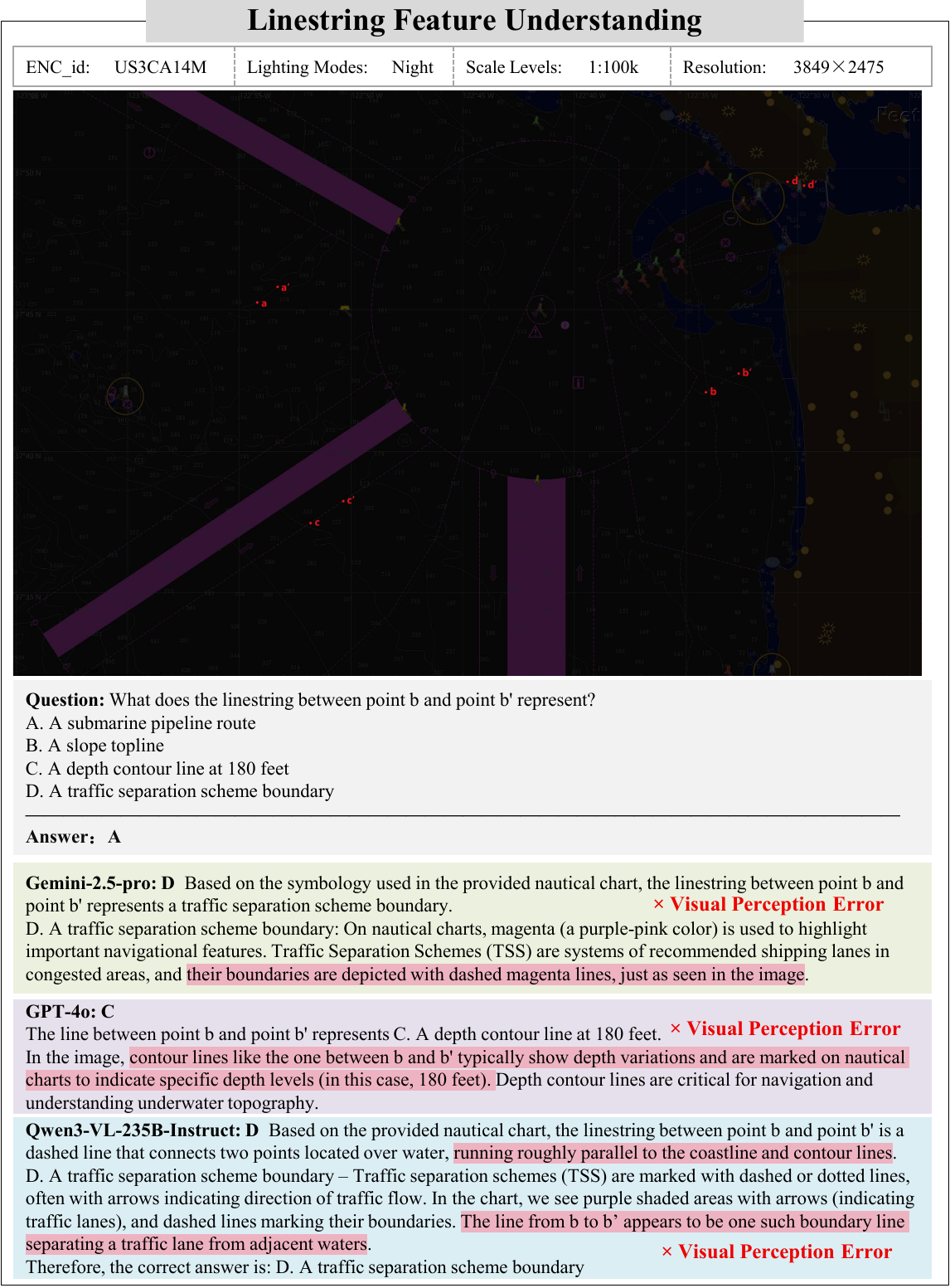}
    \caption{A sample case of \textbf{Linestring Feature Understanding} (Night Mode).}
    \label{fig:case_10}
\end{figure*}

\begin{figure*}[t]
    \centering
    \includegraphics[width=0.9\textwidth, keepaspectratio]{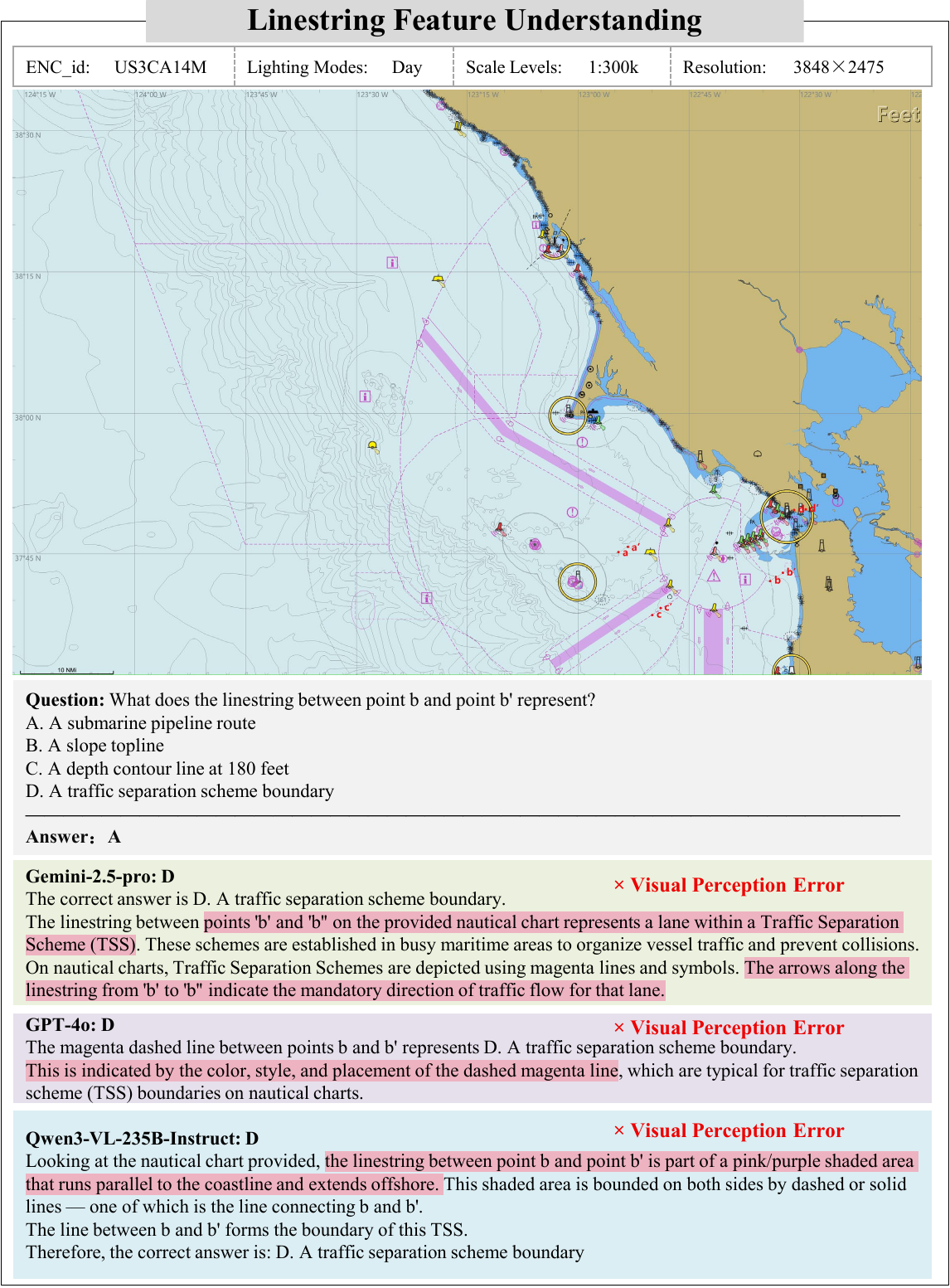}
    \caption{A sample case of \textbf{Linestring Feature Understanding} (Small Scale 1:300k). Identifying a Traffic Separation Scheme boundary.}
    \label{fig:case_11}
\end{figure*}

\begin{figure*}[t]
    \centering
    \includegraphics[width=0.9\textwidth, keepaspectratio]{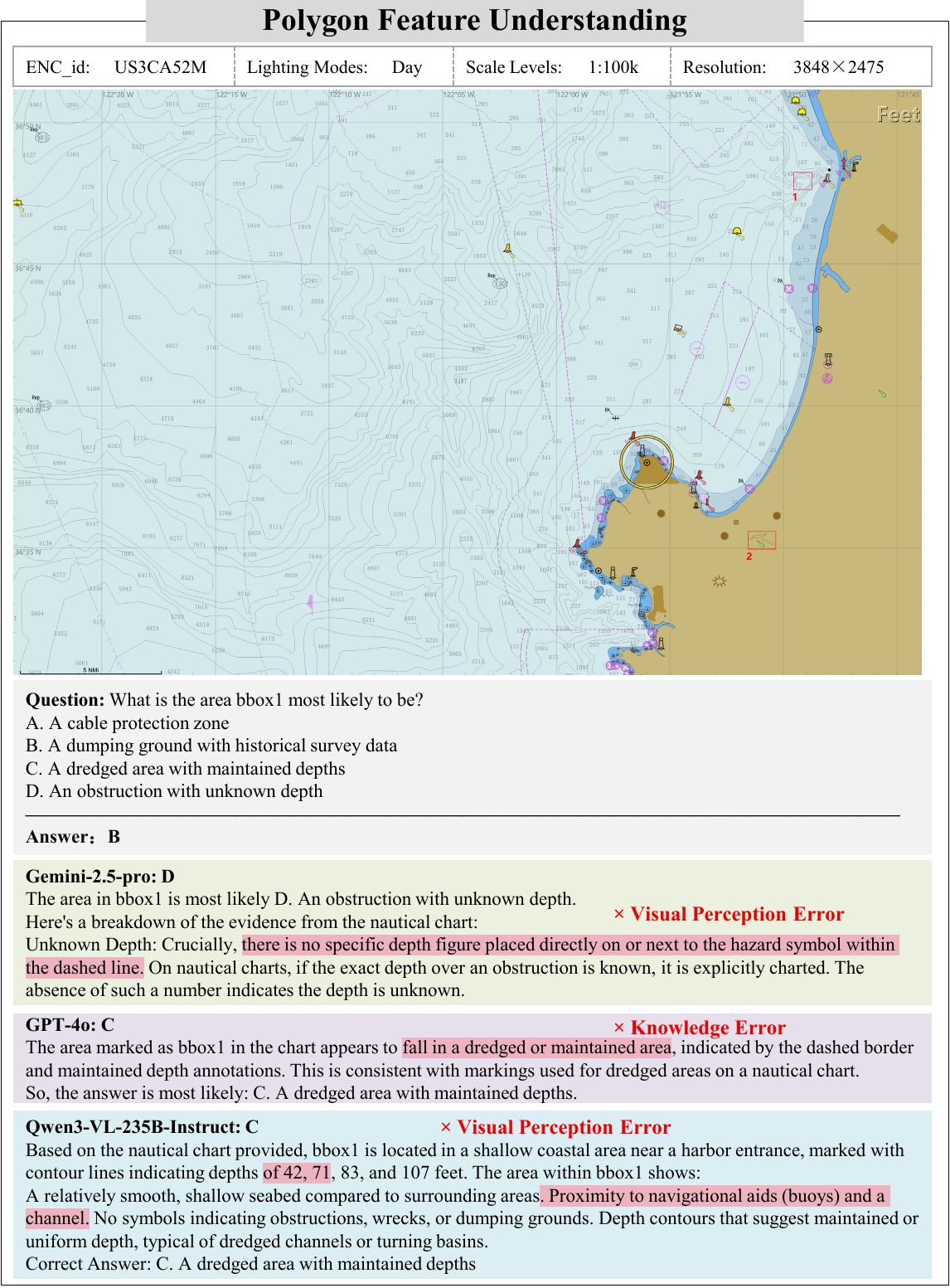}
    \caption{A sample case of \textbf{Polygon Feature Understanding} (Day Mode). Identifying a Dumping Ground.}
    \label{fig:case_12}
\end{figure*}

\begin{figure*}[t]
    \centering
    \includegraphics[width=0.9\textwidth, keepaspectratio]{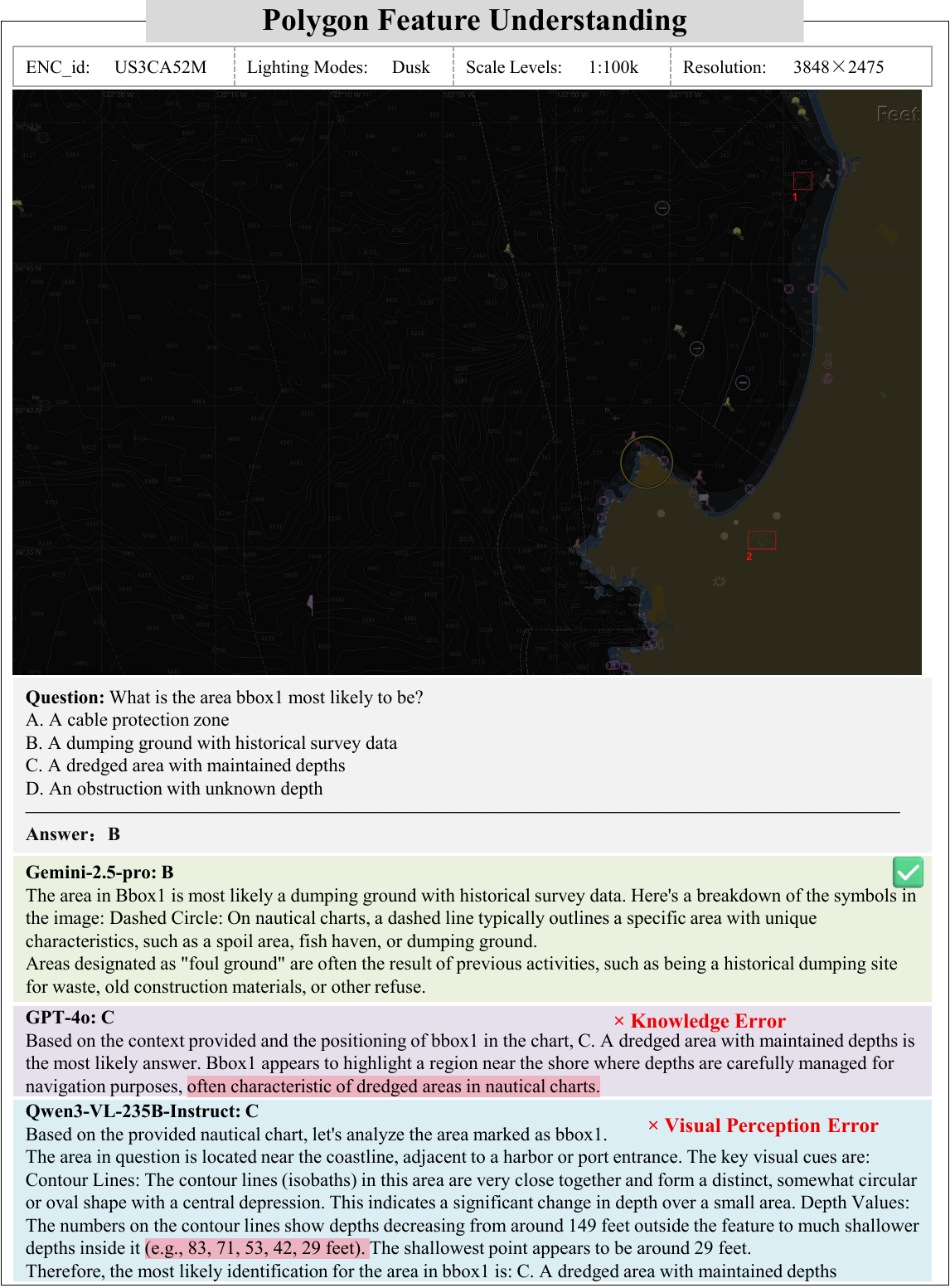}
    \caption{A sample case of \textbf{Polygon Feature Understanding} (Dusk Mode).}
    \label{fig:case_13}
\end{figure*}

\begin{figure*}[t]
    \centering
    \includegraphics[width=0.9\textwidth, keepaspectratio]{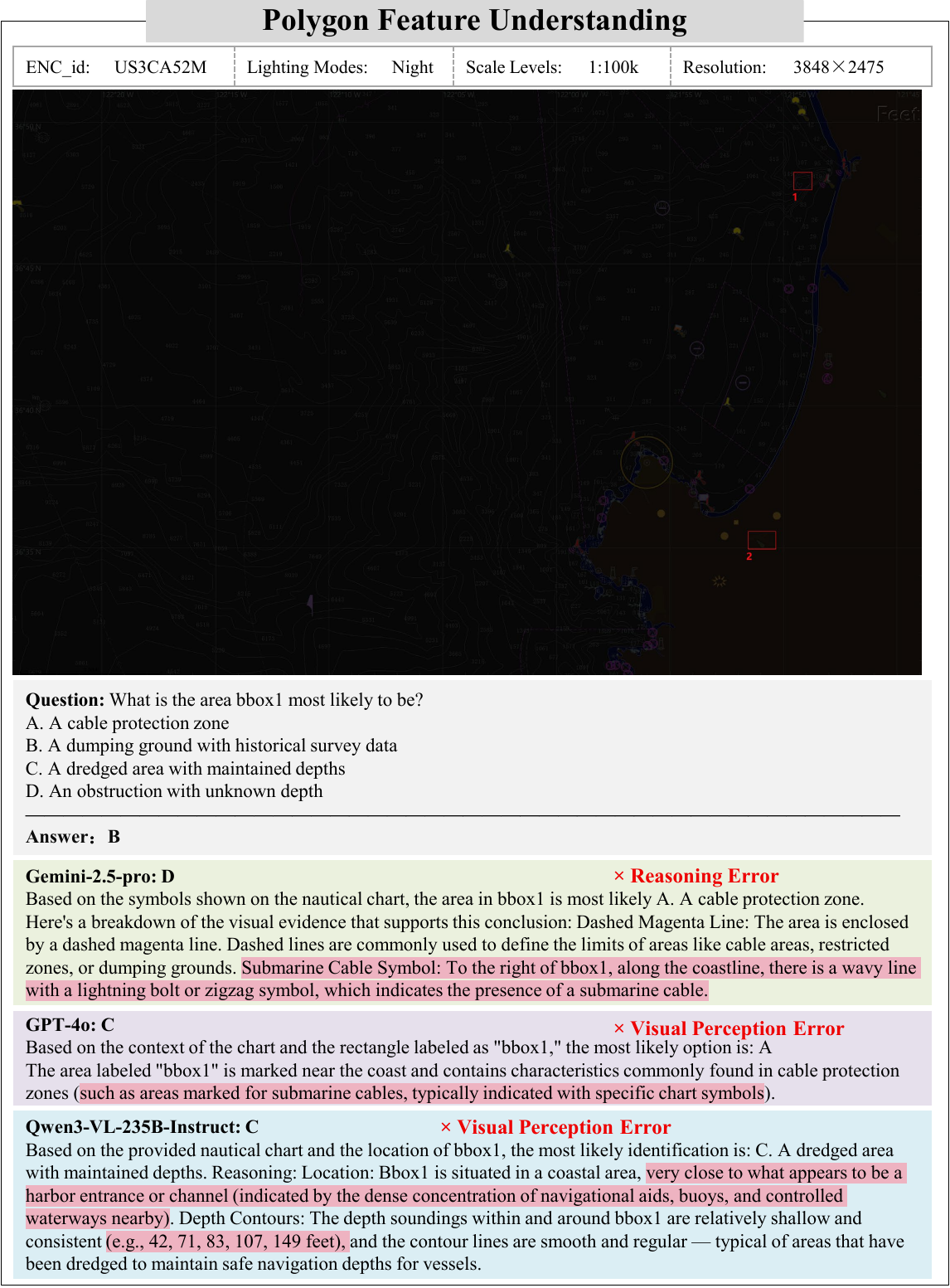}
    \caption{A sample case of \textbf{Polygon Feature Understanding} (Night Mode).}
    \label{fig:case_14}
\end{figure*}

\begin{figure*}[t]
    \centering
    \includegraphics[width=0.9\textwidth, keepaspectratio]{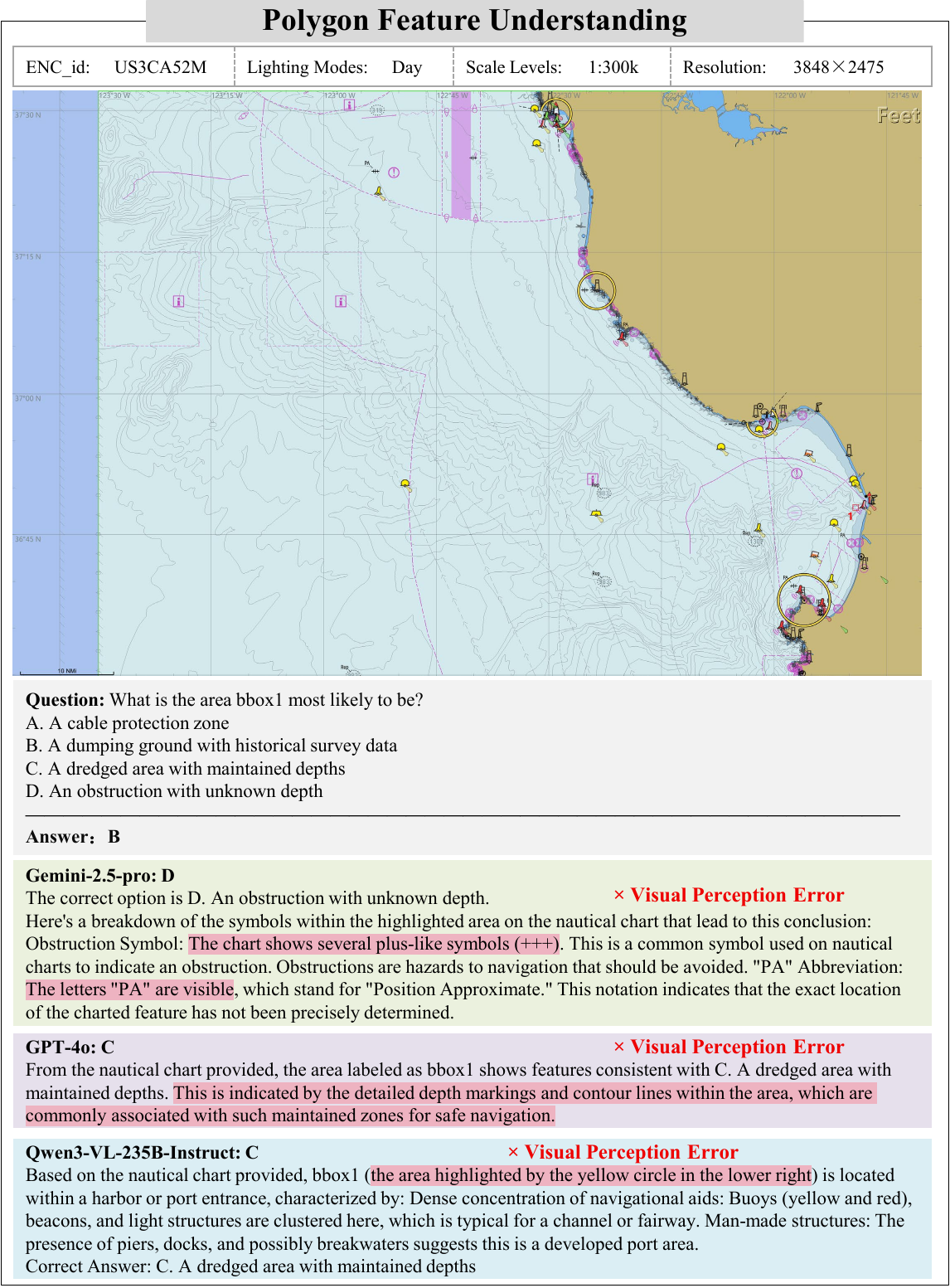}
    \caption{A sample case of \textbf{Polygon Feature Understanding} (Small Scale 1:300k).}
    \label{fig:case_15}
\end{figure*}

\begin{figure*}[t]
    \centering
    \includegraphics[width=0.9\textwidth, keepaspectratio]{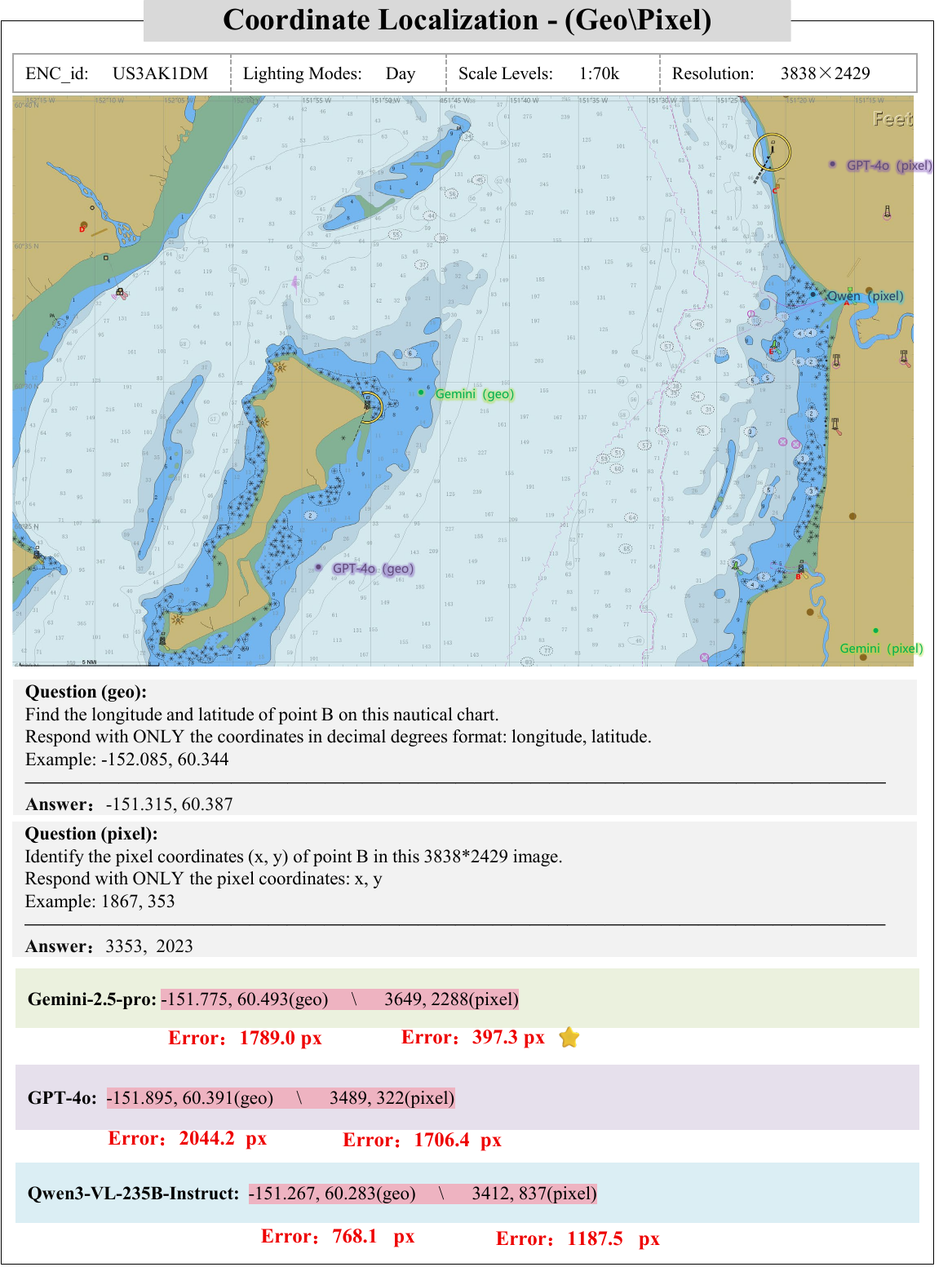}
    \caption{A sample case of \textbf{Coordinate Localization} (Day Mode). Comparison of Geographic vs. Pixel localization errors.}
    \label{fig:case_16}
\end{figure*}

\begin{figure*}[t]
    \centering
    \includegraphics[width=0.9\textwidth, keepaspectratio]{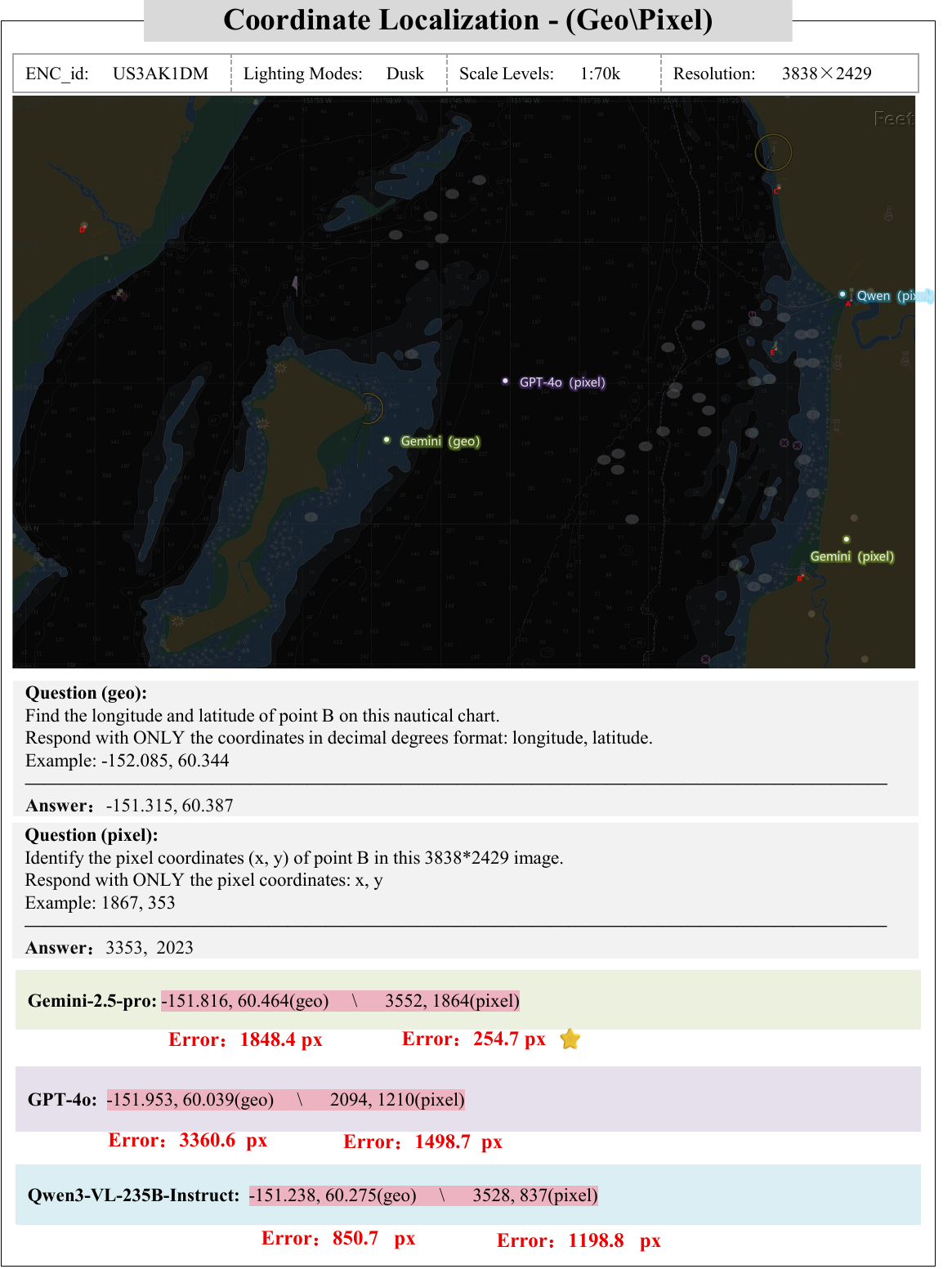}
    \caption{A sample case of \textbf{Coordinate Localization} (Dusk Mode).}
    \label{fig:case_17}
\end{figure*}

\begin{figure*}[t]
    \centering
    \includegraphics[width=0.9\textwidth, keepaspectratio]{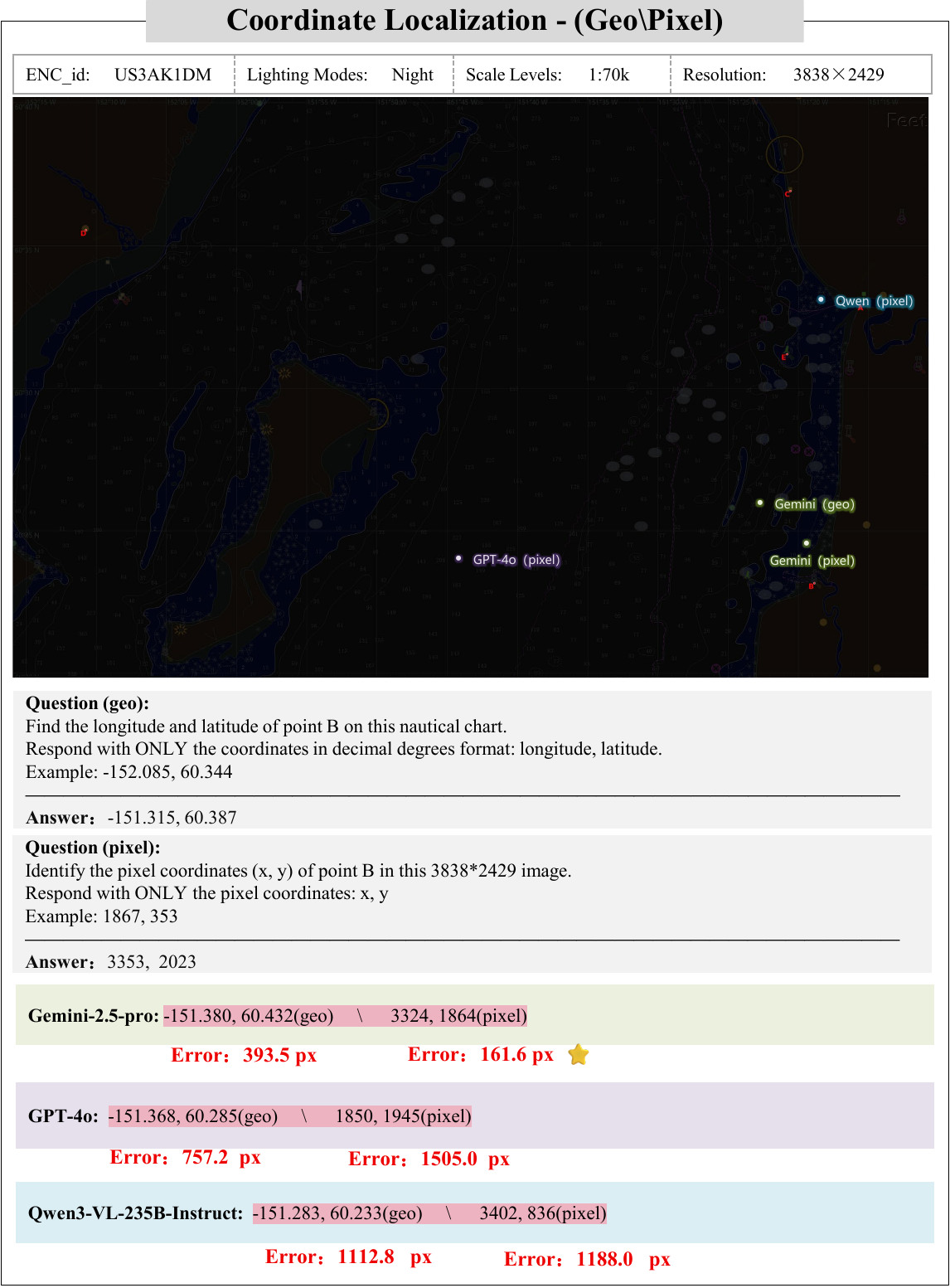}
    \caption{A sample case of \textbf{Coordinate Localization} (Night Mode).}
    \label{fig:case_18}
\end{figure*}

\begin{figure*}[t]
    \centering
    \includegraphics[width=0.9\textwidth, keepaspectratio]{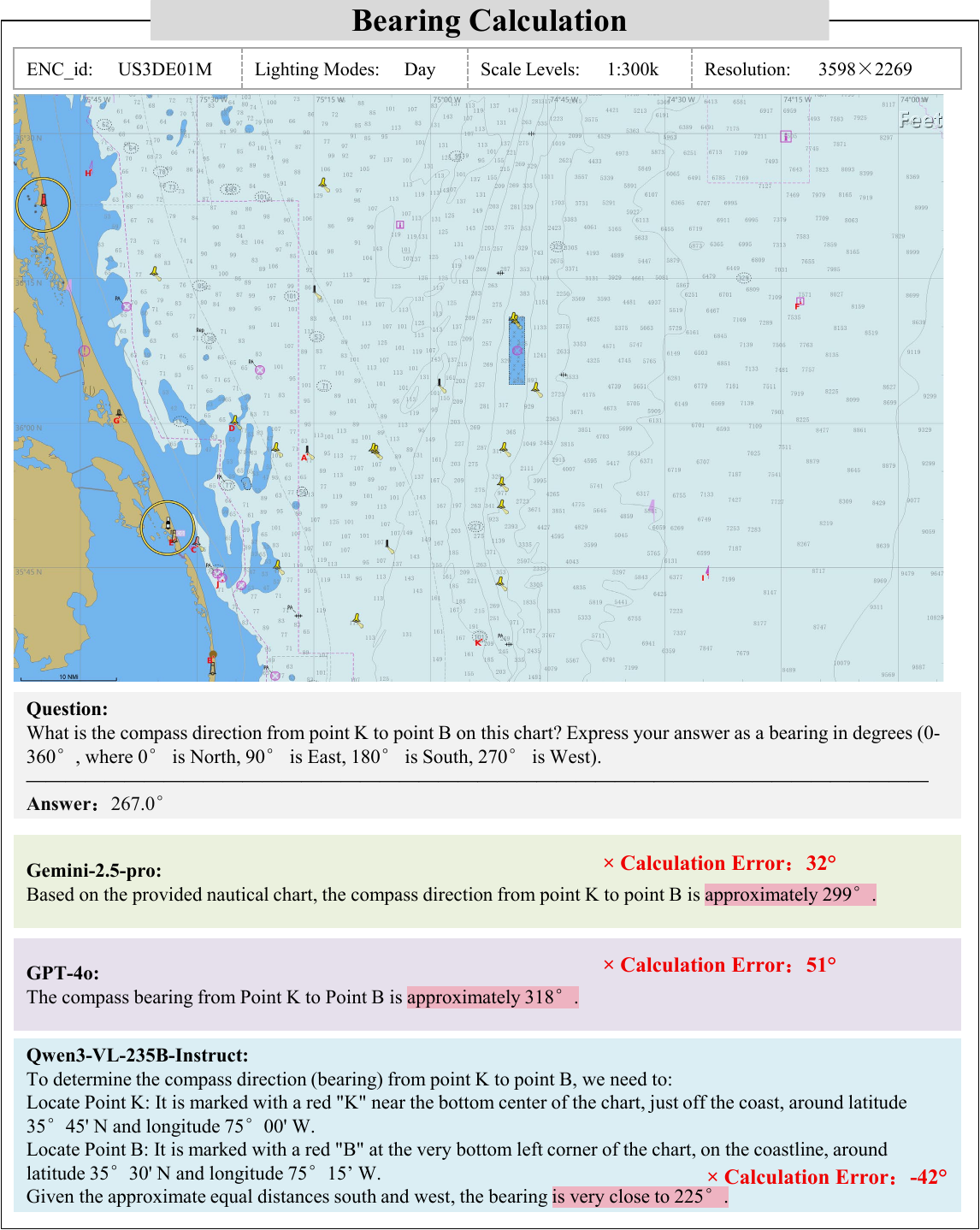}
    \caption{A sample case of \textbf{Bearing Calculation} (Day Mode).}
    \label{fig:case_19}
\end{figure*}

\begin{figure*}[t]
    \centering
    \includegraphics[width=0.9\textwidth, keepaspectratio]{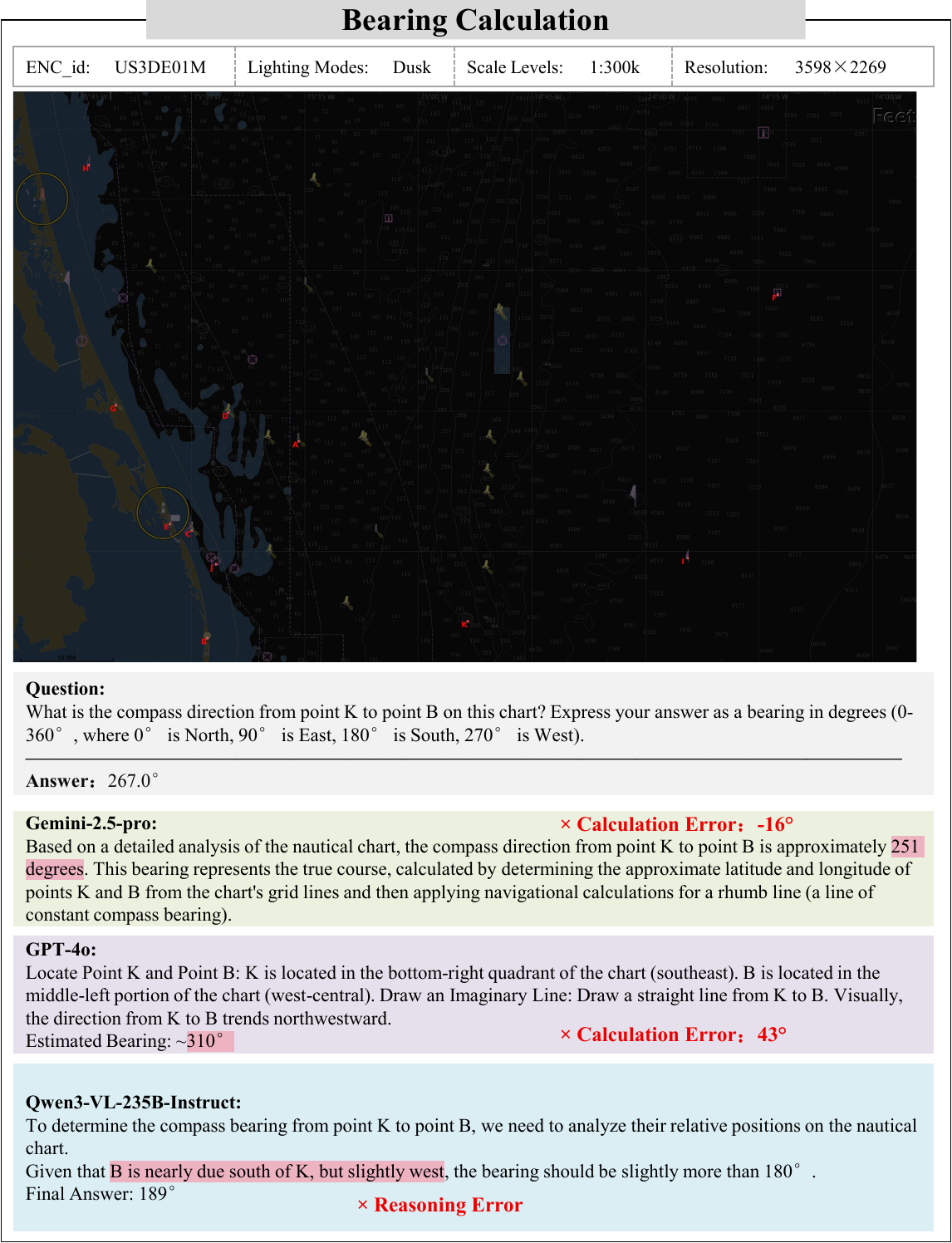}
    \caption{A sample case of \textbf{Bearing Calculation} (Dusk Mode).}
    \label{fig:case_20}
\end{figure*}

\begin{figure*}[t]
    \centering
    \includegraphics[width=0.9\textwidth, keepaspectratio]{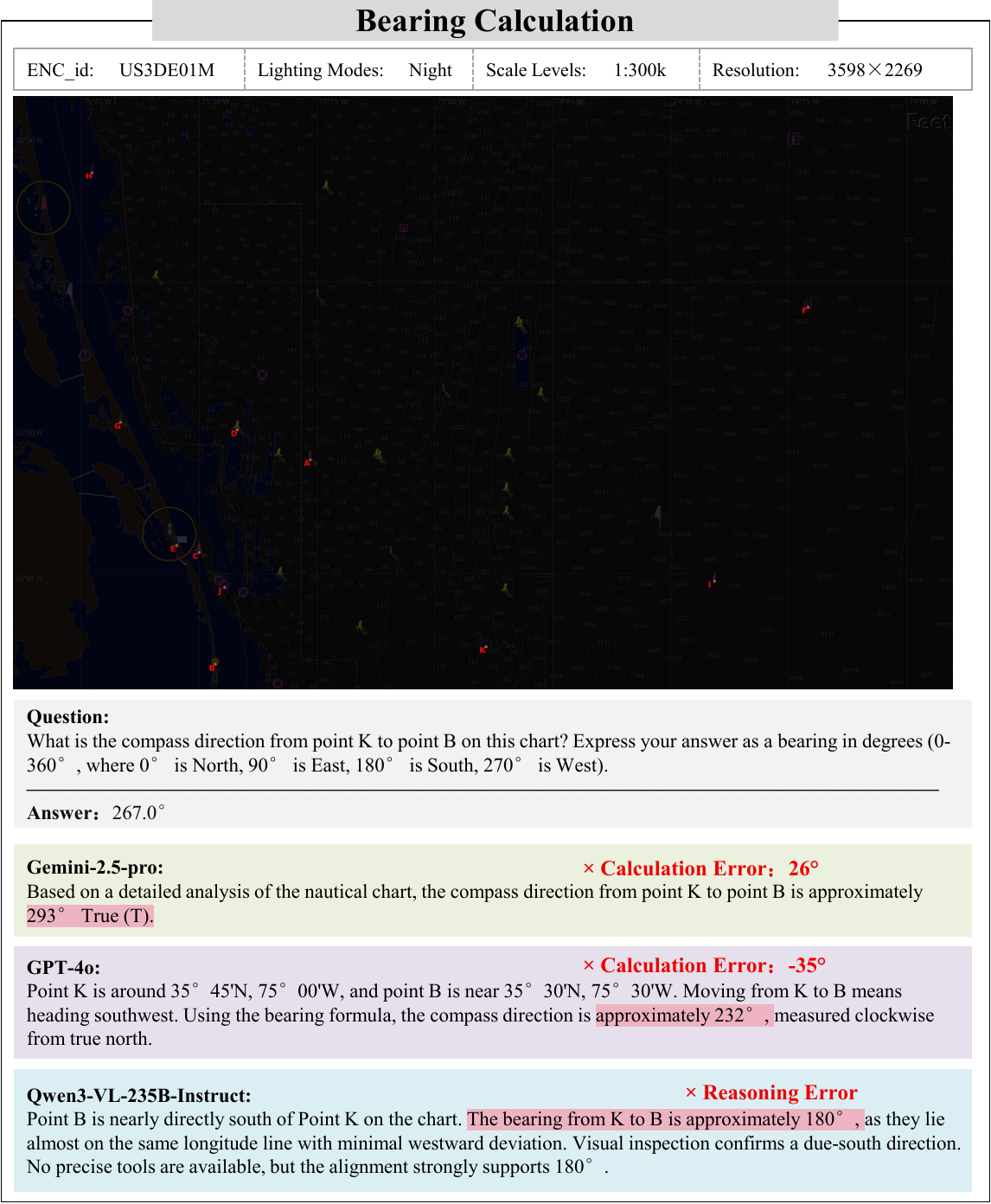}
    \caption{A sample case of \textbf{Bearing Calculation} (Night Mode).}
    \label{fig:case_21}
\end{figure*}

\begin{figure*}[t]
    \centering
    \includegraphics[width=0.9\textwidth, keepaspectratio]{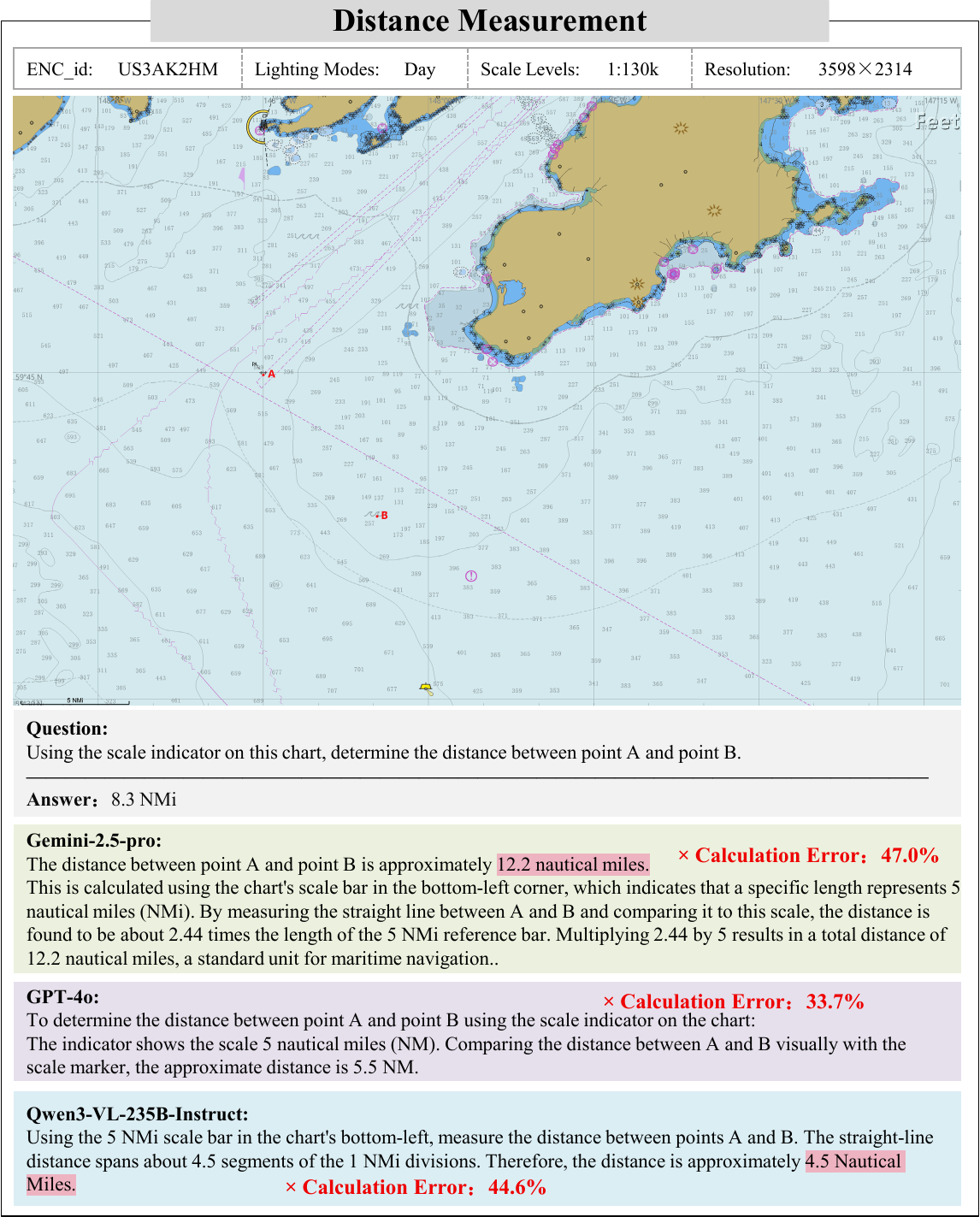}
    \caption{A sample case of \textbf{Distance Measurement} (Day Mode).}
    \label{fig:case_22}
\end{figure*}

\begin{figure*}[t]
    \centering
    \includegraphics[width=0.9\textwidth, keepaspectratio]{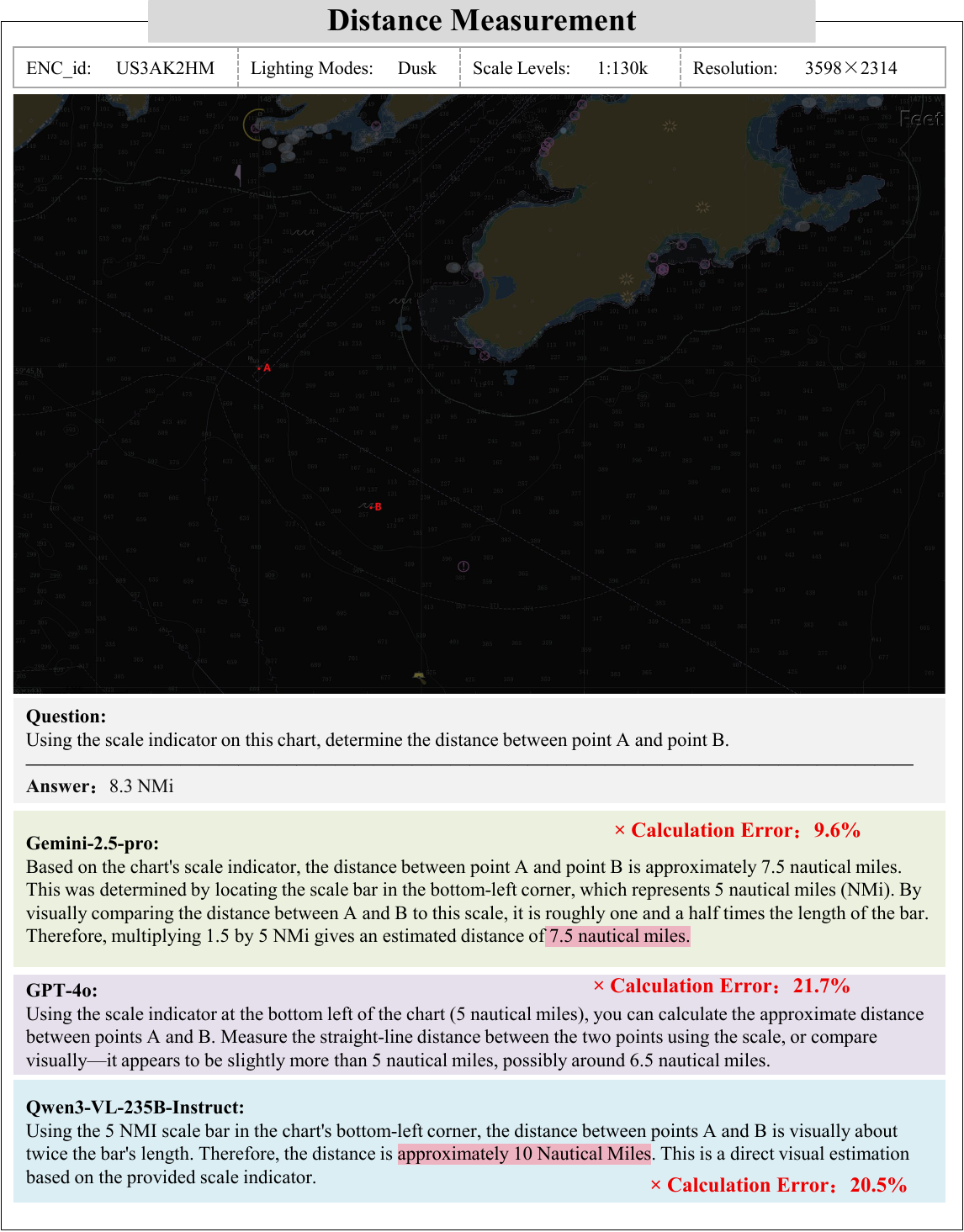}
    \caption{A sample case of \textbf{Distance Measurement} (Dusk Mode).}
    \label{fig:case_23}
\end{figure*}

\begin{figure*}[t]
    \centering
    \includegraphics[width=0.9\textwidth, keepaspectratio]{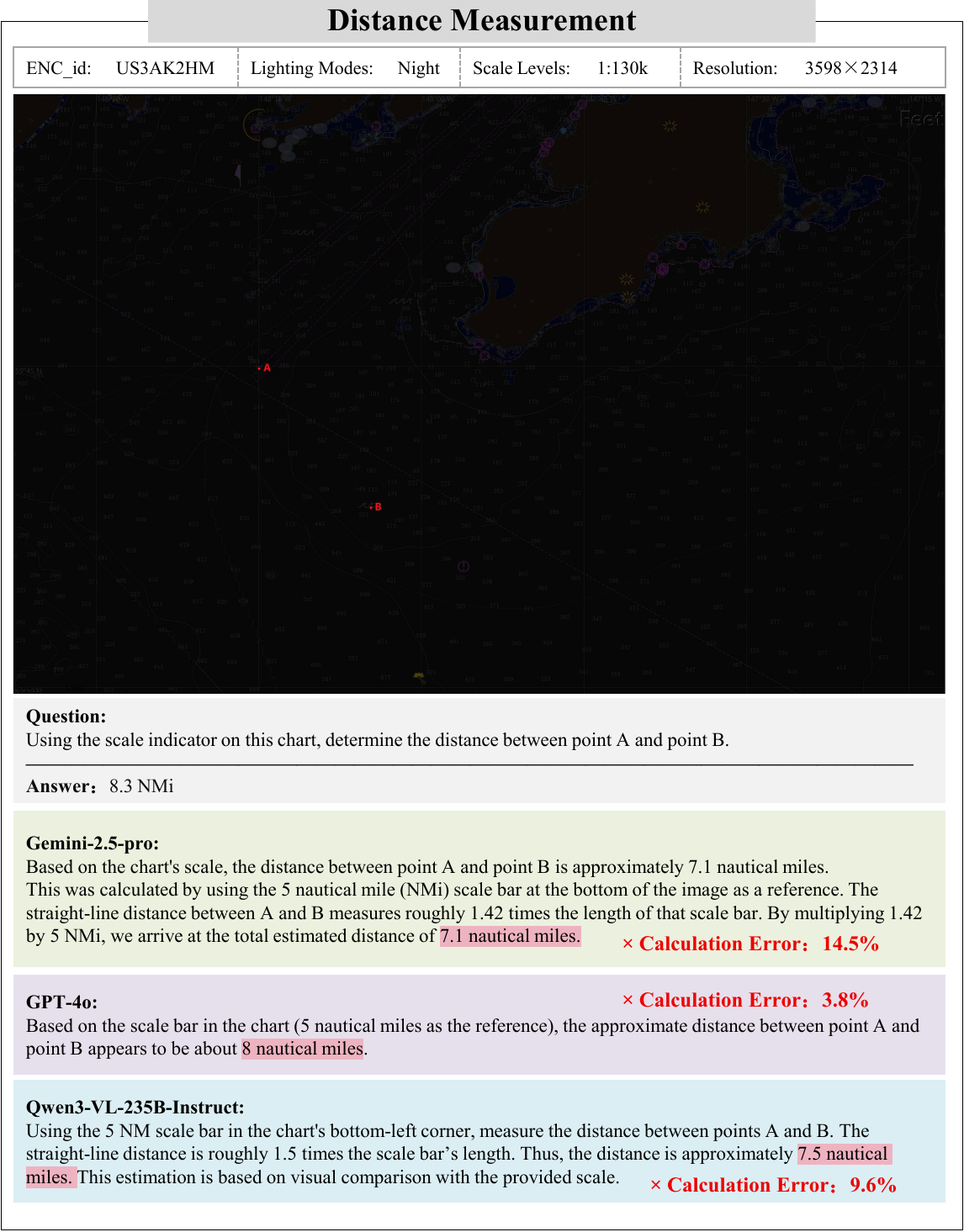}
    \caption{A sample case of \textbf{Distance Measurement} (Night Mode).}
    \label{fig:case_24}
\end{figure*}

\begin{figure*}[t]
    \centering
    \includegraphics[width=0.9\textwidth, keepaspectratio]{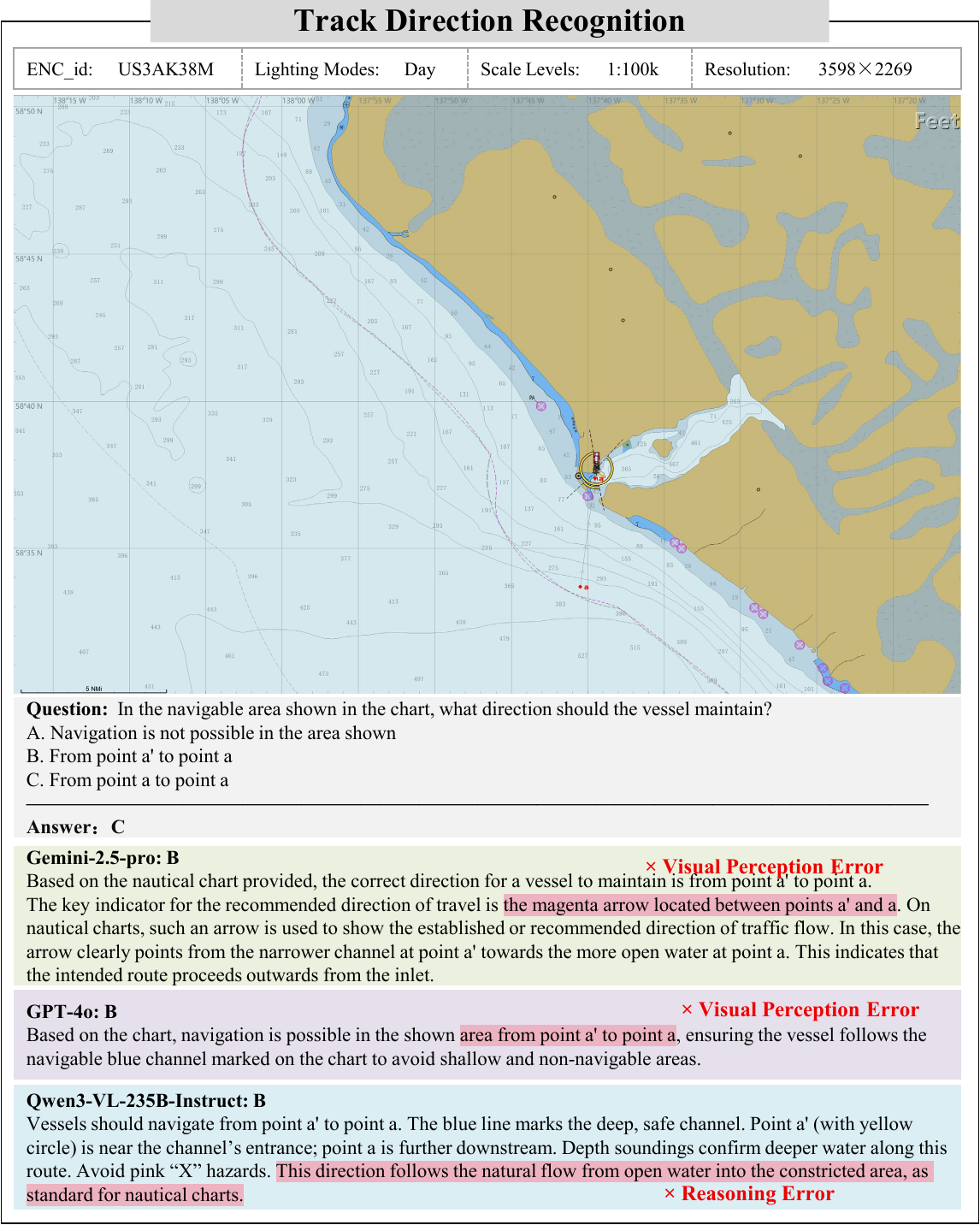}
    \caption{A sample case of \textbf{Track Direction Recognition} (Large Scale). Determining legal traffic flow direction.}
    \label{fig:case_25}
\end{figure*}

\begin{figure*}[t]
    \centering
    \includegraphics[width=0.9\textwidth, keepaspectratio]{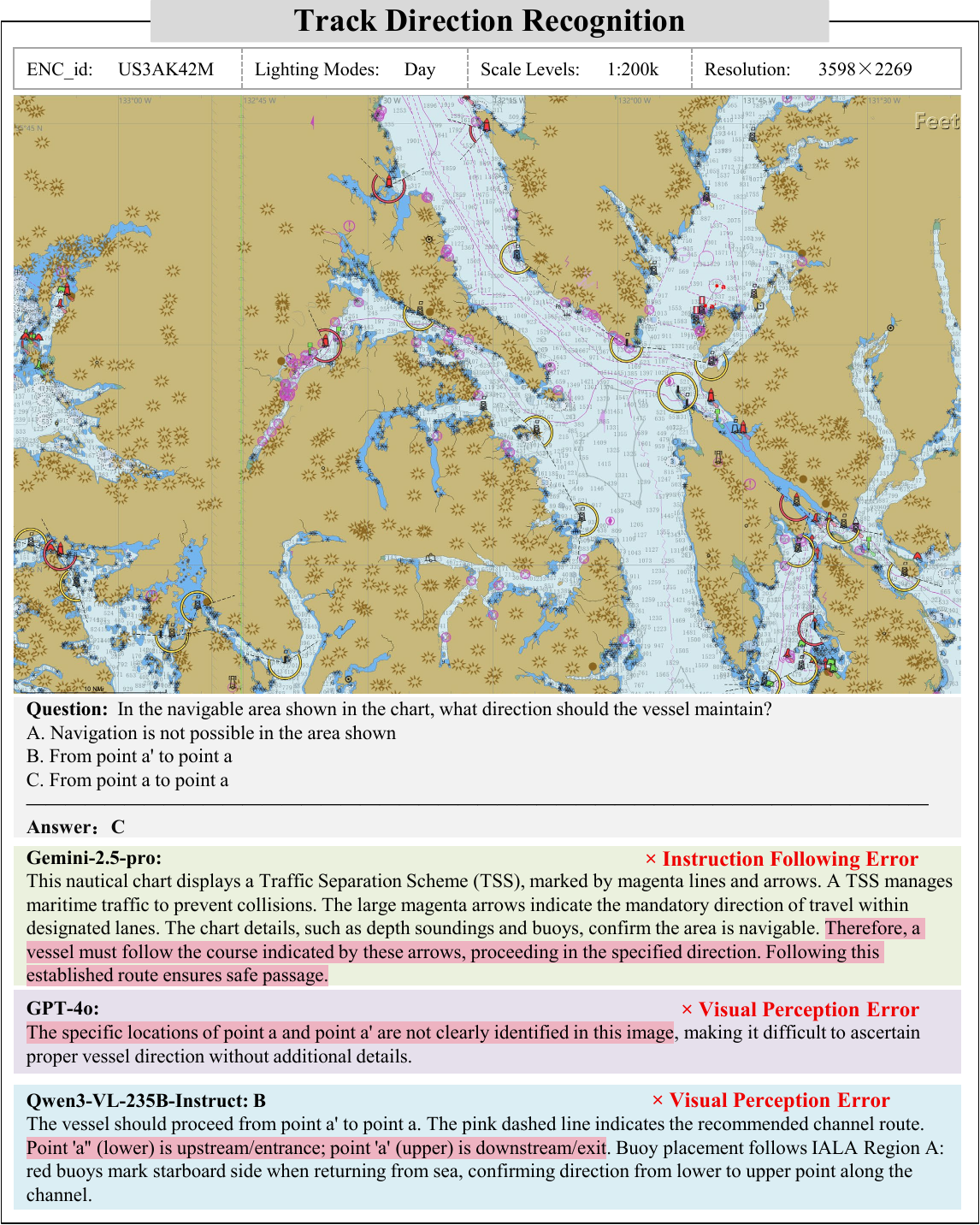}
    \caption{A sample case of \textbf{Track Direction Recognition} (Small Scale).}
    \label{fig:case_26}
\end{figure*}

\begin{figure*}[t]
    \centering
    \includegraphics[width=0.9\textwidth, keepaspectratio]{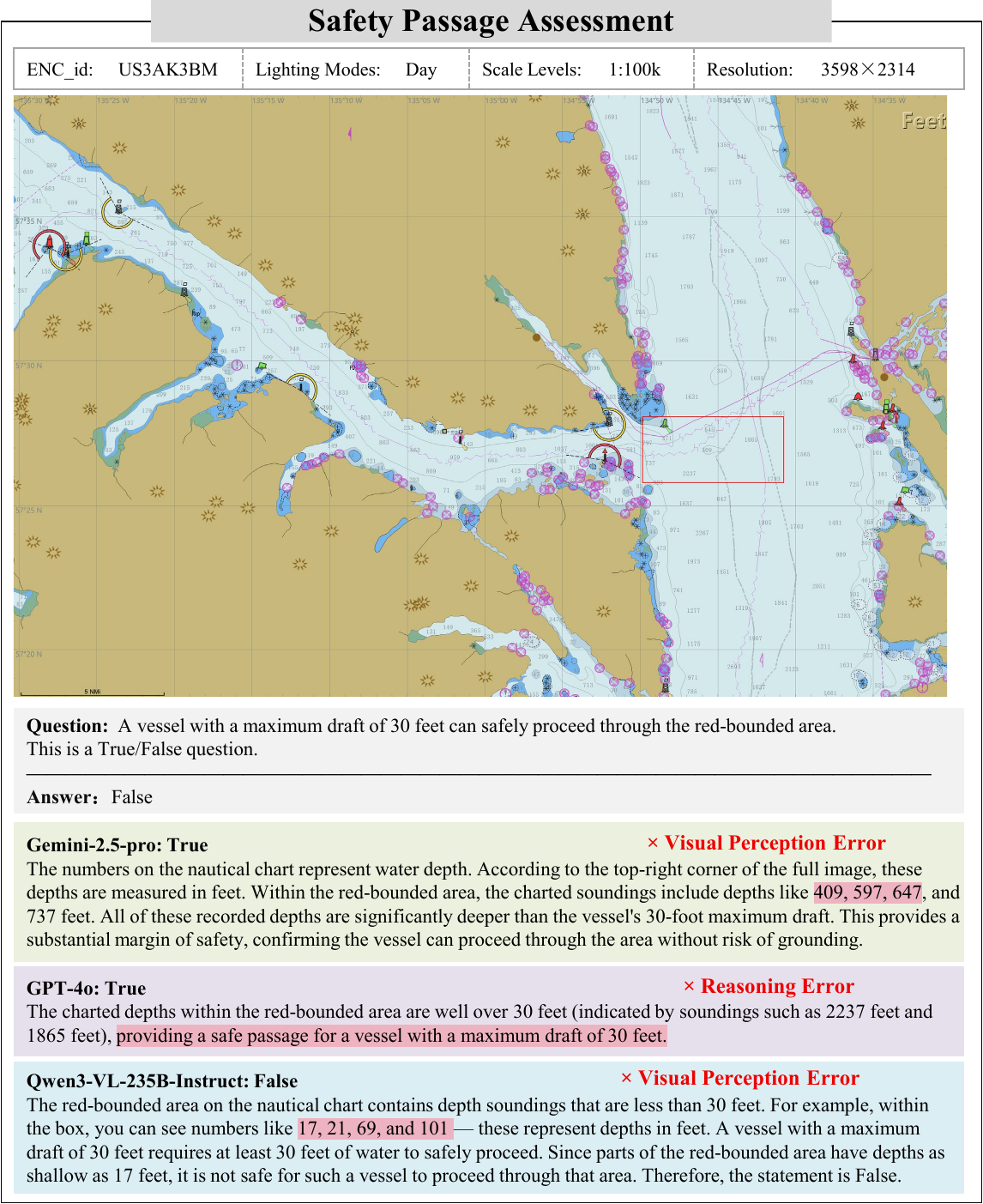}
    \caption{A sample case of \textbf{Safety Passage Assessment}. Evaluating depth constraints for a vessel draft.}
    \label{fig:case_27}
\end{figure*}

\begin{figure*}[t]
    \centering
    \includegraphics[width=0.9\textwidth, keepaspectratio]{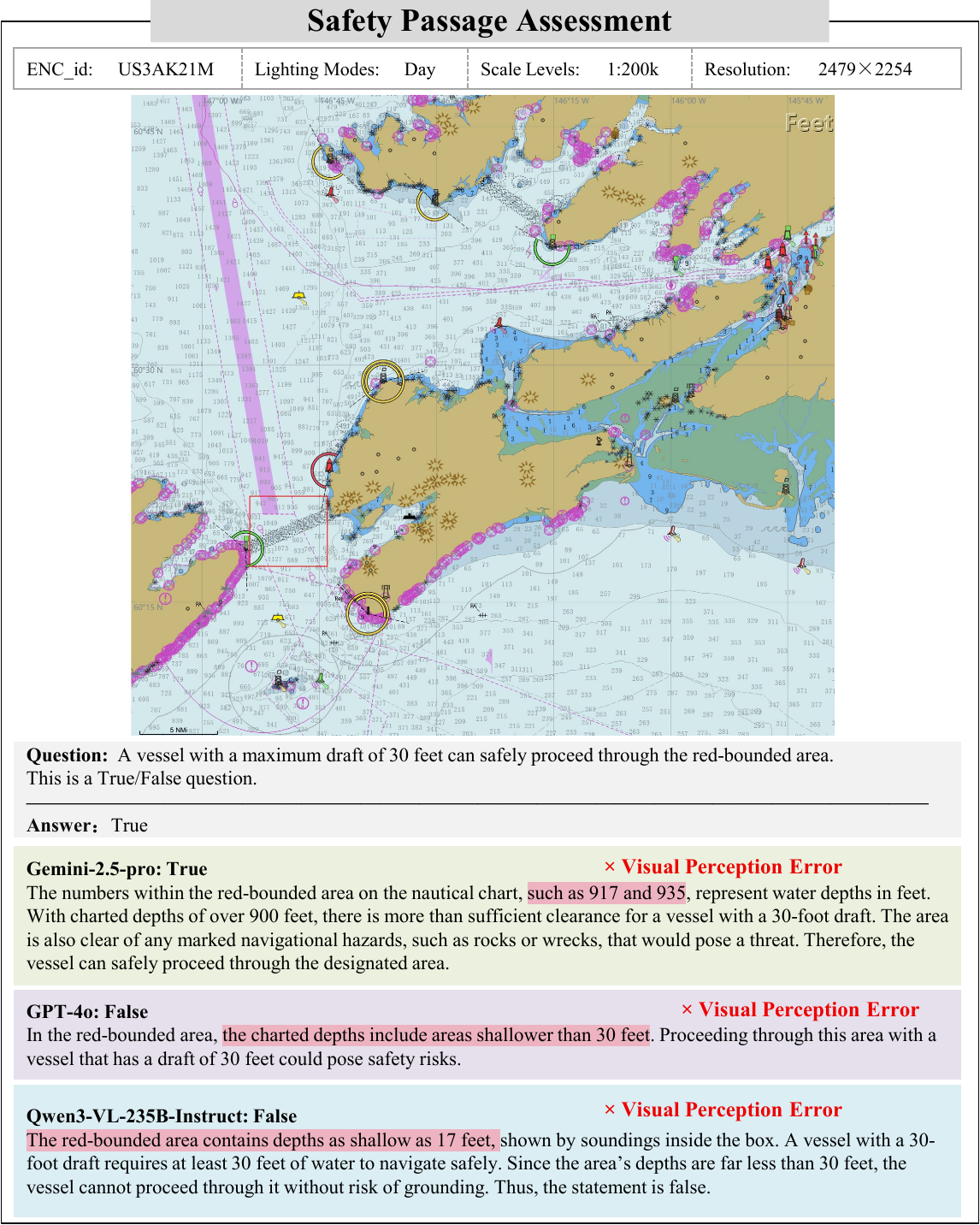}
    \caption{A sample case of \textbf{Safety Passage Assessment} (Small Scale).}
    \label{fig:case_28}
\end{figure*}

\begin{figure*}[t]
    \centering
    \includegraphics[width=0.9\textwidth, keepaspectratio]{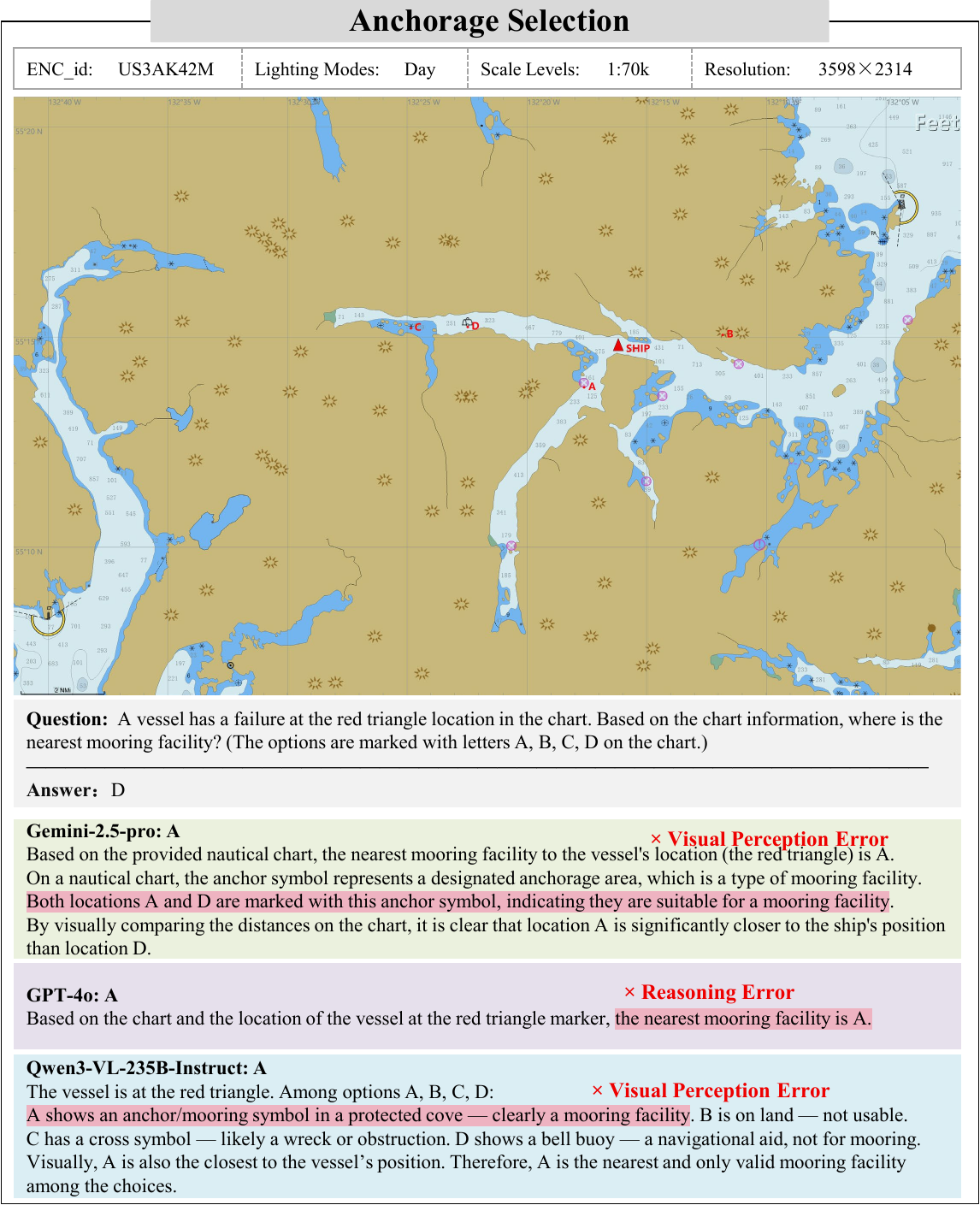}
    \caption{A sample case of \textbf{Anchorage Selection}. Choosing the nearest valid anchorage in an emergency.}
    \label{fig:case_29}
\end{figure*}

\begin{figure*}[t]
    \centering
    \includegraphics[width=0.9\textwidth, keepaspectratio]{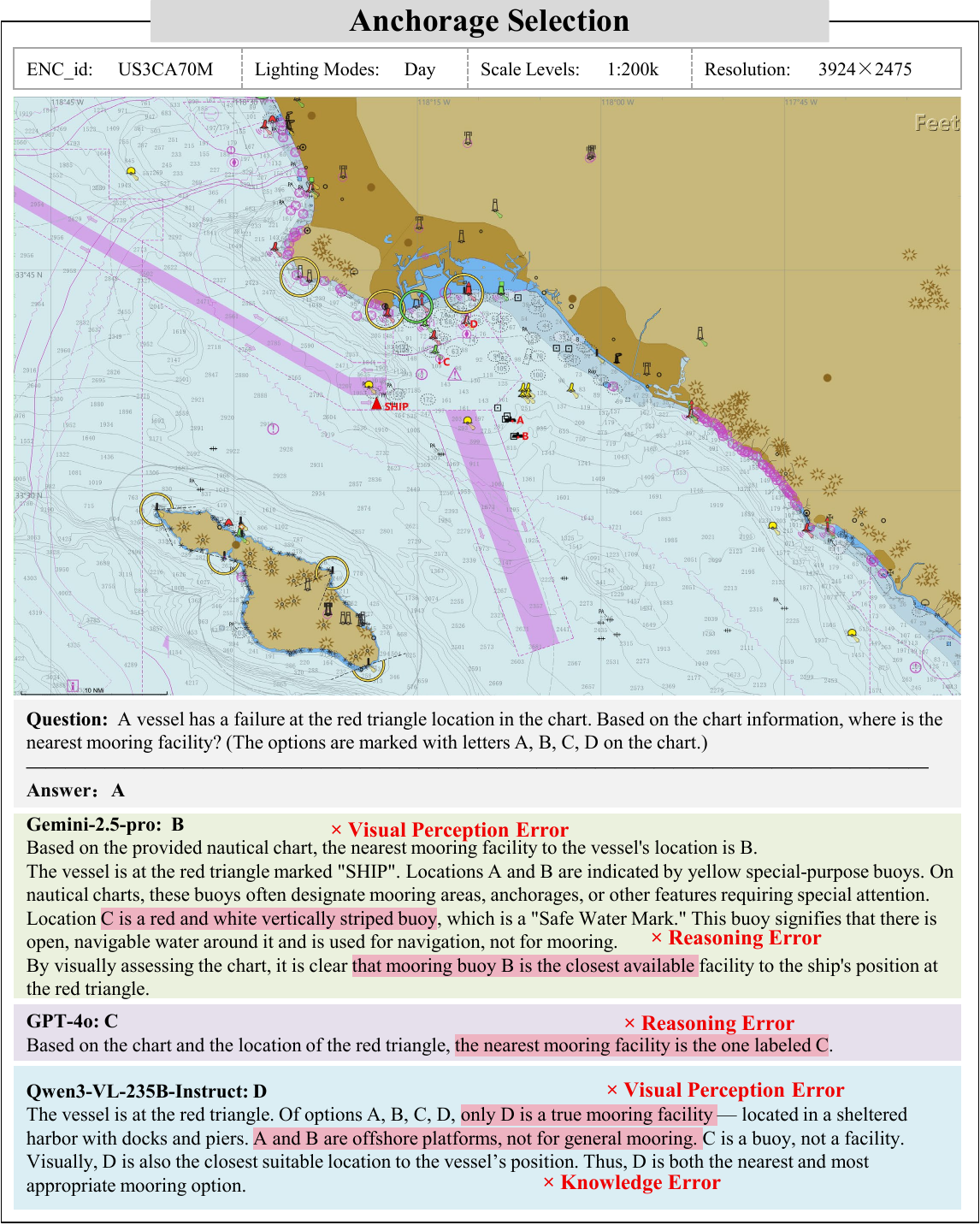}
    \caption{A sample case of \textbf{Anchorage Selection} (Small Scale).}
    \label{fig:case_30}
\end{figure*}

This appendix provides additional details on dataset construction, evaluation protocols, and experimental analysis:

\begin{itemize}[leftmargin=2em, labelsep=1em]
    \item \textbf{Sec.~\ref{sec:appendix-enc-primer}: Electronic Navigational Chart Primer.} \\
    Background on IHO S-57 vector structure, standardized symbology for point/line/polygon features, operational lighting modes, and scale-dependent rendering.

    \item \textbf{Sec.~\ref{sec:appendix-dataset-details}: Dataset Construction Details.} \\
    Semantic decoding of binary S-57 files, conflict-graph algorithm for density control, and task-specific generation: visible property filtering, analytical ground truth computation, and procedural scenario simulation.

    \item \textbf{Sec.~\ref{sec:appendix-prompts}: Evaluation Prompts.} \\
    Complete zero-shot prompt templates for all 10 benchmark tasks.

    \item \textbf{Sec.~\ref{sec:appendix-quantitative}: More Analysis of Results on ENC-Bench.} \\
    Fine-grained accuracy at varying error tolerances (Acc@T), performance comparison across distance/bearing/coordinate tasks, and analysis of the symbolic grounding gap between pixel and geographic localization.

    \item \textbf{Sec.~\ref{sec:appendix_cases}: Qualitative Analysis Case Studies.} \\
    30 annotated case studies (Figures \ref{fig:case_1}--\ref{fig:case_30}) demonstrating five error categories—Visual Perception, Reasoning, Knowledge, Calculation, and Instruction Following—across lighting modes and scale levels.

    \item \textbf{Sec.~\ref{sec:appendix-impact}: Limitations and Broader Impact.} \\
    Limitations of the current benchmark and the broader societal implications of deploying AI in safety-critical maritime navigation.
\end{itemize}

\begin{figure*}[t]
    \centering
    \includegraphics[width=1.0\textwidth]{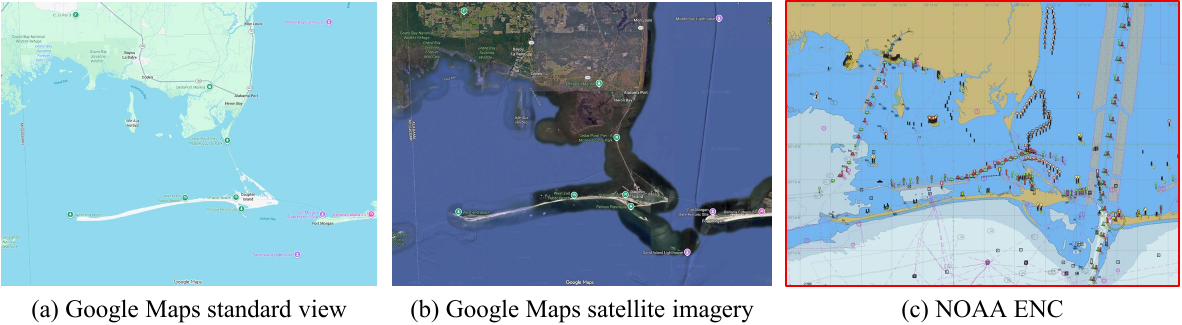}
    \caption{\textbf{Domain Contrast: Consumer Mapping vs. Professional Hydrography.} Comparison of the same geographic region across (a) Google Maps Standard View, (b) Google Maps Satellite Imagery, and (c) NOAA Electronic Navigational Chart (ENC). While consumer maps focus on road networks and vague water representations, ENCs illustrate a complex, vector-based hydrographic reality. Note how the ENC explicitly renders critical depth contours, shipping channels, and navigational aids absent in consumer views, prioritizing safety-critical information over visual realism.}
    \label{fig:app-domain-contrast} 
\end{figure*}

\section{Electronic Navigational Chart Primer}
\label{sec:appendix-enc-primer}

Electronic Navigational Charts (ENCs) differ fundamentally from natural images and consumer maps commonly used in computer vision. ENCs are safety-critical geospatial databases governed by the \textit{International Convention for the Safety of Life at Sea} (SOLAS) and encoded following the IHO S-57 standard. This section provides essential background to understand the domain-specific 
challenges in \textbf{ENC-Bench}.

\subsection{Vector-Based Data Structure}

Unlike pixel-based raster images, ENCs store data as object-oriented vector features with explicit geometric primitives (points, lines, polygons) and semantic attributes. Figure~\ref{fig:app-domain-contrast} contrasts consumer maps with ENCs: the latter employs standardized vector symbology defined by IHO S-57, encoding navigational semantics through formal geometric primitives rather than photorealistic textures. This symbolic representation ensures unambiguous interpretation across different display systems but challenges vision models trained on natural images.

\begin{figure*}[t]
    \centering
    \includegraphics[width=0.7\textwidth]{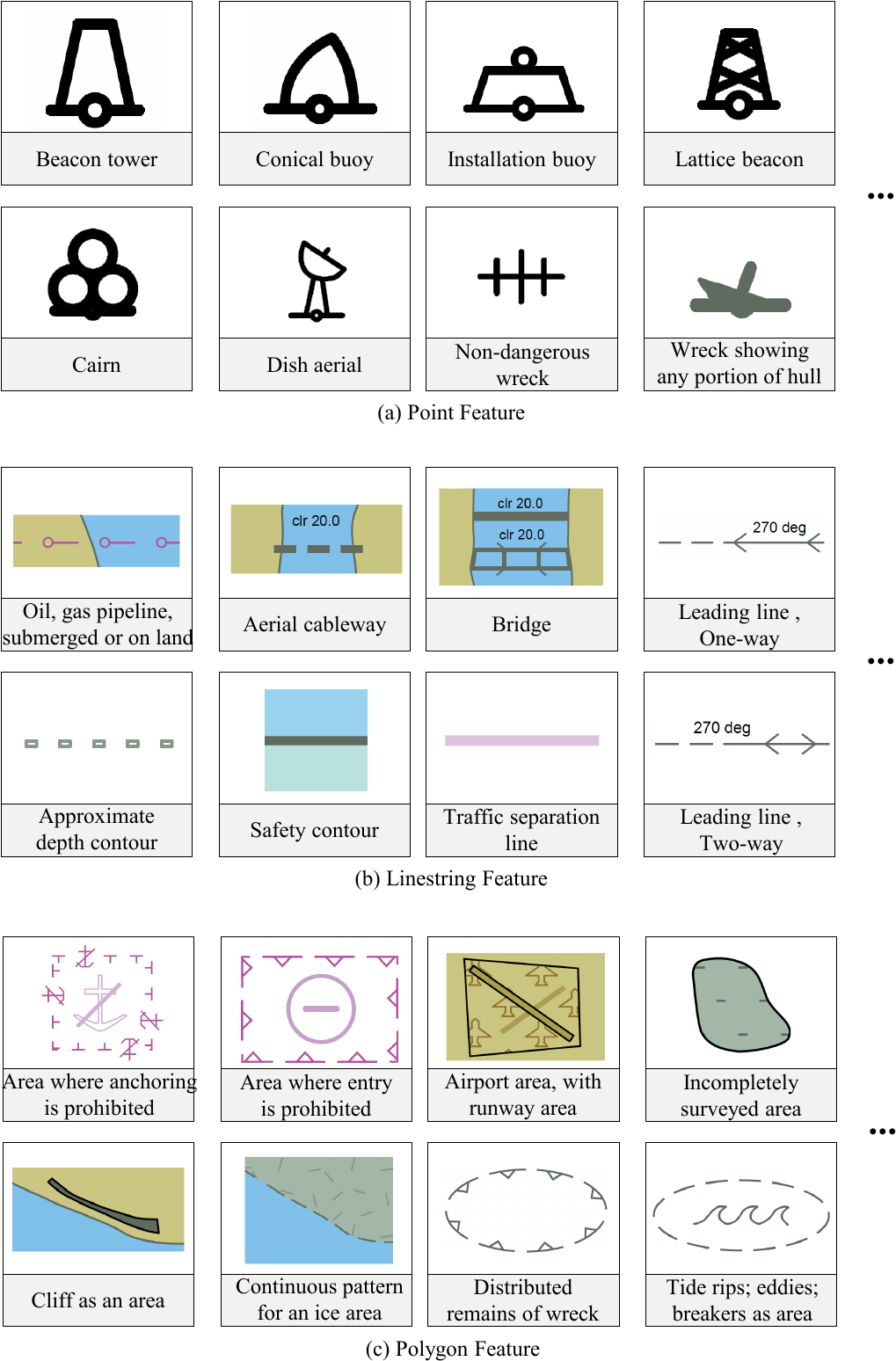}
    \caption{\textbf{Standardized IHO S-57 Symbology Types.} ENC features are classified into three geometric primitives with rigorous semantic definitions. \textbf{(a) Point Features:} Complex composite symbols where shape and topmark denote function (e.g., \textit{Beacon tower}, \textit{Non-dangerous wreck}). \textbf{(b) Linestring Features:} Define boundaries with directionality constraints, such as \textit{Pipelines} or \textit{Traffic separation lines}. \textbf{(c) Polygon Features:} Denote regional regulations, such as \textit{Anchoring prohibited areas} or \textit{Incompletely surveyed areas}. Recognizing these symbols requires distinguishing fine-grained visual details that carry heavy semantic weight.}
    \label{fig:app-symbols} 
\end{figure*}

Figure~\ref{fig:app-symbols} shows representative symbols for each 
geometric primitive type. Three feature types encode distinct 
navigational information:

\begin{itemize}
    \item \textbf{Point Features:} Navigational aids (buoys, beacons, lights) are rendered based on categorical attributes. A single object class (e.g., ``BOYCAR'' for cardinal buoys) can manifest in numerous distinct visual symbols depending on cardinal direction (N/E/S/W), topmark configuration, color pattern, and light characteristics (color, rhythm, period). Models must map visual symbols to structured attribute combinations rather than recognizing holistic object appearance.
    \item \textbf{Linestring Features:} Linear objects encode both geometric constraints and directional semantics. Recommended tracks specify safe navigation paths through complex waters; submarine cable areas mark zones where anchoring risks infrastructure damage; depth contours indicate transitions between bathymetric zones. Critically, line direction often carries regulatory meaning: a traffic lane permits one-way vessel movement, while a bidirectional ferry route allows traffic in both directions. Interpreting these features requires associating geometric primitives with their regulatory meanings rather than detecting visual edges.
    \item \textbf{Polygon Features:} Polygons define legal and operational zones rather than physical boundaries. Traffic separation schemes mandate vessel routing patterns; anchoring areas specify permissible mooring locations; restricted zones prohibit or constrain entry. These regions represent operational constraints rather than visual objects, requiring models to reason about rule compliance instead of object segmentation.
\end{itemize}

\subsection{Lighting Modes}
\label{sec:app-lighting}
Maritime navigation operates around the clock. To protect mariners' night vision during dark watches, ECDIS systems render charts in three distinct color schemes mandated by IHO S-52: \textbf{Day} ,\textbf{ Dusk} , and \textbf{Night}. 

\begin{figure*}[t]
    \centering
    \includegraphics[width=1\textwidth]{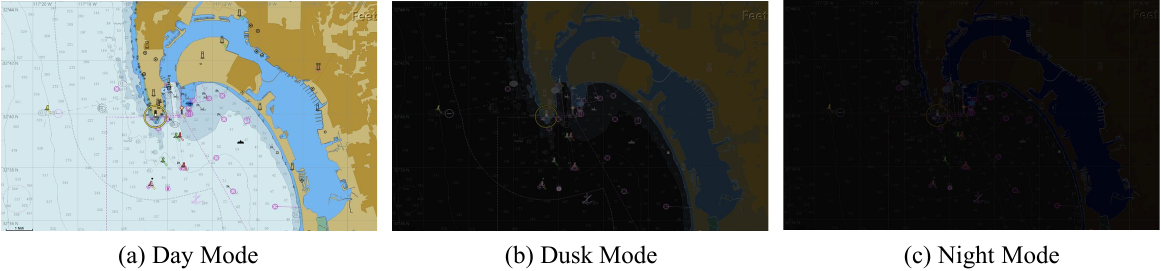}
    \caption{\textbf{Operational Lighting Modes.} Demonstration of the ECDIS standardized color palettes. \textbf{(a) Day Mode:} High contrast with white/blue backgrounds. \textbf{(b) Dusk Mode:} Reduced glare with grey backgrounds. \textbf{(c) Night Mode:} Black background with non-linear color shifts. Note how the blue shallow water in Day mode transforms into dark grey/black tones in Night mode, and text labels shift colors to maintain visibility. This drastic ``palette swapping'' challenges MLLMs reliant on natural image color statistics.}
    \label{fig:app-lighting} 
\end{figure*}

Figure~\ref{fig:app-lighting} shows how the same chart transforms across these modes. Unlike adjusting screen brightness, mode switching involves systematic color remapping that creates two fundamental challenges for vision models:
\begin{itemize}
    \item \textbf{Background Inversion:} Day mode employs bright backgrounds with darker symbols; Night mode reverses this to preserve night vision, using dark backgrounds with bright, high-contrast symbols. This inversion fundamentally changes visual patterns: the same buoy that appears as a dark icon in Day mode becomes a bright marker in Night mode. Models relying on consistent contrast polarity must adapt to these reversals.
    \item \textbf{Color Meaning Changes:} The S-52 standard defines distinct color palettes for each mode. Features do not simply darken or brighten—they undergo systematic color transformation to maintain discriminability against different backgrounds. Consequently, color alone becomes unreliable for feature recognition: a model must identify features through their geometric structure and symbolic form rather than learned color-category associations.
\end{itemize}

\subsection{Scale-Dependent Rendering (SCAMIN)}

In operational use, mariners frequently adjust chart scales—zooming out for regional route planning and zooming in for precise maneuvering. To maintain readability, the system implements scale-based feature filtering: less important objects hide at small scales and appear at large scales. This follows the IHO S-57 SCAMIN (Scale Minimum) attribute, which specifies the minimum display scale for each feature.

\begin{figure*}[t]
    \centering
    \includegraphics[width=1\textwidth]{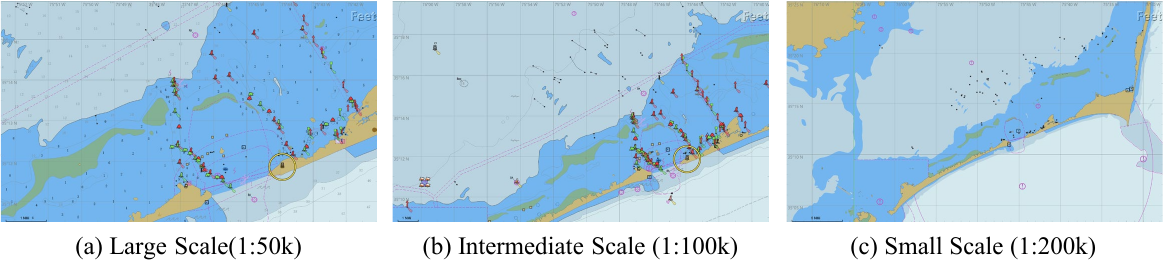}
    \caption{\textbf{Scale-Dependent Rendering and Cartographic Generalization.} The same chart region rendered at three standard scale levels. \textbf{(a) Large Scale (1:50k):} Maximum detail with dense soundings (e.g., depth values like \textit{15} or \textit{16} feet) and all navigational aids visible. \textbf{(b) Intermediate Scale (1:100k):} Partial generalization where minor soundings are suppressed to reduce clutter. \textbf{(c) Small Scale (1:200k):} High-level abstraction where only critical features remain, and dense depth numbers are removed. Models must handle this dynamic appearance where features logically ``exist'' but visually vanish based on zoom level.}
    \label{fig:app-scale} 
\end{figure*}

Figure~\ref{fig:app-scale} illustrates this mechanism at three representative scales. The information density varies systematically:

\begin{itemize}[leftmargin=*,noitemsep,topsep=2pt]
    \item \textbf{1:200k (zoomed out):} Major shipping channels, primary navigation aids, and critical depth contours appear. Minor buoys, individual depth soundings, and local hazards remain hidden.
    \item \textbf{1:100k (intermediate):} Additional buoys and secondary routes become visible. Depth information increases but detailed soundings remain sparse.
    \item \textbf{1:50k (zoomed in):} Full detail emerges—every buoy, individual depth measurements, submarine cables, and local restricted areas appear.
\end{itemize}

This scale-dependent visibility creates a fundamental challenge: the same location appears visually different at different scales. A region that looks "clear" at 1:200k may reveal numerous hazards at 1:50k. Models must recognize that feature absence at small scales does not imply physical absence—features may simply be display-suppressed due to scale filtering.

ENC-Bench systematically samples six scale levels (1:50k, 1:70k, 1:100k, 1:130k, 1:200k, 1:300k) for the same geographic regions. This evaluates whether models can correctly interpret features at different scales, adjusting their reasoning based on which details should be visible at each zoom level—mirroring how human navigators mentally account for scale when reading charts.

\clearpage

\clearpage

\clearpage
\section{Dataset Construction Details}
\label{sec:appendix-dataset-details}

The construction of \textbf{ENC-Bench} is governed by a rigorous, scientifically controlled pipeline designed to transform raw S-57 hydrographic data into a verifiable visual-language benchmark. Unlike datasets relying on crowdsourced annotations, our pipeline derives ground truth directly from the official IHO S-57 vector database, ensuring logical consistency and nautical plausibility.

\subsection{Semantic Decoding and Entity Resolution}
Raw ENC data is encapsulated in the ISO/IEC 8211 binary format, which optimizes storage but obscures semantics behind cryptic acronyms. To enable automated reasoning, we developed a custom semantic parsing engine to decode these binary structures into usable semantic primitives.

\noindent\textbf{Attribute Expansion.} 
We utilize the official IHO S-57 Object Catalogue lookup tables (specifically \texttt{s57objectclasses.csv}, \texttt{s57attributes.csv}, and \texttt{s57expectedinput.csv}) to map binary codes to human-readable semantics. For instance, the acronym \texttt{BOYLAT} is expanded to \textit{Buoy, Lateral}, and the enumerated attribute \texttt{COLOUR:3} is mapped to \textit{Red}. Depth values stored in meters (\texttt{VALSOU}) are standardized and converted to feet/fathoms where required to match the visual labels rendered on specific chart styles.

\noindent\textbf{Cross-Layer Entity Resolution.} 
A critical challenge in S-57 data is that a single maritime object is often fragmented across multiple geometric layers (e.g., a lighthouse represented as a point for the light and a separate polygon for the structure). We implement an entity resolution algorithm using the unique \texttt{LNAM} (Long Name) identifier. Features sharing \texttt{LNAM} or listed in \texttt{LNAM\_REFS} are aggregated into a single semantic object. This ensures that questions query the holistic entity rather than disjoint geometric fragments.

\subsection{Adaptive Density Control via Graph Coloring}
Professional ENCs are visually dense. Naively sampling features for evaluation often leads to overlapping bounding boxes or ambiguous references (e.g., two buoys too close to distinguish). We solve this using an \textbf{Adaptive Density Control} strategy modeled as a Graph Coloring problem.

We construct a \textit{Conflict Graph} $G=(V,E)$ where each node $v \in V$ represents a candidate feature. An undirected edge $e_{ij}$ is drawn between two features if their projected pixel distance is below a visual threshold $\tau$ (set to 40 pixels in our implementation). We then apply a greedy graph coloring algorithm to partition $V$ into $k$ independent sets (groups). For a single chart view, we generate multiple distinct evaluation samples, where each sample contains only the features from one independent set. This mathematically guarantees that all targeted features in a given question (labeled A, B, C...) are visually separated and unambiguous.

\subsection{Hierarchical Task Generation Logic}
We employ distinct generation strategies for each of the three capability tiers, ensuring that ground truth is derived from precise analytical constraints rather than visual estimation.

\subsubsection{L-1 Perception: Visible Property Filtering}
A common pitfall in VQA dataset generation is "information leakage," where models access metadata present in the database but not visually rendered (e.g., the unique identifier or installation date of a buoy). 

To prevent this, we implement a strict \textbf{Visible Property Filter}. Based on ECDIS rendering rules, we defined a whitelist of visually observable attributes for each feature class. For example, for a \textit{Cardinal Mark}, we retain \texttt{COLOUR}, \texttt{BCNSHP} (shape), and \texttt{TOPSHP} (topmark), while explicitly stripping non-visual attributes like \texttt{OBJNAM} (Object Name) and \texttt{PEREND} (Period End). The VLM prompt generator is conditioned \textit{only} on these filtered attributes, ensuring that the resulting QA pairs test genuine visual recognition.

\subsubsection{L-2 Spatial Reasoning: Analytical Ground Truth}
Spatial reasoning tasks are grounded in computational geometry using the precise geodetic coordinates extracted from the S-57 vector data.

\begin{itemize}
    \item \textbf{Coordinate Localization:} We evaluate both geodetic (Latitude/Longitude) and pixel-space localization. Pixel coordinates are computed via an affine transformation matrix $\mathbf{M}$ calibrated using manually annotated control points for each chart, ensuring sub-pixel accuracy.
    \item \textbf{Distance Measurement:} Ground truth distance $d$ is calculated using the \textbf{Haversine formula} with an Earth radius $R = 3440.065$ nautical miles. This accounts for the curvature of the earth, which is significant at the scales of maritime navigation.
    \item \textbf{Bearing Calculation:} True north bearings are calculated using the inverse geodetic problem ($\text{atan2}(\Delta\lambda, \Delta\phi)$), adjusted to the 0-360° compass convention.
\end{itemize}

\subsubsection{L-3 Decision-Making: Procedural Simulation}
For high-level tasks, we developed a simulation engine to synthesize navigation scenarios that require multi-constraint reasoning.

\noindent\textbf{Safety Passage Assessment.} 
We generate queries "Can a vessel with draft $D$ safely pass?" by spatially joining a procedurally generated random polygon with underlying depth layers (\texttt{DEPARE}, \texttt{SOUNDG}). The ground truth is determined by a strict inequality:
\begin{equation}
    \text{Safe} \iff D + \delta_{safety} < \min(|d_{val}|)
\end{equation}
where $\delta_{safety}$ is a fixed safety margin (set to 2.0 feet) and $\min(|d_{val}|)$ is the absolute minimum depth found within the query polygon. This tests the model's ability to perform OCR on depth numbers and apply arithmetic logic.

\noindent\textbf{Track Direction Recognition.}
This task tests adherence to traffic flow regulations. We extract \texttt{RCRTCL} (Recommended Track) features which contain an \texttt{ORIENT} attribute specifying the mandatory heading. For a track segment with endpoints $a$ and $a'$, we calculate the geometric bearing $\theta_{a \rightarrow a'}$ and its reciprocal $\theta_{a' \rightarrow a}$. The correct direction is identified by minimizing the angular distance to the database \texttt{ORIENT} value, forcing the model to visually correlate the line's orientation with implicit traffic rules.

\noindent\textbf{Anchorage Selection.}
We synthesize emergency scenarios by digitally superimposing a red triangle symbol (representing a vessel) onto the chart image. The ground truth is computed by:
1. Querying all \texttt{MORFAC} (Mooring Facility) objects in the database.
2. Computing the geodesic distance from the synthetic vessel to each facility.
3. Selecting the nearest valid facility as the target.
This requires the model to perform multi-hop reasoning: \textit{detect vessel $\rightarrow$ detect anchorages $\rightarrow$ estimate distances $\rightarrow$ select minimum}.

\section{Evaluation Prompts}
\label{sec:appendix-prompts}

To ensure reproducibility and facilitate future benchmarking, we detail the specific prompt templates used for ENC-Bench. All models were evaluated in a zero-shot setting, with task instructions and output formatting constraints directly embedded in the user query.

Table~\ref{tab:prompt-templates} presents the complete list of templates. Placeholders such as \texttt{\{label\}}, \texttt{\{width\}}, and \texttt{\{options\}} are dynamically filled with instance-specific data during evaluation.

\begin{table*}[h]
\centering
\caption{\textbf{Task-Specific Prompt Templates.} We provide distinct prompts for each task. Note that for Coordinate Localization, we evaluate both Geographic (Latitude/Longitude) and Pixel-space prediction using separate prompts.}
\label{tab:prompt-templates}
\small
\renewcommand{\arraystretch}{1}
\begin{tabular}{@{}p{0.22\textwidth} p{0.74\textwidth}@{}}
\toprule
\textbf{Task} & \textbf{Prompt Template} \\
\midrule
\multicolumn{2}{c}{\cellcolor{gray!15}\textbf{L-1 Perception}} \\
\midrule
\textbf{Symbol Recognition} & The image displays a symbol used to represent ENC (Electronic Navigational Chart) data on an ECDIS. According to the description provided, this symbol specifically portrays what type of object or feature?\newline 
Please choose the correct answer from the following options:\newline
\texttt{\{options\}}\newline
Please respond with only the letter (A, B, C, or D) of the correct answer. \\
\midrule
\textbf{Point Feature} & What is the nautical symbol located above point labeled ``\texttt{\{label\}}''?\newline
Please choose the correct answer from the following options:\newline
\texttt{\{options\}}\newline
Please respond with only the letter (A, B, C, or D) of the correct answer. \\
\midrule
\textbf{Linestring Feature} & What does the linestring between point \texttt{\{label\_start\}} and point \texttt{\{label\_end\}} represent?\newline
Please choose the correct answer from the following options:\newline
\texttt{\{options\}}\newline
Please respond with only the letter (A, B, C, or D) of the correct answer. \\
\midrule
\textbf{Polygon Feature} & What is the most likely meaning of the area \texttt{\{bbox\_label\}}?\newline
Please choose the correct answer from the following options:\newline
\texttt{\{options\}}\newline
Please respond with only the letter (A, B, C, or D) of the correct answer. \\
\midrule
\multicolumn{2}{c}{\cellcolor{gray!15}\textbf{L-2 Spatial Reasoning}} \\
\midrule
\textbf{Coordinate Localization} \newline \textit{(Geographic)} & Find the longitude and latitude of point \texttt{\{label\}} on this nautical chart.\newline
Respond with ONLY the coordinates in decimal degrees format: longitude,latitude\newline
Example: -152.085,60.344 \\
\midrule
\textbf{Coordinate Localization} \newline \textit{(Pixel)} & Identify the pixel coordinates (x, y) of point \texttt{\{label\}} in this \texttt{\{width\}}*\texttt{\{height\}} image.\newline
Respond with ONLY the pixel coordinates: x,y\newline
Example: 1867,353 \\
\midrule
\textbf{Bearing Calculation} & On this nautical chart, what is the bearing from point \texttt{\{label\_start\}} to point \texttt{\{label\_end\}}? Please provide your answer as a compass bearing in degrees (0-360°, where 0° is North, 90° is East, 180° is South, 270° is West).\newline
Respond with ONLY the numerical bearing value in degrees (0-360°).\newline
Example: 285.1° \\
\midrule
\textbf{Distance Measurement} & On this chart, with reference to the scale, calculate the distance between point \texttt{\{label\_start\}} and point \texttt{\{label\_end\}}.\newline
Respond with ONLY the numerical distance value in nautical miles.\newline
Example: 1.46 \\
\midrule
\multicolumn{2}{c}{\cellcolor{gray!15}\textbf{L-3 Maritime Decision-Making}} \\
\midrule
\textbf{Track Direction} & What is the correct direction for vessel navigation in the navigable area shown?\newline
Please choose the correct answer from the following options:\newline
\texttt{\{options\}}\newline
Please respond with only the letter (A, B, or C) of the correct answer. \\
\midrule
\textbf{Safety Assessment} & It is safe for a vessel with a maximum draft of \texttt{\{draft\}} feet to pass through the area indicated by the red box.\newline
This is a True/False question.\newline
Please answer strictly with a single word: True or False. \\
\midrule
\textbf{Anchorage Selection} & A vessel has a failure at the red triangle location in the chart. Based on the chart information, where is the nearest mooring facility?\newline
Please look at the chart image carefully. The red triangle shows the vessel's failure location. The options are marked with letters A, B, C, D on the chart. Please identify the nearest anchorage to the red triangle location and respond with only the letter (A, B, C, or D). \\
\bottomrule
\end{tabular}
\end{table*}

\section{More Analysis of Results on ENC-Bench}
\label{sec:appendix-quantitative}

Due to space limitations, more in-depth analyses to advance MLLM research in specialized maritime domains are provided in this appendix. This section highlights the fine-grained performance of spatial reasoning capabilities.

\textbf{Performance Sensitivity to Error Tolerance.}
As shown in Tables~\ref{tab:app-distance-acc} and \ref{tab:app-bearing-acc}, model accuracy exhibits significant sensitivity to error tolerance thresholds. In the Distance Measurement task, while Gemini-2.5-Flash achieves 25.93\% accuracy at the standard 20\% tolerance, its performance declines to 6.92\% at the stricter 5\% threshold. This trend is consistent across all evaluated models, with even the top-performing Gemini-2.5-Pro failing to exceed 9\% accuracy for high-precision measurement. This degradation suggests that current MLLMs rely primarily on approximate visual estimation rather than precise pixel-to-scale calibration, lacking the capability for sub-pixel measurement required in professional navigation contexts.

\textbf{Impact of Reasoning Mechanisms on Directional Estimation.}
A notable divergence is observed in the Bearing Calculation task (Table~\ref{tab:app-bearing-acc}). Unlike distance estimation, where extended reasoning models show marginal gains, the \textbf{Qwen3-VL-235B-Thinking} model achieves 55.64\% accuracy at the 20$^{\circ}$ threshold, significantly outperforming its standard instruction-tuned counterpart (36.95\%) and commercial SOTA models such as GPT-4o (25.15\%). This indicates that directional orientation benefits substantially from geometric reasoning. Chain-of-thought processes likely enable the model to explicitly deduce relative positions (e.g., inferring cardinal directionality) before committing to a numerical bearing, thereby mitigating the stochastic generation of angular values common in standard MLLMs.

\textbf{Disparity Between Visual and Symbolic Localization.}
We constructed a comparative analysis (Table~\ref{tab:app-combined-coord}) of coordinate localization performance across two modalities: \textit{Geographic} (requiring OCR and grid interpolation) and \textit{Pixel} (requiring pure visual localization). On ENC-Bench, MLLMs generally demonstrate lower proficiency in symbolic grounding compared to visual localization. For instance, GLM-4.5V achieves 14.38\% accuracy in pixel space (Acc@200px) but decreases to 9.52\% in geographic space. Similarly, Gemini-2.5-Pro performance drops from 21.43\% (Pixel) to 17.36\% (Geo). These results confirm that the primary bottleneck lies not in object detection, but in mapping visual detections to the semantic coordinate system defined by the chart's marginalia.

\textbf{Constraints on Fine-Grained Feature Localization.}
Across all models, performance approaches near-zero ($<5\%$) at the strictest localization threshold (Acc@50px), as shown in Table~\ref{tab:app-combined-coord}. This limitation in fine-grained precision can be attributed to three primary factors:
\begin{enumerate}
    \item \textbf{Sparse Feature Representation.} ENC symbols are often fine-grained (e.g., single-point features). Current visual encoders, typically optimized for global semantic alignment, may lack the spatial resolution to resolve these minute features with 50px tolerance.
    \item \textbf{Text-Image Alignment Accuracy.} Geographic localization requires precise interpolation between grid lines. Misalignments between text embeddings (reading grid numbers) and patch embeddings (perceiving grid lines) can introduce cumulative errors.
    \item \textbf{Input Resolution Limits.} Most models process images at fixed resolutions (e.g., $1024^2$), necessitating downsampling that obscures the precise pixel location of small navigational aids, rendering sub-100px accuracy mathematically challenging for certain feature sizes.
\end{enumerate}

\clearpage
\begin{table*}[t]
\centering
\caption{\textbf{Fine-Grained Distance Measurement Accuracy (Acc@T).} Evaluation of model performance across varying relative error tolerances. \textbf{Bold} indicates the best performance for each threshold.}
\label{tab:app-distance-acc}
\small
\setlength{\tabcolsep}{12pt}
\renewcommand{\arraystretch}{1.1}
\begin{tabular}{lcccc}
\toprule
\textbf{Model} & \textbf{Acc@5\%} & \textbf{Acc@10\%} & \textbf{Acc@15\%} & \textbf{Acc@20\%} \\
\midrule
Gemini-2.5-Flash & 6.92 & 11.76 & 17.99 & \textbf{25.93} \\
Gemini-2.5-Pro & \textbf{8.76} & \textbf{12.99} & \textbf{19.64} & 25.67 \\
Qwen3-VL-235B-Thinking & 6.10 & 10.50 & 15.20 & 20.33 \\
GPT-4o & 3.95 & 8.33 & 13.16 & 19.29 \\
Qwen3-VL-235B-Instruct & 5.95 & 9.52 & 13.69 & 16.71 \\
InternVL-3-38B & 3.50 & 6.80 & 10.20 & 13.09 \\
Qwen3-VL-32B-Instruct & 1.78 & 4.73 & 8.28 & 12.75 \\
GLM-4.5V & 0.60 & 3.31 & 6.33 & 9.04 \\
Llama-4-Maverick-17B & 0.93 & 3.10 & 5.20 & 7.41 \\
Qwen3-VL-32B-Thinking & 0.50 & 1.80 & 3.10 & 4.91 \\
\bottomrule
\end{tabular}
\end{table*}

\begin{table*}[t]
\centering
\caption{\textbf{Fine-Grained Bearing Calculation Accuracy (Acc@T).} Evaluation of angular precision with error thresholds of $5^{\circ}, 10^{\circ}, 20^{\circ},$ and $30^{\circ}$. \textbf{Bold} indicates the best performance for each threshold.}
\label{tab:app-bearing-acc}
\small
\setlength{\tabcolsep}{12pt}
\renewcommand{\arraystretch}{1.1}
\begin{tabular}{lcccc}
\toprule
\textbf{Model} & \textbf{Acc@5$^{\circ}$} & \textbf{Acc@10$^{\circ}$} & \textbf{Acc@20$^{\circ}$} & \textbf{Acc@30$^{\circ}$} \\
\midrule
Qwen3-VL-235B-Thinking & \textbf{16.20} & \textbf{31.50} & \textbf{55.64} & \textbf{70.20} \\
Gemini-2.5-Pro & 13.35 & 27.95 & 46.86 & 62.42 \\
Qwen3-VL-32B-Thinking & 11.50 & 24.80 & 44.10 & 59.50 \\
InternVL-3-38B & 10.80 & 22.50 & 41.09 & 56.80 \\
Qwen3-VL-235B-Instruct & 9.72 & 19.75 & 36.95 & 55.17 \\
GLM-4.5V & 5.52 & 12.99 & 28.60 & 52.60 \\
Gemini-2.5-Flash & 8.33 & 16.67 & 28.77 & 35.61 \\
Qwen3-VL-32B-Instruct & 7.29 & 11.85 & 26.44 & 41.64 \\
GPT-4o & 5.85 & 11.70 & 25.15 & 33.33 \\
Llama-4-Maverick-17B & 3.31 & 4.42 & 14.90 & 25.41 \\
\bottomrule
\end{tabular}
\end{table*}

\begin{table*}[t]
\centering
\caption{\textbf{Comparative Analysis of Coordinate Localization Modalities (Geo vs. Pixel).} Accuracy at varying pixel error thresholds for Geographic (Geo) and Pixel-based (Pix) localization. \textbf{Geo} requires symbolic grid interpretation; \textbf{Pix} requires only visual localization. \textbf{Bold} indicates the best performance for each column.}
\label{tab:app-combined-coord}
\small
\setlength{\tabcolsep}{5.5pt}
\renewcommand{\arraystretch}{1.1}
\begin{tabular}{lcccccccc}
\toprule
& \multicolumn{2}{c}{\textbf{Acc@50px}} & \multicolumn{2}{c}{\textbf{Acc@100px}} & \multicolumn{2}{c}{\textbf{Acc@200px}} & \multicolumn{2}{c}{\textbf{Acc@500px}} \\
\cmidrule(lr){2-3} \cmidrule(lr){4-5} \cmidrule(lr){6-7} \cmidrule(lr){8-9}
\textbf{Model} & \textbf{Geo} & \textbf{Pix} & \textbf{Geo} & \textbf{Pix} & \textbf{Geo} & \textbf{Pix} & \textbf{Geo} & \textbf{Pix} \\
\midrule
Gemini-2.5-Pro & 0.00 & 2.07 & \textbf{8.70} & \textbf{8.28} & \textbf{17.36} & \textbf{21.43} & 56.52 & 50.34 \\
Gemini-2.5-Flash & 0.76 & 1.45 & 3.05 & 5.80 & 16.77 & 21.03 & \textbf{58.02} & \textbf{56.52} \\
InternVL-3-38B & 0.50 & 0.80 & 2.50 & 2.50 & 13.19 & 10.52 & 48.20 & 35.40 \\
GPT-4o & 0.00 & 1.74 & 1.22 & 2.61 & 12.20 & 6.94 & 50.00 & 43.48 \\
Qwen3-VL-235B-Instruct & \textbf{4.29} & 2.04 & 4.29 & 2.04 & 11.31 & 12.20 & 45.71 & 38.78 \\
Qwen3-VL-235B-Thinking & 3.80 & \textbf{2.50} & 4.10 & 3.10 & 11.01 & 12.90 & 44.50 & 41.20 \\
GLM-4.5V & 3.95 & 0.96 & 5.26 & 3.85 & 9.52 & 14.38 & 52.63 & 47.12 \\
Llama-4-Maverick-17B & 1.14 & 0.00 & 1.14 & 3.23 & 9.13 & 12.90 & 46.59 & 40.32 \\
Qwen3-VL-32B-Thinking & 0.00 & 0.00 & 0.80 & 0.50 & 6.94 & 7.34 & 25.40 & 18.50 \\
Qwen3-VL-32B-Instruct & 0.00 & 0.00 & 0.00 & 0.00 & 5.26 & 5.16 & 18.33 & 13.33 \\
\bottomrule
\end{tabular}
\end{table*}
\clearpage

\section{Qualitative Analysis Case Studies}
\label{sec:appendix_cases}

In this section, we conduct a comprehensive case study analysis of the error types exhibited by three representative MLLMs (Gemini-2.5-Pro, GPT-4o, and Qwen3-VL-235B-Instruct) across the 30 qualitative examples in ENC-Bench. We categorize the errors into 5 distinct types based on the failure modes observed in the model outputs. Error Category Definitions:
\begin{itemize}
    \item \textbf{Visual Perception Error:} The model fails to recognize basic chart features, misidentifies symbols, or fails to locate labeled points. 
    \item \textbf{Reasoning Error:} The model correctly perceives visual elements but fails in the logical deduction required for the task.
    \item \textbf{Knowledge Error:} The model hallucinates the meaning of specialized IHO S-57 symbols.
    \item \textbf{Calculation Error:} In spatial tasks, the model fails to perform accurate numerical computation or pixel-to-geo mapping, resulting in large deviations in coordinates, bearings, or distances.
    \item \textbf{Instruction Following Error:} The model fails to adhere to the output format or specific constraints defined in the prompt.
\end{itemize}
Table~\ref{tab:case-index} provides a detailed index of all 30 case figures, mapping each sub-task to the specific error categories observed for each model.

\section{Limitations and Broader Impact}
\label{sec:appendix-impact}

In this section, we discuss the limitations and potential societal impact of this work.

\subsection{Potential Limitations}

While ENC-Bench provides a comprehensive benchmark for evaluating MLLMs in professional maritime domains, there are several limitations to consider:

\begin{itemize}
    \item \textbf{Geographic and Source Bias:} Our dataset is derived exclusively from NOAA (United States) Electronic Navigational Charts. Although these charts conform to the IHO S-57 standard, regional variations in charting practices exists across different hydrographic offices (e.g., UKHO, JHO). Consequently, findings based on this benchmark may not fully generalize to non-US waters or proprietary S-63 encrypted data formats used in commercial shipping.
    
    \item \textbf{Temporal and Dynamic Constraints:} Maritime navigation inherently involves continuous monitoring of temporal changes. As a static VQA benchmark, ENC-Bench evaluates snapshot interpretation capabilities but does not assess temporal reasoning (e.g., collision avoidance over time) or the integration of real-time sensor feeds such as Radar or AIS.
    
    \item \textbf{Simplified Decision Space:} To ensure rigorous evaluation, our L-3 decision-making tasks rely on closed-set options and explicit visual cues. Real-world operations involve open-ended decisions constrained by implicit factors—such as weather forecasts, tidal windows, and standing orders—that are not fully captured in visual charts alone.
    
    \item \textbf{Absence of Domain Adaptation:} This study focuses on the zero-shot evaluation of general-purpose MLLMs. We do not explore domain adaptation or instruction tuning on hydrographic data. While establishing a baseline for off-the-shelf capability, this may underestimate the potential of MLLMs when specifically aligned with maritime instructions.
\end{itemize}

\subsection{Potential Societal Impact}

\begin{itemize}
    \item \textbf{Safety Risks and Automation Bias:} Our results indicate that state-of-the-art MLLMs exhibit symbolic grounding gaps and reasoning hallucinations. In safety-critical domains where minor errors can lead to catastrophic outcomes, excessive reliance on these systems may lead to automation bias among operators. It is crucial to implement rigorous uncertainty quantification and maintain human supervision (Human-in-the-Loop) when deploying such models in operational environments.
    
    \item \textbf{Maritime Safety Advancement:} By establishing a rigorous standard for AI chart understanding, ENC-Bench supports the development of Intelligent Bridge Systems (IBS) and Maritime Autonomous Surface Ships (MASS). Reliable AI assistants can serve to cross-verify human decisions, potentially reducing accidents caused by human error or fatigue.
\end{itemize}

\clearpage
\definecolor{correct_green}{HTML}{D9EAD3}
\definecolor{visual_yellow}{HTML}{FFF2CC}
\definecolor{reason_red}{HTML}{F4CCCC}
\definecolor{know_purple}{HTML}{D9D2E9}
\definecolor{Calcul_blue}{HTML}{C9DAF8}

\newcommand{\cCorrect}{\cellcolor{correct_green}\textbf{Correct}}
\newcommand{\cPercept}{\cellcolor{visual_yellow}Visual Perception Error}
\newcommand{\cReason}{\cellcolor{reason_red}Reasoning Error}
\newcommand{\cKnow}{\cellcolor{know_purple}Knowledge Error}
\newcommand{\cCalcul}{\cellcolor{Calcul_blue}Calculation Error}
\newcommand{\cInstruct}{\cellcolor{orange!20}Instruction Following Error}

\begin{table*}[h]
\centering
\caption{\textbf{Index of case study figures by sub-tasks (L-1 to L-3) with associated error categories for representative MLLMs.} Errors are color-coded: \colorbox{visual_yellow}{Visual Perception Error}, \colorbox{reason_red}{Reasoning Error}, \colorbox{know_purple}{Knowledge Error}, \colorbox{Calcul_blue}{Calculation Error}, and \colorbox{orange!20}{Instruction Following Error}. \colorbox{correct_green}{\textbf{Correct}} indicates a successful prediction.}
\label{tab:case-index}
\resizebox{\textwidth}{!}{%
\begin{tabular}{llllll}
\toprule
\textbf{Figure} & \textbf{L-1/2/3 Task} & \textbf{Specific Sub-task} & \textbf{Gemini-2.5-Pro} & \textbf{GPT-4o} & \textbf{Qwen3-VL} \\
\midrule
Fig.~\ref{fig:case_1} & L-1 Perception & Symbol Recognition & \cReason & \cPercept & \cPercept \\
Fig.~\ref{fig:case_2} & L-1 Perception & Symbol Recognition (Cardinal) & \cKnow & \cPercept & \cKnow \\
Fig.~\ref{fig:case_3} & L-1 Perception & Symbol Recognition (Area) & \cKnow & \cKnow & \cKnow \\
\midrule
Fig.~\ref{fig:case_4} & L-1 Perception & Point Feature (Day) & \cPercept & \cPercept & \cKnow \\
Fig.~\ref{fig:case_5} & L-1 Perception & Point Feature (Dusk) & \cPercept & \cPercept & \cCorrect \\
Fig.~\ref{fig:case_6} & L-1 Perception & Point Feature (Night) & \cCorrect & \cPercept & \cCorrect \\
Fig.~\ref{fig:case_7} & L-1 Perception & Point Feature (Small Scale) & \cPercept & \cPercept & \cPercept \\
\midrule
Fig.~\ref{fig:case_8} & L-1 Perception & Linestring (Day) & \cPercept & \cCorrect & \cKnow \\
Fig.~\ref{fig:case_9} & L-1 Perception & Linestring (Dusk) & \cKnow & \cCorrect & \cPercept \\
Fig.~\ref{fig:case_10} & L-1 Perception & Linestring (Night) & \cPercept & \cPercept & \cPercept \\
Fig.~\ref{fig:case_11} & L-1 Perception & Linestring (Small Scale) & \cPercept & \cPercept & \cPercept \\
\midrule
Fig.~\ref{fig:case_12} & L-1 Perception & Polygon (Day) & \cPercept & \cKnow & \cPercept \\
Fig.~\ref{fig:case_13} & L-1 Perception & Polygon (Dusk) & \cCorrect & \cKnow & \cPercept \\
Fig.~\ref{fig:case_14} & L-1 Perception & Polygon (Night) & \cReason & \cPercept & \cPercept \\
Fig.~\ref{fig:case_15} & L-1 Perception & Polygon (Small Scale) & \cPercept & \cPercept & \cPercept \\
\midrule
Fig.~\ref{fig:case_16} & L-2 Spatial & Coordinate (Day) & \cCalcul & \cCalcul & \cCalcul \\
Fig.~\ref{fig:case_17} & L-2 Spatial & Coordinate (Dusk) & \cCalcul & \cCalcul & \cCalcul \\
Fig.~\ref{fig:case_18} & L-2 Spatial & Coordinate (Night) & \cCalcul & \cCalcul & \cCalcul \\
\midrule
Fig.~\ref{fig:case_19} & L-2 Spatial & Bearing (Day) & \cCalcul & \cCalcul & \cCalcul \\
Fig.~\ref{fig:case_20} & L-2 Spatial & Bearing (Dusk) & \cCalcul & \cCalcul & \cReason \\
Fig.~\ref{fig:case_21} & L-2 Spatial & Bearing (Night) & \cCalcul & \cCalcul & \cReason \\ 
\midrule
Fig.~\ref{fig:case_22} & L-2 Spatial & Distance (Day) & \cCalcul & \cCalcul & \cCalcul \\
Fig.~\ref{fig:case_23} & L-2 Spatial & Distance (Dusk) & \cCalcul & \cCalcul & \cCalcul \\
Fig.~\ref{fig:case_24} & L-2 Spatial & Distance (Night) & \cCalcul & \cCalcul & \cCalcul \\
\midrule
Fig.~\ref{fig:case_25} & L-3 Decision & Track Direction (Large) & \cPercept & \cPercept  & \cReason \\
Fig.~\ref{fig:case_26} & L-3 Decision & Track Direction (Small) & \cInstruct & \cPercept & \cPercept \\
\midrule
Fig.~\ref{fig:case_27} & L-3 Decision & Safety Assessment & \cPercept & \cReason & \cPercept \\
Fig.~\ref{fig:case_28} & L-3 Decision & Safety Assessment (Small) & \cPercept & \cPercept & \cPercept \\
\midrule
Fig.~\ref{fig:case_29} & L-3 Decision & Anchorage Selection & \cPercept & \cReason & \cPercept \\
Fig.~\ref{fig:case_30} & L-3 Decision & Anchorage Selection (Small) & \cPercept & \cReason & \cKnow \\
\bottomrule
\end{tabular}%
}
\end{table*}


\clearpage

\clearpage

\clearpage


\clearpage

\clearpage

\clearpage

\clearpage


\clearpage

\clearpage

\clearpage

\clearpage


\clearpage

\clearpage

\clearpage

\clearpage


\clearpage

\clearpage

\clearpage


\clearpage

\clearpage

\clearpage


\clearpage

\clearpage

\clearpage


\clearpage

\clearpage


\clearpage

\clearpage


\clearpage

\clearpage

\end{document}